\pdfoutput=1

\documentclass[11pt]{article}

\PassOptionsToPackage{table}{xcolor}

\usepackage[final]{acl}

\usepackage{times}
\usepackage{latexsym}
\usepackage{booktabs}
\usepackage{array}
\usepackage{graphicx}
\usepackage{siunitx}
\usepackage{adjustbox}
\usepackage{framed}
\usepackage{listings}
\usepackage{tabularx} 
\usepackage{longtable} 

\usepackage{subcaption}
\usepackage{placeins}
\usepackage{soul}
\usepackage{lipsum}
\usepackage{makecell}

\usepackage[T1]{fontenc}

\usepackage[utf8]{inputenc}

\usepackage{microtype}

\usepackage{inconsolata}

\usepackage{footmisc}
\usepackage{amsmath,amssymb}
\usepackage{algorithm}
\usepackage{algorithmic}
\usepackage{multirow}

\usepackage{parskip}
\usepackage{enumitem}
\usepackage[skins,breakable]{tcolorbox}

\newcommand{\MetaSynth}{\textsc{MetaSynth}}

\definecolor{darkgreen}{RGB}{0,100,0}

\definecolor{boxbg}{RGB}{235, 240, 255}
\definecolor{boxborder}{RGB}{100, 150, 255}
\definecolor{highlightorange}{RGB}{240, 140, 0}

\newcommand{\orangetag}[1]{{\color{highlightorange}\{#1\}}}

\title{\MetaSynth: Meta\textendash Prompting\textendash Driven Agentic Scaffolds 
for Diverse Synthetic Data Generation}

\author{
  \textbf{Haris Riaz\textsuperscript{1†*}},
  \textbf{Sourav Bhabesh\textsuperscript{2*}},
  \textbf{Vinayak Arannil\textsuperscript{2*}}, \\
  \textbf{Miguel Ballesteros\textsuperscript{2}},
  \textbf{Graham Horwood\textsuperscript{2}}
\\
\\
  \textsuperscript{1}University of Arizona,
  \textsuperscript{2}AWS AI Labs
\\
  \small{
    \textbf{Correspondence:} \href{mailto:hriaz@arizona.edu}{hriaz@arizona.edu}
  }
}

\begin{document}

\maketitle
\def\thefootnote{†}\footnotetext{This work was done while Haris Riaz was an intern at
Amazon Web Services.}
\def\thefootnote{*}\footnotetext{Equal contribution.}

\def\thefootnote{\arabic{footnote}}

\begin{abstract}
Recent smaller language models such Phi-3.5 and Phi-4 rely on synthetic data generated using larger Language models. Questions remain about leveraging synthetic data for other use cases, such as adapting LLMs to specific domains. A key limitation of synthetic data is \textit{low diversity}, which negatively impacts its downstream applicability for improving other models.
To address this, we propose \textsc{MetaSynth}, a method for generating synthetic data that enhances diversity through meta-prompting, where a language model orchestrates multiple ``expert'' LLM \textit{agents} to collaboratively generate data. Using only \textbf{25 million} tokens of synthetic data generated with \textsc{MetaSynth}, we successfully adapt a well-trained LLM (Mistral-7B-v0.3) to two specialized domains--Finance and Biomedicine--without compromising the capabilities of the resulting model in general tasks. In addition, we evaluate the diversity of our synthetic data using seven automated metrics, and find that it approaches the diversity of LLM pre-training corpora.

Continually pre-training Mistral-7B-v0.3 with \textsc{MetaSynth} notably outperforms the base LLM, showing improvements of up to 4.08\% in Finance and 13.75\% in Biomedicine. The same model shows degraded performance when trained on data generated using a template prompt, even when the template includes prior generations and varying In-Context exemplars of real data. Our findings suggest that a few million tokens of diverse synthetic data without mixing any real data, is sufficient for effective domain adaptation when using MetaSynth.
\end{abstract}

\section{Introduction}
Human generated public text data cannot sustain the continued scaling and expansion of large language models (LLMs). It has been argued by \newcite{villalobos2024run} that the available stock of public human text data will be fully utilized by 2028 if current LLM development trends continue, or earlier if LLMs are trained on more data than is compute optimal. This is evidenced in e.g., Llama 3 \cite{grattafiori2024llama3herdmodels} which uses one order of magnitude more data compared to only two year old estimates of compute optimal large language models \cite{hoffmann2022trainingcomputeoptimallargelanguage}.
As a potential remedy, synthetic data generated with LLMs has shown remarkable potential to alleviate the impending issue of data scarcity for future model scaling. 


However, low diversity is a key issue in any type of synthetic data. In this work, we hypothesize that there are two prominent reasons that affect the diversity of data synthesized by LLMs: a) the choice of seed instances used to initialize data generation and b) the prompts used, which commonly follow predefined templates, where variation in the prompt is mainly introduced via placeholders whose content is populated dynamically. 
Examples of data generation methods which use template-like prompts with variation include: \textit{Self-prompting} \cite{li-etal-2024-self-prompting}, \textit{Attrprompt} \cite{yu2023largelanguagemodelattributed}, \textit{CLINGEN} \cite{xu2025knowledgeinfusedpromptingassessingadvancing} and \textit{Explore-Instruct} \cite{wan2023exploreinstructenhancingdomainspecificinstruction}, among others. We contend that this variation yields limited diversity. For instance, generating a collection of domain-specific (e.g., financial) texts with similar prompts results in repetitive sentence structures—many texts begin with lexical patterns such as \texttt{``In today's ever-changing financial landscape''} or \texttt{``as the financial world evolves''}—and often contain recurring phrases, and generic buzzwords.




Recently, \newcite{suzgun2024metaprompting} find that \textit{Meta-prompting} \cite{zhang2024metapromptingaisystems} approaches -- where an LLM itself writes the prompts to solve a problem -- can elicit more diverse and creative outputs, significantly improving problem-solving capabilities for mathematical and algorithmic reasoning tasks, largely due to the feedback, self-verification, chain-of-thought \cite{wei2023chainofthoughtpromptingelicitsreasoning}, and planning dynamics inherent in these approaches. It has also been shown that using an optimized meta-prompt can improve the quality and downstream effectiveness of LLM generated synthetic data \cite{kim2024evaluatinglanguagemodelssynthetic}. We argue that a key use case for synthetic
data arises when abundant real data exists in the form of pre-training corpora, but one wishes to effectively tailor an LLM to a specific domain using only a small amount of carefully generated
synthetic data \cite{arannil2024dopaminedomainspecificpretrainingadaptation}. 
In this work, we investigate \textit{data-efficient domain adaptation} through meta-prompting, where a language model is instructed to act as a supervisor that writes specialized prompts for other models to collaboratively generate diverse data. 
Our contributions are as follows:\\
(1) We propose \textsc{MetaSynth}, a method to create diverse synthetic documents for continual pre-training (CPT) by leveraging a meta language model (which we refer to as meta-LM) and \textit{Conditional Instance Generation} -- where the meta-LM categorizes, and keeps track of each generated instance in memory, to ensure distinctness between them.
\\
(2) We propose \textsc{MetaSynth-\textit{Instruct}}, which can generate and iteratively evolve complex instructions for \textit{instruction pre-training}. Notably, this evolution is entirely driven by prompts written by the meta-LM itself. Furthermore, unlike other instruction-pretraining approaches e.g., 
(\cite{cheng-etal-2024-instruction}, \cite{cheng2024adaptinglargelanguagemodels}) 
our instructions are purely evolved from contexts synthesized by \textsc{MetaSynth} i.e., without using any human-written text (section \ref{instruction-gen}). \\
(3) \textsc{MetaSynth-\textit{Instruct}} can also synthesize training data for fine-tuning encoder models such as BERT \cite{devlin2019bertpretrainingdeepbidirectional}, RoBERTa \cite{liu2019robertarobustlyoptimizedbert} etc. We observe that encoders fine-tuned on this data can outpeform those fine-tuned on data generated with template-based prompting (section \ref{encoder-tasks}).\\
(4) We generate synthetic documents for continual pre-training by prompting an LLM with its prior outputs (memory) where the prompt follows a predefined template containing in-context exemplars of real data. 
However, when these synthetic documents are mixed with real data in a 1:1 ratio over 25M tokens, we find that it \textit{does not} improve the Mistral-7B base model, and even leads to slight performance degradation across two domains. In contrast, using 25M tokens of diverse synthetic data from MetaSynth yields substantial improvements to the base model across various ratios of mixing real and synthetic data.
Experiments on ten datasets in Finance and Biomedicine—evaluating nine mixing ratios following \citet{cheng2024adaptinglargelanguagemodels}--
indicate that \textbf{mixing real data with synthetic data is not needed if synthetic data is diverse}. (section \ref{metasynth-domain-adaptation-results}). \\
(5) We systematically measure the diversity of LLM generated synthetic data across multiple dimensions using seven automated metrics, including the \textit{Task2Vec} diversity coefficient \cite{lee2023scale}, which encapsulates formal notions of data diversity, among others (see Section \ref{diversity-section}). We find that our approach significantly improves the diversity of generated data relative to template-based prompting (section \ref{diversity-section}).
We argue that model degradation and ``model collapse'' (as discussed, inter alia, in \cite{shumailov2024curserecursiontraininggenerated, seddik2024badtrainingsyntheticdata, gerstgrasser2024modelcollapseinevitablebreaking}) can be avoided even when training solely on a small amount of synthetic data if it is sufficiently diverse. We present our method below, which we view through the lens of inference-time compute scaling, to ensure diversity in synthetic data. 





\begin{figure*}[t]
    \centering
    \includegraphics[width=0.9\textwidth, height=0.3\textheight, keepaspectratio]
    {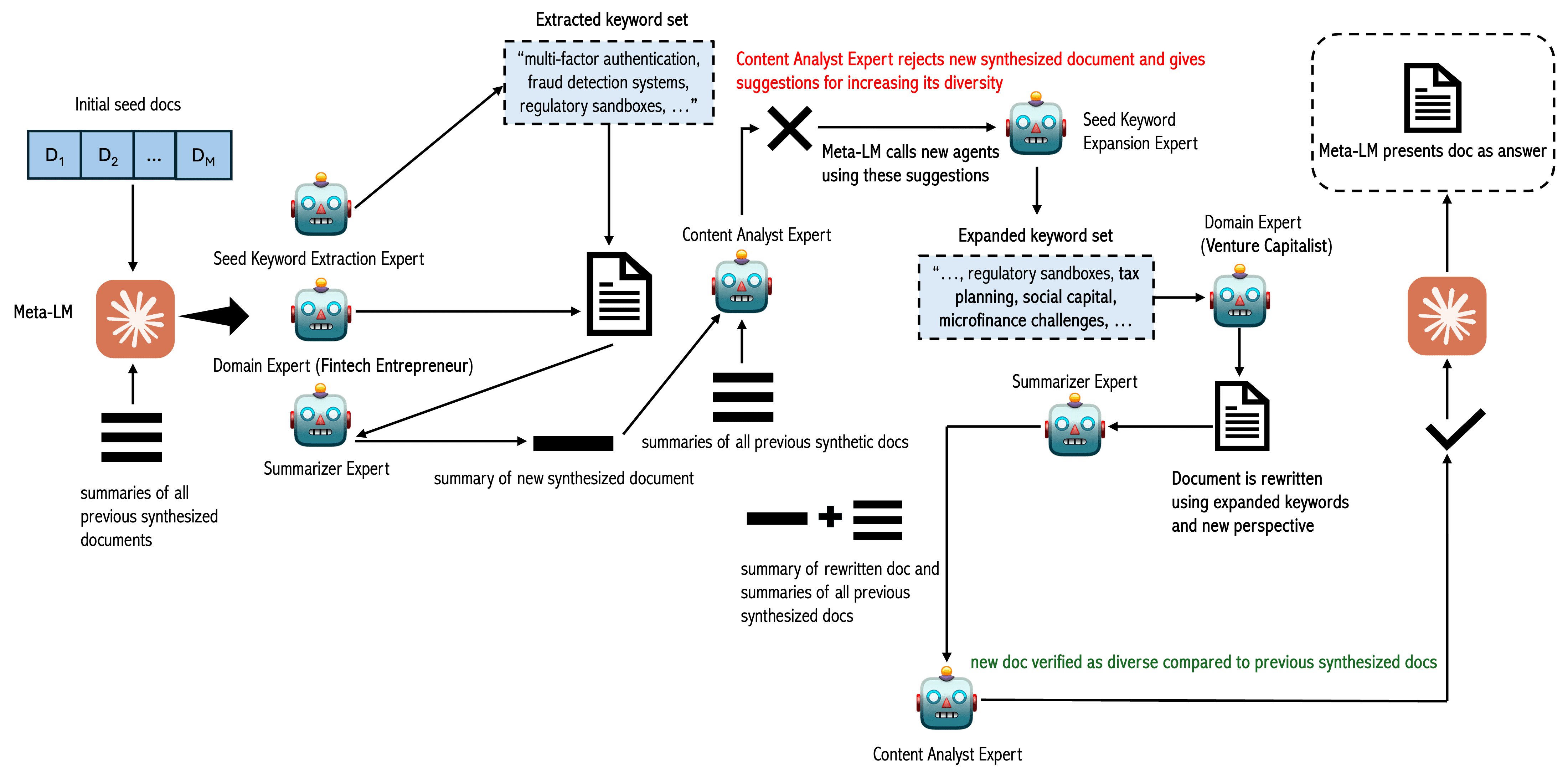}
    
    \caption{Demonstration of an example \textsc{MetaSynth} agentic workflow for synthesizing a financial document. A meta-LM orchestrates various expert agents that iteratively refine and generate diverse documents conditioned on an initial set of seed documents and previously synthesized documents. Refer to Section \ref{meta-prompting-execution} for a detailed description of the workflow.}
    \label{fig:metasynth_figure}
\end{figure*}

\begin{algorithm}[h] 
\footnotesize
\caption{Conditional Instance Generation}
\label{alg:conditional-generation}
\begin{algorithmic}[1]
\REQUIRE $S_{0}$: initial set of seeds;\\ $\theta$: parameters of data generating $\mathrm{LM}$; \\$\mathrm{div}(\cdot,\cdot)$: implicit diversity measure between a set of instances; \( T \in \mathbb{N} \)

\STATE \textbf{Step 1} Generate an initial instance:
$\displaystyle I_{0} \sim p(\,\cdot \mid S_{0}; \theta)$



\STATE \textbf{Step 2} Expand seed set and then generate another instance: \\ \STATE $\displaystyle
  S_{1} = \mathrm{ExpandSeeds}\bigl(S_{0}, \{I_{0}, I_{1}\}\bigr)
$
\STATE $\displaystyle
\begin{aligned}[t]
I_{1} 
&= \underset{I}{\arg\max} 
    \Bigl[
      p\bigl(I \mid I_{0}, S_{1}; \theta\bigr) 
      \times
      \mathbb{E}\bigl[\mathrm{div}(I_{0}, I)\bigr]
    \Bigr] \\
&\quad \therefore \text{subject to $I$ conforming to $S_{1}$}
\end{aligned}
$


\STATE \textbf{Step 3} Iteratively generate additional instances:
\FOR{$i = 2$ to $T$}
   \STATE $S_{i} = \mathrm{ExpandSeeds}\bigl(S_{i-1}, \{I_{0}, \dots, I_{i}\}\bigr)$
   \\ $\displaystyle
   \begin{aligned}[t]
   I_{i}
   &= \underset{I}{\arg\max}
      \Bigl[
         p\bigl(I \mid \{I_{0}, \dots, I_{i-1}\}, S_{i}; \theta\bigr)
         \\
         &\quad \times\;         \mathbb{E}\bigl[\mathrm{div}(\{I_{0}, \dots, I_{i-1}\}, I)\bigr]
      \Bigr] \\
   &\quad \therefore \text{subject to $I$ conforming to $S_{i}$}
   \end{aligned}
   $
\ENDFOR

\STATE \textbf{return} $\{I_{0}, \dots, I_{T}\}$
\end{algorithmic}
\end{algorithm}

\section{Meta-Prompting}
\subsection{Meta-LM}
\label{agentic-scaffolding}
At a high level our procedure for synthesizing a diverse collection of documents leverages two ideas: Meta-Prompting \cite{suzgun2024metaprompting, zhang2024metapromptingaisystems} and \textit{Conditional Instance Generation} (algorithm \ref{alg:conditional-generation}). 
Meta-prompting leverages a central meta-LM to coordinate and execute multiple independent inquiries and subsequently synthesize their responses to render a final response. This is realized via a high-level ``meta'' prompt which instructs an LM to break down complex tasks (such as generating a diverse collection of documents) into smaller or more manageable subtasks.  Each of these subtasks is assigned to specialized expert models (also known as ``agents'') where both the choice of the agent and the instructions to the agent are dynamically composed by the meta-LM depending upon the nature of the subtask.  In this work, we adapt the task agnostic meta-prompt from \newcite{suzgun2024metaprompting} to specifically focus on generating diverse synthetic data. The meta-LM serves as an orchestrator overseeing communication between these agents in a centralized multi-agent system (MAS) \cite{ijcai2024p890}, where the agents cannot cannot directly interact with each other; and also carries forward the thread of the process by applying its own critical thinking, reasoning and verification skills throughout. 
Further, to enable conditional instance generation, the meta-LM is equipped with memory to become stateful--a message history comprising it's own responses (which include the selection of agents and formulation of instructions for them) and the responses from various agents. \textit{Only the meta-LM} has access to the complete history, while the agents it invokes are limited to selectively shared information, seeing only what the meta-LM chooses to share with them. Being provided with only partial information pertaining to the task to solve, allows an agent to consider new perspectives with ``fresh eyes'' \cite{suzgun2024metaprompting} and potentially correct the meta-LM's errors. In this work, we use Claude 3 Sonnet \cite{anthropic2024claude3} as the meta-LM.
We further motivate the need for agentic scaffolding by drawing an analogy to multidisciplinary problem solving: complex tasks are often best addressed by leveraging diverse expertise rather than relying on a single, monolithic approach. As shown by \newcite{wu2023autogenenablingnextgenllm,yao2023reactsynergizingreasoningacting}, distinct agents can specialize in decision making, problem decomposition, and mitigating issues such as error propagation in chain-of-thought reasoning.
To generate a single synthetic instance (e.g., document or instruction), the meta-LM can invoke agents arbitrarily. However, to ensure that: a) each synthetic instance (document or instruction) is sufficiently distinct from all previously generated instances, and b) the meta procedure does not degenerate: we impose specific constraints within the meta-prompt which specify that certain types of agents must always be invoked, accompanied by an in-context exemplar that demonstrates the invocation process for those agents. The required agents depend on the task—whether synthesizing documents, instructions, or instances from an existing dataset (refer to the meta-prompts in Appendix~\ref{sec:appendixJ} and~\ref{sec:appendixK}). Without these constraints,\footnote{Even with these constraints, degeneration can still occur due to noisy message passing between the meta-LM and agents.} the procedure risks degenerative loops, where repetitive exchanges between a meta-LM and agent(s) may hinder task completion.  
Beyond this, the procedure remains open-ended, allowing the meta-LM to leverage any type of agent to enhance instance diversity in the final set of synthesized instances. In this work, the meta-LM (Claude 3 Sonnet) also serves as the agent LM, though any advanced instruction-following models can fulfill these roles. 




\textbf{Conditional Instance Generation}
Our document synthesis approach relies on a continuously expanding \textit{instance classification table} 
appended to the meta-LM’s history in each iteration, tracking and categorizing previously generated instances. 
After each newly synthesized instance (i.e., a document), the set of seed keywords is expanded with related yet distinct terms. Each newly synthesized document must also satisfy the following two criteria: conform to the current seed set, and be distinct from all previously synthesized documents. 
This process is guided by two agents: the ``Seed Keyword Expansion Expert'' and the ``Content Analyst Expert.'' Documents are compared via summaries (generated by a ``Summarizer Expert''). Summarization mitigates LLM context window limitations when keeping track of a large set of prior documents. The Content Analyst Expert categorizes each document, and suggests diversity-enhancing modifications, such as expanding the seed keyword set with related terms or incorporating new personas. The idea of conditional instance generation for synthesizing documents using seed
keywords is expressed in algorithm \ref{alg:conditional-generation} and the meta-prompting procedure that we adapt from \newcite{suzgun2024metaprompting} is shown in algorithm \ref{alg:meta-prompting-algorithm}, appendix \ref{sec:appendixA}.

\textbf{Exit Criteria and Error Handling} 
At each iteration, the meta-LM, conditioned on its history, must either call an agent or return a final response marked by the \texttt{<end>} token (indicating that the desired number of instances have been synthesized from the initial seeds). Otherwise, an error is appended to its history and the model is prompted to retry. After $N$ attempts the iteration is discarded.

\subsection{Execution}
\label{meta-prompting-execution}
Figure \ref{fig:metasynth_figure} shows an example 
agentic workflow that synthesizes a new financial domain document given initial seed documents and previously synthesized documents: \textbf{(1)} The meta-LM consults a ``Seed Keyword Extraction Expert,'' a ``Domain Expert,'' and a ``Summarizer Expert.''
\textbf{(2)} The Seed Keyword Extraction Expert extracts representative keywords (e.g., ``\texttt{multi-factor authentication},'' ``\texttt{fraud detection},'' ``\texttt{regulatory sandboxes}''), which the meta-LM uses to instruct a Domain Expert (e.g., a ``Fintech Entrepreneur'') to generate a document. 
\textbf{(3)} The Domain Expert writes the document, which the Summarizer Expert condenses before the meta-LM accepts it. 
\textbf{(4)} The meta-LM then instructs the Domain Expert to generate a second document that adheres to the same keywords while differing in content and style. 
\textbf{(5)} To verify diversity, the meta-LM consults the Summarizer Expert and a ``Content Analyst Expert.'' If the second document is deemed too similar to the first, the Content Analyst provides feedback. 
\textbf{(6)} In response, the meta-LM calls a ``Seed Keyword \textit{Expansion} Expert'' to enrich the keyword set and instructs a new Domain Expert (e.g., a ``Venture Capitalist'') to rewrite the document from a fresh perspective. \textbf{(7)} The Summarizer and Content Analyst Experts reassess the revised document, and a ``Writing/Linguistic Expert'' may be consulted for stylistic diversity. Once confirmed as sufficiently distinct, the document is accepted, and the process continues for generating subsequent documents. For a more thorough illustration of ``fresh eyes'' and ``conditional instance generation'', refer to the example meta-prompting execution history in figure \ref{fig:metasynth_history}, appendix \ref{sec:metasynth-execution-history}.
\begin{figure*}[t]
\setlength{\tabcolsep}{4pt}  
\begin{minipage}{\textwidth}
\renewcommand{\arraystretch}{2.5}  
\makebox[\textwidth][c]{%
\resizebox{\textwidth}{!}{%
\begin{tabular}{>
{\raggedright\arraybackslash}p{5.2cm}|c|c|c|c|c|c|c}
\hline
\textbf{Setting} & \normalsize{\textbf{Compression Ratio $\downarrow$}} & \normalsize{\textbf{Task2Vec Div. Coeff $\uparrow$}} & \normalsize{\textbf{Remote Clique $\uparrow$}} & \normalsize{\textbf{Chamfer Distance $\uparrow$}} & \normalsize{\textbf{1-GD $\uparrow$}} & \normalsize{\textbf{4-GD $\uparrow$}} & \normalsize{\textbf{MIF $\uparrow$}} \\
\hline
{\large Template Prompting} & {\Large \underline{3.6674}} & {\Large \underline{0.1576}} & {\Large \underline{0.1964}} & {\Large \underline{0.0897}} & {\Large \underline{0.0198}} & {\Large \underline{0.9224}} & {\Large \underline{8.5614}} \\
\hline
{\large Common Crawl} & {\Large 2.7380} \textcolor{darkgreen}{{\Large (-25.34\%)}} & {\Large 0.212} \textcolor{darkgreen}{{\Large (+34.52\%)}} & {\Large 0.3036} \textcolor{darkgreen}{{\Large (+54.58\%)}} & {\Large 0.2359} \textcolor{darkgreen}{{\Large (+162.99\%)}} & {\Large 0.0621} \textcolor{darkgreen}{{\Large (+213.64\%)}} & {\Large 1.6080} \textcolor{darkgreen}{{\Large (+74.33\%)}} & {\Large 8.1263} \textcolor{red}{{\Large (-5.08\%)}} \\
\hline
{\large Synth. Docs (Seed Keywords)} & {\Large 3.4443} \textcolor{darkgreen}{{\Large (-6.08\%)}} & {\Large 0.1757} \textcolor{darkgreen}{{\Large (+11.49\%)}} & {\Large 0.2191} \textcolor{darkgreen}{{\Large (+11.56\%)}} & {\Large 0.1351} \textcolor{darkgreen}{{\Large (+50.61\%)}} & {\Large 0.0345} \textcolor{darkgreen}{{\Large (+74.24\%)}} & {\Large 1.1749} \textcolor{darkgreen}{{\Large (+27.37\%)}} & {\Large 9.0016} \textcolor{darkgreen}{{\Large (+5.14\%)}} \\
\hline
{\large Synth. Docs (Seed Documents)} & {\Large 3.1495} \textcolor{darkgreen}{{\Large (-14.12\%)}} & {\Large 0.1788} \textcolor{darkgreen}{{\Large (+13.45\%)}} & {\Large 0.2047} \textcolor{darkgreen}{{\Large (+4.23\%)}} & {\Large 0.1383} \textcolor{darkgreen}{{\Large (+54.18\%)}} & {\Large 0.0390} \textcolor{darkgreen}{{\Large (+96.97\%)}} & {\Large 1.3468} \textcolor{darkgreen}{{\Large (+46.01\%)}} & {\Large 8.9150} \textcolor{darkgreen}{{\Large (+4.13\%)}} \\
\hline
{\large Wikipedia} & {\Large 2.6088} \textcolor{darkgreen}{{\Large (-24.82\%)}} & {\Large 0.1892} \textcolor{darkgreen}{{\Large (+20.05\%)}} & {\Large 0.2868} \textcolor{darkgreen}{{\Large (+46.03\%)}} & {\Large 0.2416} \textcolor{darkgreen}{{\Large (+169.34\%)}} & {\Large 0.1046} \textcolor{darkgreen}{{\Large (+428.28\%)}} & {\Large 1.6997} \textcolor{darkgreen}{{\Large (+84.27\%)}} & {\Large 8.3149} \textcolor{red}{{\Large (-2.88\%)}} \\
\hline
\end{tabular}}}
\caption*{\footnotesize {(a) Evaluating diversity metrics of synthetic data generation methods from \textbf{finance} domain.}}
\makebox[\textwidth][c]{%
\resizebox{\textwidth}{!}{%
\begin{tabular}{>
{\raggedright\arraybackslash}p{5.5cm}|c|c|c|c|c|c|c}
\hline
\textbf{Setting} & \textbf{Compression Ratio $\downarrow$} & \textbf{Task2Vec Div. Coeff $\uparrow$} & \normalsize{\textbf{Remote Clique $\uparrow$}} & \textbf{Chamfer Distance $\uparrow$} & \textbf{1-GD $\uparrow$} & \textbf{4-GD $\uparrow$} & \textbf{MIF $\uparrow$} \\
\hline
{\large Template Prompting} & {\Large \underline{3.4699}} & {\Large \underline{0.1575}} & {\Large \underline{0.2295}} & {\Large \underline{0.1056}} & {\Large \underline{0.0278}} & {\Large \underline{1.0035}} & {\Large \underline{8.7463}} \\
\hline
{\large Common Crawl} & {\Large 2.6717} \textcolor{darkgreen}{{\Large (-23.00\%)}} & {\Large 0.2068} \textcolor{darkgreen}{{\Large (+31.30\%)}} & {\Large 0.3130} \textcolor{darkgreen}{{\Large (+36.38\%)}} & {\Large 0.2451} \textcolor{darkgreen}{{\Large (+132.10\%)}} & {\Large 0.0703} \textcolor{darkgreen}{{\Large (+152.88\%)}} & {\Large 1.6524} \textcolor{darkgreen}{{\Large (+64.66\%)}} & {\Large 8.2744} \textcolor{red}{{\Large (-5.40\%)}} \\
\hline
{\large Synth. Docs (Seed Keywords)} & {\Large 3.1537} \textcolor{darkgreen}{{\Large (-9.11\%)}} & {\Large 0.1760} \textcolor{darkgreen}{{\Large (+11.75\%)}} & {\Large 0.2277} \textcolor{red}{{\Large (-0.79\%)}} & {\Large 0.1426} \textcolor{darkgreen}{{\Large (+35.04\%)}} & {\Large 0.0403} \textcolor{darkgreen}{{\Large (+44.96\%)}} & {\Large 1.3323} \textcolor{darkgreen}{{\Large (+32.77\%)}} & {\Large 8.9503} \textcolor{darkgreen}{{\Large (+2.33\%)}} \\
\hline
{\large Synth. Docs (Seed Documents)} & {\Large 3.0649} \textcolor{darkgreen}{{\Large (-11.67\%)}} & {\Large 0.1793} \textcolor{darkgreen}{{\Large (+13.84\%)}} & {\Large 0.2395} \textcolor{darkgreen}{{\Large (+4.36\%)}} & {\Large 0.1478} \textcolor{darkgreen}{{\Large (+39.96\%)}} & {\Large 0.0432} \textcolor{darkgreen}{{\Large (+55.40\%)}} & {\Large 1.3794} \textcolor{darkgreen}{{\Large (+37.46\%)}} & {\Large 8.9044} \textcolor{darkgreen}{{\Large (+1.81\%)}} \\
\hline
{\large Wikipedia} & {\Large 2.6088} \textcolor{darkgreen}{{\Large (-24.82\%)}} & {\Large 0.1892} \textcolor{darkgreen}{{\Large (+20.05\%)}} & {\Large 0.2868} \textcolor{darkgreen}{{\Large (+46.03\%)}} & {\Large 0.2416} \textcolor{darkgreen}{{\Large (+169.34\%)}} & {\Large 0.1046} \textcolor{darkgreen}{{\Large (+428.28\%)}} & {\Large 1.6997} \textcolor{darkgreen}{{\Large (+84.27\%)}} & {\Large 8.3149} \textcolor{red}{{\Large (-2.88\%)}} \\
\hline
\end{tabular}}}
\caption*{\footnotesize {(b) Evaluating diversity metrics of synthetic data generation methods from \textbf{biomedicine} domain.}}
\caption{\footnotesize{Metrics are annotated with $\uparrow$ or $\downarrow$ arrows which indicate if higher or lower values are better, respectively. \textbf{1-GD} refers to 1-Gram diversity and \textbf{4-GD} refers to 4-Gram diversity. \textbf{MIF} refers to the Mean Inverse Frequency metric (refer to section \ref{diversity-section}). For a particular domain, diversity metrics for synthetic data generated using template prompting the base LLM are \underline {underlined} as reference points. We include diversity metrics over a subset of Wikipedia as a generic example of a dataset regarded to be diverse. For each synthetic data generation method and each metric, percentage increases in diversity relative to template prompting are shown in parentheses. Improvements in measured diversity are highlighted in \textcolor{darkgreen}{green} and reductions in diversity are highlighted in \textcolor{red}{red}.} All metrics are mean values of 95\% CI computed with boostrap resampling (refer to Appendix \ref{sec:boostrap-resampling}). We control for length in all diversity comparisons by constraining synthetic documents to 400 words (Section~\ref{document-gen-seed-document-approach}) and sampling from a similar-length distribution for other sources (e.g., Common Crawl, Wikipedia; Appendix~\ref{sec:appendixD}).}
\label{fig:diversity-metrics}
\end{minipage}
\end{figure*}

\section{Synthetic Data Generation}
\label{synthdata-details}

\subsection{Baseline: Template Prompting}
\label{template-prompting}
We introduce a strong baseline for synthetic data generation that uses a static template-based prompt with a placeholder populated by five-shot examples of real documents randomly selected from a domain specific subset of Common Crawl\footnote{\url{https://commoncrawl.org/}} (refer to appendix \ref{sec:appendixI} for the template prompt used in this work). Additionally, this generation process is also \textit{conditional}, as the data generator is equipped with memory, allowing it to reference previously generated documents while being instructed to ensure that each new document remains distinct from prior outputs. Refer to appendix \ref{sec:appendixM} for examples of documents synthesized with template prompting.



\subsection{MetaSynth: Synthetic Document Generation}
\label{document-gen-seed-document-approach}

\textbf{Random Seed Selection} 
For generating synthetic documents, we propose two methods for selecting a set of seed instances. The first is keyword based which initializes the generation process using random domain-specific keywords synthesized by an agent. 

\textbf{Topic-Aware Seed Selection}
We introduce a second \textit{topic-aware} seed selection approach using a dynamically adaptive $k$-NN algorithm. Starting with $N$ seed documents from domain-specific Common Crawl, each assigned an LLM-generated topic label (via ``Topic Labeling Expert''), we update the seed set after every $M$ MetaSynth-generated documents. New seeds are retrieved from the $k$ nearest neighbors of synthesized documents in embedding space, with each new seed selected such that its topic label (assigned via ``Topic Labeling Expert'') differs from the topic labels of the $M$ documents that were synthesized using the current seed set. If not enough candidate seeds meeting these criteria are found, $k$ is incremented\footnote{Initially we set $k=5$; embeddings are computed using \url{https://huggingface.co/jinaai/jina-embeddings-v2-base-en}}. This approach ensures topical variation while maintaining semantic relevance to the initial seeds.  
To prevent seed data leakage (see Table~\ref{tab:data_contamination}), MetaSynth always extracts keywords via the ``Seed Keyword Extraction Expert'', ensuring data synthesis is keyword-driven, regardless of whether seeds are documents or keywords. Motivated by \newcite{eldan2023tinystoriessmalllanguagemodels}, who show that short, diverse, grammatically correct texts (\textit{TinyStories}) induce language learning in small LMs, we cap the length of both synthesized and seed documents at 400 words (approximately 530 tokens). Appendix \ref{sec:appendixN} contains examples of MetaSynth generated documents.
\subsection{MetaSynth-\textit{Instruct}: Synthetic Instruction Synthesis \& Evolution Using Synthetic Documents}
\label{instruction-gen}
We design a meta-prompting driven instruction synthesizer to leverage the synthetic documents synthesized in the previous step (section \ref{document-gen-seed-document-approach}) to derive and evolve complex instructions. As part of the meta-prompt, we use a \textit{task description} string to define an instruction as a complex problem about a particular context leveraging various formats and styles e.g. reading comprehension, multiple-choice, fill-in-the-blank and inferential questions.
To prevent instruction synthesis from degenerating, the meta-prompt for instruction synthesis also contains invocation calls for a certain group of predetermined agents to always be invoked e.g. \texttt{Document Transformation Expert}, \texttt{Persona Suggestion Expert} (inspired by \newcite{ge2024scalingsyntheticdatacreation}), 
\texttt{Complexity Expert} and \texttt{Question Editor Expert} (similar to the suggestor-editor agents proposed by \textit{AgentInstruct} \cite{mitra2024agentinstructgenerativeteachingagentic}). In contrast to the work by \newcite{xu2023wizardlmempoweringlargelanguage} and \newcite{honovich2022unnaturalinstructionstuninglanguage}, in our method the instruction evolution prompts (which involves choosing the method of evolution) are open-ended; composed by the meta-LM taking into account the content of the synthesized document, the responses of agents from previous execution steps and it's own best judgment. We illustrate an example agentic flow for MetaSynth-\textit{Instruct} in Appendix \ref{sec:appendixH} and Figure \ref{fig:metasynth_instruct_figure}. Synthesized instructions (Appendix \ref{sec:appendixK}, Appendix \ref{sec:appendixO}) are limited to 100 words and the responses to each instruction are generated using Claude 3 Sonnet with varied prompt formats (see Appendix~\ref{sec:appendixF}).

\section{Measuring The Diversity of Generated Synthetic Data}
\label{diversity-section}
The premise of this work is that diverse data is high quality data. Thus, in lieu of human judgment of diversity; it is necessary to use an appropriate set of automated metrics which can quantify the diversity of LLM generated data such that these measures also align with \textit{human notions of variability and diversity}. 
\subsection{Metrics:}
\textbf{Task2Vec Diversity Coefficient} 
To quantify semantic and structural diversity in MetaSynth-generated data, we adopt the \textit{Task2Vec} diversity coefficient from \newcite{lee2023scale}. \textit{Task2Vec} formalizes diversity by embedding sampled data batches (e.g., synthesized documents) using the Fisher Information Matrix of a probe network\footnote{We use GPT-2 \cite{Radford2019LanguageMA} as the probe network} fine-tuned on the data . The coefficient, defined as the average pairwise cosine distance between \textit{Task2Vec} embeddings \cite{achille2019task2vectaskembeddingmetalearning}, has been shown to correlate with human diversity judgments.

\textbf{Compression Ratio \& \textit{N}-Gram Diversity} 
Following the recommendations of \newcite{shaib2024detectionmeasurementsyntactictemplates, shaib2024standardizingmeasurementtextdiversity}, we select $\mathcal{G}$zip compression ratio and \textit{N}-Gram diversity score (ratio of the unique n-gram counts to all n-gram counts in a dataset) as appropriate metrics which can detect aspects of repetition in LLM generated texts (such as the presence of pre-defined syntactic templates).

\textbf{Remote-Clique \& Chamfer Distance} Following (\newcite{10.1145/3411764.3445782,li2023syntheticdatagenerationlarge}), we also compute language model embedding based diversity with the \textit{Remote Clique Score} (average mean pairwise distance of a data instance to other instances) and the \textit{Chamfer Distance Score} (average minimum pairwise distance of a data instance to other instances).

\textbf{Mean Inverse Frequency (MIF) Score} We propose an additional metric which captures the average ``lexical rarity'' of synthesized documents, where instances that use a rarer vocabulary (relative to a reference corpus such as Wikipedia) are assigned high scores, and vice versa, somewhat similar to Inverse Document Frequency from \textit{TF-IDF} \cite{Ramos2003UsingTT}. Refer to appendix \ref{sec:diversity-appendix} for further details on diversity metrics.

Figure~\ref{fig:diversity-metrics} shows that MetaSynth documents seeded with Common Crawl are more diverse than those seeded with random keywords, with both exceeding the diversity of template-prompted documents.

\begin{table*}[th!]
  \footnotesize
  \centering
  \resizebox{\textwidth}{!}{%
  \begin{tabular}{l|c|c|c|c|c|c|c}
      \toprule
      \rowcolor[rgb]{0.93,0.93,0.93}\multicolumn{8}{c}{\textit{Finance}}\\
      \midrule
      CPT Setting & Token Mix & ConvFinQA & NER & FPB & Headline & FiQA\_SA & Average \\
      \midrule
      Mistral-7B Base (No CPT) & 0M & 38.9 & \textbf{58.14} & 65.09 & 79.26 & 75.62 &  63.40 \\
      \midrule
      \underline{Real Docs + Template-Prompting Docs} & 12.5M:12.5M & 48.79 & 52.64 & 64.24 & 76.00 & 74.47 & 63.23 \\
      \midrule
      Real Docs & 25M & 46.51 & 55.59 & 65.07 & 78.30 & 76.09 & 64.31 \\
      \midrule
      \underline{Real Docs + MetaSynth Docs} & 12.5M:12.5M & 48.59 & 53.69 & \textbf{67.82} & 80.14 & 75.66 & 65.18 \\
      \midrule
      Real Docs + MetaSynth Docs-Instructions-Responses & 12.5M:12.5M & 43.29 & 53.77 & 62.06 & 79.75 & 71.57 & 62.09 \\
      \midrule
      Real Docs + MetaSynth Docs-Instructions-Responses & 8.33M:16.7M & 43.22 & 52.26 & 65.66 & 79.73 & 72.50 & 62.67 \\
      \midrule
      Real Docs + MetaSynth Instructions-Responses & 12.5M:12.5M & 47.51 & 52.08 & 63.16 & 79.53 & 72.56 & 62.97 \\
      \midrule
      Real Docs + MetaSynth Instructions-Responses & 8.33M:16.7M & 44.43 & 49.34 & 63.05 & 79.68 & 75.27 & 62.35 \\
      \midrule
      MetaSynth Docs & 25M & 42.28 & 48.72 & 67.37 & 79.67 & 73.65 & 62.34 \\
      \midrule
      MetaSynth Docs-Instructions-Responses & 25M & \textbf{49.30} & 54.64 & 66.43 & \textbf{83.46} & \textbf{76.13} & \textbf{65.99} \\
      \midrule
      \rowcolor[rgb]{0.93,0.93,0.93}\multicolumn{8}{c}{\textit{Biomedicine}}\\
      \midrule
      CPT Setting & Token Mix & PubMedQA & USMLE & MQP & RCT & ChemProt & Average \\
      \midrule
      Mistral-7B (No CPT) & 0M & 58.20 & 35.27 & 67.86 & 62.55 & 40.80 & 52.94 \\
      \midrule
      \underline{Real Docs + Template-Prompting Docs} & 12.5M:12.5M & 56.40 & \textbf{38.41} & 67.38 & 59.80 & 30.40 & 50.48 \\
      \midrule
      Real Docs & 25M & 59.70 & 36.37 & 62.29 & 63.70 & 28.90 & 50.19 \\
      \midrule
      \underline{Real Docs + MetaSynth Docs} & 12.5M:12.5M & 60.70 & 37.31 & 64.26 & 67.50 & 45.00 & 54.95 \\
      \midrule
      Real Docs + MetaSynth Docs-Instructions-Responses & 12.5M:12.5M & 60.30 & 37.16 & 74.75 & 71.85 & 38.40 & 56.49 \\
      \midrule
      Real Docs + MetaSynth Docs-Instructions-Responses & 8.33M:16.7M & 59.50 & 36.61 & 76.06 & 71.05 & 42.20 & 57.08 \\
      \midrule
      Real Docs + MetaSynth Instructions-Responses & 12.5M:12.5M & \textbf{62.90} & 35.98 & 71.80 & 71.40 & 39.60 & 56.34 \\
      \midrule
      Real Docs + MetaSynth Instructions-Responses & 8.33M:16.7M & 60.20 & 36.44 & 73.77 & 71.75 & 42.10 & 56.85 \\
      \midrule
      MetaSynth Docs & 25M & 60.20 & 37.23 & 70.16 & 68.15 & 40.40 & 55.23 \\
      \midrule
      MetaSynth Docs-Instructions-Responses & 25M & 61.80 & 36.60 & \textbf{77.87} & \textbf{74.45} & \textbf{50.40} & \textbf{60.22} \\
      \midrule
  \end{tabular}%
  }
  \caption{\footnotesize Performance on domain-specific tasks for Mistral-7B under \textbf{nine} different continual pre-training (CPT) settings with varying mixing ratios of real and synthetic data. \textbf{Bold} indicates the best result for a dataset across all settings within a particular domain. Settings are \underline{underlined} to indicate the corresponding setting from MetaSynth which can be compared with Template-Prompting.}
  \label{tab:mistral-7B-cpt}
\end{table*}

\begin{table*}
    \centering
    \small
    \label{tab:general-eval-results}
    \begin{adjustbox}{max width=\textwidth}
    \begin{tabular}{l|cccccccccc}
    \hline
    & ARC-ch & ARC-easy & BoolQ & HellaSwag & MMLU & OBQA & PIQA & SIQA & Winogrande & Avg \\
    \hline
    \multicolumn{11}{l}{\textbf{Base Model}}\\
    Mistral-7B & 52.1 & 78.4 & 82.0 & 80.4 & 59.1 & 44.2 & 82.3 & 45.9 & 73.4 & 66.4 \\
    \hline
    \multicolumn{11}{l}{\textbf{Finance}}\\
     Real Docs + Template Prompting Docs & 53.8 & 78.4 & 78.0 & 80.7 & 59.0 & 45.6 & 81.9 & 48.1 & 71.4 & 66.3 \\
    Real Docs + MetaSynth Docs & 55.9 & 77.3 & 84.3 & 80.7 & 58.5 & 44.4 & 81.1 & 49.6 & 71.7 & 67.1 \\
    MetaSynth Docs-Instr-Responses & 50.9 & 75.1 & 84.1 & 79.4 & 56.3 & 43.0 & 80.7 & 48.1 & 69.3 & 65.2 \\
    \hline
    \multicolumn{11}{l}{\textbf{Biomedicine}}\\
    Real Docs + Template Prompting Docs & 54.9 & 79.5 & 80.8 & 81.1 & 58.1 & 45.4 & 82.6 & 46.9 & 71.7 & 66.8 \\
    Real Docs + MetaSynth Docs & 53.4 & 76.1 & 83.5 & 80.6 & 58.0 & 44.6 & 81.0 & 46.9 & 70.2 & 66.0 \\
    MetaSynth Docs-Instr-Responses & 54.2 & 75.2 & 83.2 & 79.1 & 57.5 & 43.2 & 81.0 & 47.5 & 70.8 & 65.8 \\
    \hline
    \end{tabular}
    \end{adjustbox}
    \caption{\footnotesize General evaluation across domains and Settings. \texttt{Real docs + Template Prompting docs} refers to Continual Pre-training (CPT) over 12.5M tokens of synthetic data generated with template prompting mixed with 12.5 tokens of Common Crawl data. \texttt{Real Docs + MetaSynth Docs} refers to CPT over 12.5M tokens of synthetic data generated by our method mixed with 12.5M tokens of Common Crawl data. \texttt{MetaSynth Docs-Instr-Responses} refers to CPT over 25M tokens of MetaSynth documents and their associated synthetic instruction-response pairs.}
    \label{tab:general-eval}
\end{table*}

\begin{figure}[t]  
    \centering
    \includegraphics[width=\columnwidth]{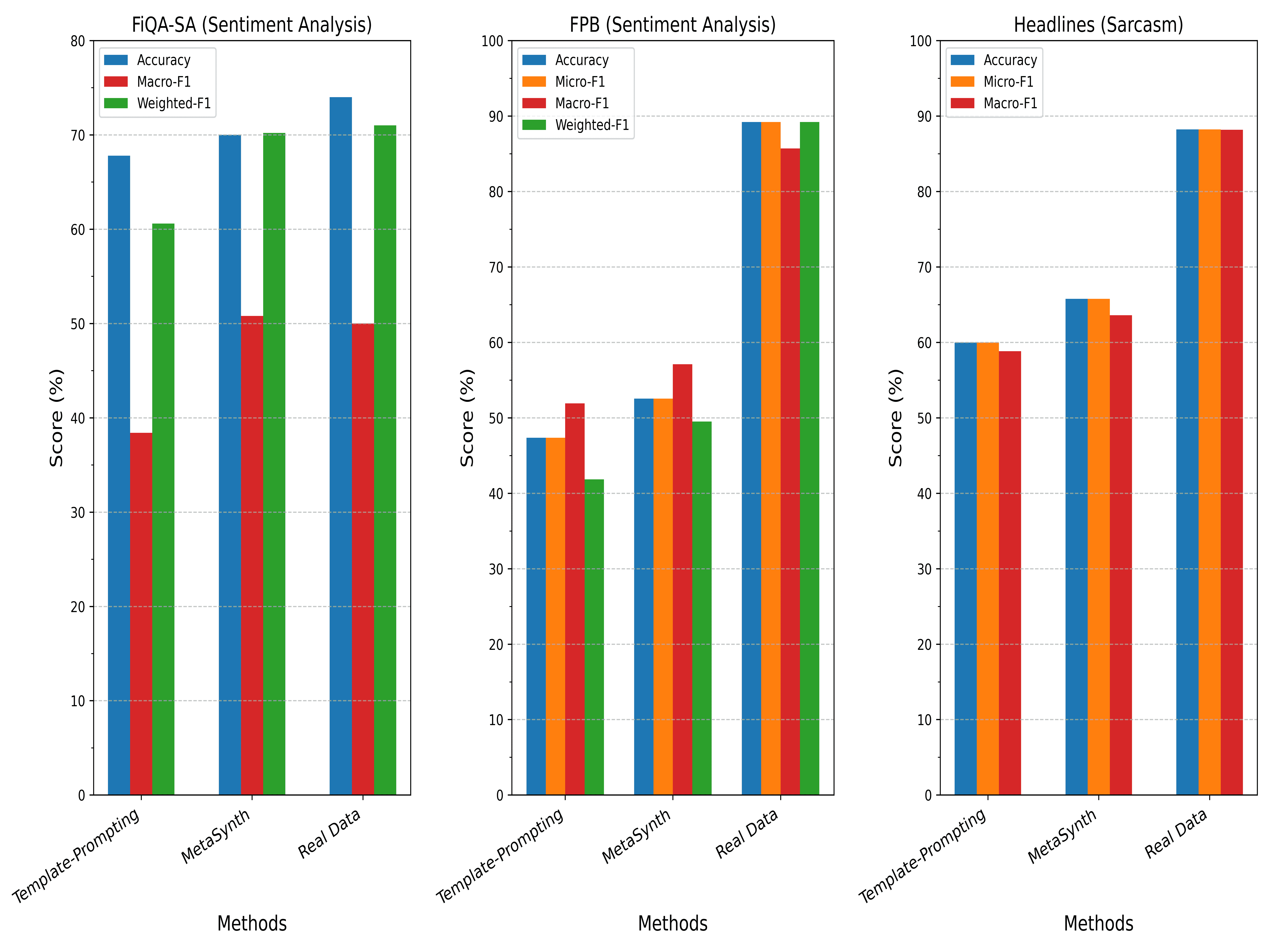}
    \caption{\footnotesize Comparing the performance of BERT finetuned on data synthesized with template-prompting and MetaSynth versus real data on: \textbf{(Left)} FiQA-SA; \textbf{(Middle)} FPB; \textbf{(Right)} Headlines.}
    \label{fig:encoder-perf}
\end{figure}


\begin{table}[h]
    \centering
    \footnotesize 
    \renewcommand{\arraystretch}{1.1} 
    \setlength{\tabcolsep}{3pt} 
    \begin{adjustbox}{max width=\linewidth} 
    \begin{tabularx}{\linewidth}{lXXXXX}
        \toprule
        \textbf{Dataset} & \textbf{EM-1} & \textbf{EM-2} & \textbf{EM-3} & \textbf{EM-5} & \textbf{EM-10} \\
        \midrule
        \multicolumn{6}{l}{\textbf{Finance Datasets}} \\
        \midrule
        ConvFinQA & 0.9784 & 0.7756 & 0.2603 & 0.0310 & 0.0000 \\
        NER       & 0.9923 & 0.7416 & 0.2431 & 0.0287 & 0.0000 \\
        FPB       & 0.9681 & 0.7024 & 0.3137 & 0.0222 & 0.0000 \\
        Headline  & 0.9957 & 0.6752 & 0.1727 & 0.0075 & 0.0000 \\
        FiQA\_SA  & 0.9619 & 0.5745 & 0.1852 & 0.0069 & 0.0000 \\
        \midrule
        \multicolumn{6}{l}{\textbf{Biomedicine Datasets}} \\
        \midrule
        ChemProt  & 0.9329 & 0.5933 & 0.2298 & 0.0111 & 0.0000 \\
        MQP       & 0.9893 & 0.8411 & 0.4211 & 0.0279 & 0.0000 \\
        PubMedQA  & 0.9867 & 0.7431 & 0.3214 & 0.0257 & 0.0000 \\
        RCT       & 0.9886 & 0.7726 & 0.4108 & 0.0422 & 0.0000 \\
        USMLE     & 0.9951 & 0.8137 & 0.4190 & 0.0495 & 0.0000 \\
        \bottomrule
    \end{tabularx}
    \end{adjustbox}
    \caption{\footnotesize Data contamination check results. \textbf{EM-N} stands for Exact Match \textit{N}-gram overlap as a substring between the reference texts from each benchmark dataset and potentially contaminated target texts from Common Crawl (Real Docs).}
    \label{tab:data_contamination}
\end{table}

\section{Experiments and Results}
\label{metasynth-domain-adaptation-results}
\textbf{Domain Adaptation}
Focusing on \textbf{continual pre-training} (where the loss is computed on all tokens) \textbf{and not supervised instruction fine-tuning} (where the loss is computed only on the response conditioned on the prompt) - we continue to train Mistral-7B-v0.3 \cite{jiang2023mistral7b}.
As shown in Table \ref{tab:mistral-7B-cpt}, 25 million tokens of diverse data synthesized with MetaSynth is sufficient for domain adaptation, tested across nine different combinations of mixing Common Crawl texts with synthetic documents and instructions in \texttt{1:1} and \texttt{1:2} token mixing ratios (refer to appendix \ref{sec:domain-datasets} for prompt settings). In Finance, we observe that 25M MetaSynth-generated tokens—without real Common Crawl data—improves the base model by \textbf{4.08\%} on average, outperforming it on all datasets except NER\footnote{This aligns with \newcite{cheng-etal-2024-instruction}, who note NER's low benchmark quality, where the base model achieves the highest score}.  A \texttt{1:1} mix of real and MetaSynth-generated documents also outperforms the same mix with template-prompted data by \textbf{3.08\%}. The same holds true for Biomedicine —  Continual pretraining on 25M MetaSynth-generated tokens—without real Common Crawl data—also boosts the base model by \textbf{13.75\%} on average. Similar to finance finance, a \texttt{1:1} real–synthetic document mix outperforms the same mix with template-prompted data by \textbf{8.85\%}. Overall, in-domain gains are more pronounced in biomedicine, with more types of token mixing ratios improving the base model compared to finance, likely due to more specialized terminology and obscure knowledge required for biomedicine, which the base model lacks. 
\textbf{General Evaluation}
As shown in Table \ref{tab:general-eval}, on average, continual pre-training on \textbf{MetaSynth generated synthetic data does not compromise the generalizability of the LLM}. 
Even when comparing models that underwent CPT exclusively on 25M tokens of MetaSynth data, without any real data incorporated in the training mixture, we observe that the degradation on general benchmarks (e.g., MMLU) is minimal. 









\section{Analysis}

\label{encoder-tasks}
To evaluate the utility of our instruction synthesizer (MetaSynth-\textit{Instruct}) in creating instructions for more general tasks, we conduct the following analyses: 

\textbf{Creating Data For Fine-tuning Encoders}
We adapt our instruction synthesizer to generate synthetic data that emulates datasets used in encoder LM evaluation. We modify the \texttt{task description} in the meta-prompt (Appendix \ref{encoder-task-description}) to instruct the meta-LM to generate synthetic training instances resembling each of three finance datasets—Headline News (sarcasm detection), FiQA-SA (aspect-based sentiment analysis), and Financial Phrasebank (sentiment analysis)—selected for their simplicity and prior use in \newcite{li2023syntheticdatagenerationlarge}'s work. For each dataset, we generate a small set of synthetic instances (refer to table \ref{tab:encoder-dataset-stats}, appendix \ref{sec:encoder-dataset-stats}) with both MetaSynth and template-prompting using 3-shot examples. Fine-tuning a BERT-based classifier on the generated data and evaluating it on a real test partition of each dataset shows that models trained on MetaSynth data outperform those trained on data synthesized by template prompting but remain behind those fine-tuned on real data, consistent with \newcite{li2023syntheticdatagenerationlarge}'s findings (Figure~\ref{fig:encoder-perf}). 

\textbf{Instruction-Response Quality}
We also analyze our synthesized instruction-response pairs in terms of \textit{context relevance}, \textit{response
accuracy}, \textit{task diversity} and \textit{win rate}. Evaluating 1000 sampled instruction-response pairs from each domain and using Claude 3 Opus \cite{anthropic2024claude3} as a judge. Table \ref{tab:response_metrics}, Appendix \ref{sec:appendixE} shows that our synthesized instruction-response pairs for finance exhibit greater task diversity and slightly higher relevance and accuracy scores than Biomedicine, yet 25M biomedical tokens still yield greater improvements to the base model,
suggesting that achieving comparable gains in finance would require substantially more data than what we synthesized due to it being a more generic domain. For both domains, a Mistral-7B-v0.3 model continually pre-trained on 25M MetaSynth tokens also attains higher win rates against Claude 3 Sonnet relative to the base model (Figure \ref{fig:winrates}, appendix \ref{sec:appendixE}). Appendix \ref{sec:metasynth-vs-ipt} shows instructions synthesized with MetaSynth can still exhibit lower diversity compared to \textit{Instruction-Pretraining}. This can be attributed to Instruction-Pretraining using a synthesizer fine-tuned on 1B tokens of real corpora to generate instructions, whereas MetaSynth first generates 25M tokens of synthetic documents, then uses these to evolve instructions.

\textbf{Is It Data Contamination?}
We assess cross-contamination between Common Crawl (\textit{Real Docs}) and domain-specific benchmarks e.g., ConvFinQA using a 10-gram substring match method \cite{benallal2024cosmopedia, openai2024gpt4technicalreport}, deeming an example contaminated if a substring of the example appears in \textit{Real Docs}. Table~\ref{tab:data_contamination} shows there is no contamination between our selected Common Crawl seeds and evaluation datasets.

\textbf{On Knowledge Distillation}
We examine whether our approach constitutes knowledge distillation from Claude 3 Sonnet to Mistral-7B. While some knowledge transfer may occur, performance improvements cannot be attributed solely to knowledge distillation. Our template prompting baseline (Section \ref{template-prompting}) also uses Claude 3 Sonnet but shows no improvement—even regression—while MetaSynth demonstrates significant gains (Table \ref{tab:mistral-7B-cpt}, rows 3 and 5). MetaSynth improves the base model's performance in Finance (63.40→65.18 vs. 63.23 with template-prompting) and Biomedicine (52.94→54.95 vs. 50.48 with template-prompting).

\textbf{Which Components Enhance Data Diversity?}
Our framework uses a meta-model with reasoning and memory to iteratively generate diverse data instances while maintaining topical relevance. We hypothesize that three components contribute most to enhancing data diversity in MetaSynth: (1) The ``Content Analyst Expert'' ensures each instance differs from prior generations (Section \ref{agentic-scaffolding}); (2) The memory mechanism enables comparison between new and existing instances (Section \ref{agentic-scaffolding}); (3) \textit{Topic-aware} seed selection which selects new seeds that maximize topical variation without drifting from initial seeds (Section \ref{document-gen-seed-document-approach}).

\section{Related Work}
Previous work on generating synthetic data with LLMs has primarily focused on post-training data synthesis, particularly for conversational data or instructions \cite{honovich2022unnaturalinstructionstuninglanguage, xu2023wizardlmempoweringlargelanguage, chen2024genqageneratingmillionsinstructions, ding2023enhancing, arannil-etal-2023-adeqa} inter alia.

Recent work like \textit{AgentInstruct} \cite{mitra2024agentinstructgenerativeteachingagentic} uses predefined taxonomies and agentic workflows to generate diverse instruction-response pairs from real corpora. While similar to our approach in using iterative refinement, our method differs by leveraging the meta-model's reasoning to dynamically select synthesis flows rather than sampling from fixed taxonomies. Unlike post-training approaches that compute loss only on responses, our method aligns with pre-training approaches which compute the loss on both prompts and responses. 
We focus on increasing model domain knowledge by targeting continual pre-training (CPT) over supervised fine-tuning (SFT) based on observations that knowledge is primarily learned during pre-training, while SFT only improves in-distribution task performance and can cause forgetting of unused abilities and domain knowledge \cite{zhou2023limaalignment, sun2025amurocharanalyzingrelationship, ke2025demystifyingdomainadaptiveposttrainingfinancial}.

Other agentic approaches for synthetic data generation include UniGen \cite{wu2024unigenunifiedframeworktextual}, which generates SFT data in the format of a target dataset (unlike our method which does not require a target dataset), MATRIX \cite{tang2025synthesizingposttrainingdatallms}, which focuses on synthesizing preference-tuning data for instruction-following, not pre-training, and PersonaHub \cite{ge2024scalingsyntheticdatacreation}, which samples 1B personas from $10^{14}$ tokens of web text and then uses a template prompt e.g., ``create data with persona'' to synthesize instances (given the large scale of real data used in PersonaHub, it is not comparable to our method). We note that non-agentic prompting techniques for extracting diverse data from LLMs such as \newcite{hayati2024farextractdiverseperspectives} and \newcite{wong2024simplestratdiversifyinglanguagemodel}, and human-in-the-loop methods such as \newcite{Chung_2023} are complementary to our approach.

In this work we demonstrate that it is possible to generate useful synthetic data while using real data resource-efficiently: we utilize $\sim$26.5M tokens of real data from domain-specific Common Crawl splits (approximately 50K documents) to synthesize 25M tokens of synthetic data ($\sim$47K documents). This is substantially lower than prior work such as AdaptLLM \cite{cheng2024adaptinglargelanguagemodels} which used billions of tokens of real corpora (5.4B medical, 1.2B finance) for synthetic data generation. Additionally, our method is unsupervised, generating instructions from synthetic texts, whereas Instruction-Pretraining \cite{cheng-etal-2024-instruction} leverages an instruction synthesizer trained on at least 1B tokens of real corpora. 

\section{Conclusion}
We propose \textsc{MetaSynth}, a method that leverages meta-prompting and agentic scaffolding to generate measurably diverse documents and instructions. We demonstrate its efficacy by synthesizing diverse data and then continually
pre-training Mistral-7B on it, yielding significant improvements in two domains, without degrading the model on general tasks.

\section*{Limitations}
Our work has several limitations worth noting. Primarily, our approach of iteratively refining each synthetic instance to be more diverse, while maintaining a record of all previously generated instances incurs a significant inference cost when synthesizing a large collection of documents. 
All data generation was conducted via the Amazon Bedrock API\footnote{\url{https://docs.aws.amazon.com/bedrock/latest/APIReference/welcome.html}}. Our runtime analysis indicates that when making API calls using a single CPU thread, generating one document requires approximately 3.6 minutes (or 3 hours to produce 50 documents from initial seeds). 
To address this constraint, we implemented parallel processing with 64 concurrent CPU threads, with each thread independently generating up to 50 documents from its own seed set. This parallelization allowed us to achieve a throughput of approximately 3,200 documents ($\sim$1.7M tokens) in 3 hours. 
\\
While the inference-time trade off here is deliberate (with the objective of increasing the diversity of generated data), and while our method demonstrates near-linear scaling with CPU thread count (up to a certain rate-limit), it is still an important consideration as we assume the availability of a computing resource with substantial multi-threading capabilities, which may not always be the case when operating in resource-constrained settings; for e.g., template-prompting based methods can synthesize a document or instruction on a single thread in just a few seconds. However, we also wish to emphasize that our approach becomes increasingly viable as LLM inference costs exhibit an approximate 10x year-over-year reduction through hardware improvements and algorithmic innovations such as speculative decoding \cite{leviathan2023fastinferencetransformersspeculative}. \\
A significant challenge also lies in the stability of our agentic workflow. We observe that our procedure is prone to breakdowns, requiring many iterations to be discarded. This instability suggests that more robust methods for maintaining coherent meta-level control may be needed for deploying our approach practically. \\
In this work, we limit our synthesized documents to average 400 words ($\sim$530 tokens) in length. Generating substantially longer documents would likely present challenges in both maintaining semantic diversity across extended text spans and managing inference costs. \\
Furthermore, we find that automatic evaluation metrics for assessing data diversity may not always align well with human judgments. A concrete example of this emerges in our finance domain experiments, where we observe strong biases in the generated data towards specific topics like \texttt{``ESG''}, \texttt{``DeFi''}, and \texttt{``cryptocurrency''}. These biases likely stem from the underlying LLM--Claude 3 Sonnet's-- post-training alignment. This highlights a broader challenge with synthetic data generation methods: ensuring that the generated data not only appears diverse by automated metrics but also maintains domain-appropriate distributions and high diversity by human standards. We leave a human evaluation of the diversity of data synthesized by MetaSynth as future work.

\bibliography{custom}

\begin{thebibliography}{63}
\providecommand{\natexlab}[1]{#1}

\bibitem[{Achille et~al.(2019)Achille, Lam, Tewari, Ravichandran, Maji, Fowlkes, Soatto, and Perona}]{achille2019task2vectaskembeddingmetalearning}
Alessandro Achille, Michael Lam, Rahul Tewari, Avinash Ravichandran, Subhransu Maji, Charless Fowlkes, Stefano Soatto, and Pietro Perona. 2019.
\newblock \href {https://arxiv.org/abs/1902.03545} {Task2vec: Task embedding for meta-learning}.
\newblock \emph{Preprint}, arXiv:1902.03545.

\bibitem[{Anthropic(2024)}]{anthropic2024claude3}
Anthropic. 2024.
\newblock \href {https://www.anthropic.com/news/claude-3-family} {Introducing the next generation of claude}.
\newblock Accessed: 2025-01-25.

\bibitem[{Arannil et~al.(2023)Arannil, Deb, and Roy}]{arannil-etal-2023-adeqa}
Vinayak Arannil, Tomal Deb, and Atanu Roy. 2023.
\newblock \href {https://doi.org/10.18653/v1/2023.bionlp-1.17} {{ADEQA}: A question answer based approach for joint {ADE}-suspect extraction using sequence-to-sequence transformers}.
\newblock In \emph{The 22nd Workshop on Biomedical Natural Language Processing and BioNLP Shared Tasks}, pages 206--214, Toronto, Canada. Association for Computational Linguistics.

\bibitem[{Arannil et~al.(2024)Arannil, Narwal, Bhabesh, Thirandas, Wang, Horwood, Chirayath, and Pandeshwar}]{arannil2024dopaminedomainspecificpretrainingadaptation}
Vinayak Arannil, Neha Narwal, Sourav~Sanjukta Bhabesh, Sai~Nikhil Thirandas, Darren Yow-Bang Wang, Graham Horwood, Alex~Anto Chirayath, and Gouri Pandeshwar. 2024.
\newblock \href {https://arxiv.org/abs/2410.00260} {Dopamine: Domain-specific pre-training adaptation from seed-guided data mining}.
\newblock \emph{Preprint}, arXiv:2410.00260.

\bibitem[{Ben~Allal et~al.(2024)Ben~Allal, Lozhkov, Penedo, Wolf, and von Werra}]{benallal2024cosmopedia}
Loubna Ben~Allal, Anton Lozhkov, Guilherme Penedo, Thomas Wolf, and Leandro von Werra. 2024.
\newblock \href {https://huggingface.co/datasets/HuggingFaceTB/cosmopedia} {Cosmopedia}.

\bibitem[{Chen et~al.(2024)Chen, Qadri, Wen, Jain, Kirchenbauer, Zhou, and Goldstein}]{chen2024genqageneratingmillionsinstructions}
Jiuhai Chen, Rifaa Qadri, Yuxin Wen, Neel Jain, John Kirchenbauer, Tianyi Zhou, and Tom Goldstein. 2024.
\newblock \href {https://arxiv.org/abs/2406.10323} {Genqa: Generating millions of instructions from a handful of prompts}.
\newblock \emph{Preprint}, arXiv:2406.10323.

\bibitem[{Chen et~al.(2022)Chen, Li, Smiley, Ma, Shah, and Wang}]{chen2022convfinqaexploringchainnumerical}
Zhiyu Chen, Shiyang Li, Charese Smiley, Zhiqiang Ma, Sameena Shah, and William~Yang Wang. 2022.
\newblock \href {https://arxiv.org/abs/2210.03849} {Convfinqa: Exploring the chain of numerical reasoning in conversational finance question answering}.
\newblock \emph{Preprint}, arXiv:2210.03849.

\bibitem[{Cheng et~al.(2024{\natexlab{a}})Cheng, Gu, Huang, Bi, Huang, and Wei}]{cheng-etal-2024-instruction}
Daixuan Cheng, Yuxian Gu, Shaohan Huang, Junyu Bi, Minlie Huang, and Furu Wei. 2024{\natexlab{a}}.
\newblock \href {https://doi.org/10.18653/v1/2024.emnlp-main.148} {Instruction pre-training: Language models are supervised multitask learners}.
\newblock In \emph{Proceedings of the 2024 Conference on Empirical Methods in Natural Language Processing}, pages 2529--2550, Miami, Florida, USA. Association for Computational Linguistics.

\bibitem[{Cheng et~al.(2024{\natexlab{b}})Cheng, Huang, and Wei}]{cheng2024adaptinglargelanguagemodels}
Daixuan Cheng, Shaohan Huang, and Furu Wei. 2024{\natexlab{b}}.
\newblock \href {https://arxiv.org/abs/2309.09530} {Adapting large language models to domains via reading comprehension}.
\newblock \emph{Preprint}, arXiv:2309.09530.

\bibitem[{Chung et~al.(2023)Chung, Kamar, and Amershi}]{Chung_2023}
John Chung, Ece Kamar, and Saleema Amershi. 2023.
\newblock \href {https://doi.org/10.18653/v1/2023.acl-long.34} {Increasing diversity while maintaining accuracy: Text data generation with large language models and human interventions}.
\newblock In \emph{Proceedings of the 61st Annual Meeting of the Association for Computational Linguistics (Volume 1: Long Papers)}. Association for Computational Linguistics.

\bibitem[{Cox et~al.(2021)Cox, Wang, Abdul, von~der Weth, and Y.~Lim}]{10.1145/3411764.3445782}
Samuel~Rhys Cox, Yunlong Wang, Ashraf Abdul, Christian von~der Weth, and Brian Y.~Lim. 2021.
\newblock \href {https://doi.org/10.1145/3411764.3445782} {Directed diversity: Leveraging language embedding distances for collective creativity in crowd ideation}.
\newblock In \emph{Proceedings of the 2021 CHI Conference on Human Factors in Computing Systems}, CHI '21, New York, NY, USA. Association for Computing Machinery.

\bibitem[{Dernoncourt and Lee(2017)}]{dernoncourt-lee-2017-pubmed}
Franck Dernoncourt and Ji~Young Lee. 2017.
\newblock \href {https://aclanthology.org/I17-2052/} {{P}ub{M}ed 200k {RCT}: a dataset for sequential sentence classification in medical abstracts}.
\newblock In \emph{Proceedings of the Eighth International Joint Conference on Natural Language Processing (Volume 2: Short Papers)}, pages 308--313, Taipei, Taiwan. Asian Federation of Natural Language Processing.

\bibitem[{Devlin et~al.(2019)Devlin, Chang, Lee, and Toutanova}]{devlin2019bertpretrainingdeepbidirectional}
Jacob Devlin, Ming-Wei Chang, Kenton Lee, and Kristina Toutanova. 2019.
\newblock \href {https://arxiv.org/abs/1810.04805} {Bert: Pre-training of deep bidirectional transformers for language understanding}.
\newblock \emph{Preprint}, arXiv:1810.04805.

\bibitem[{Ding et~al.(2023)Ding, Chen, Xu, Qin, Zheng, Hu, Liu, Sun, and Zhou}]{ding2023enhancing}
Ning Ding, Yulin Chen, Bokai Xu, Yujia Qin, Zhi Zheng, Shengding Hu, Zhiyuan Liu, Maosong Sun, and Bowen Zhou. 2023.
\newblock Enhancing chat language models by scaling high-quality instructional conversations.
\newblock \emph{arXiv preprint arXiv:2305.14233}.

\bibitem[{Eldan and Li(2023)}]{eldan2023tinystoriessmalllanguagemodels}
Ronen Eldan and Yuanzhi Li. 2023.
\newblock \href {https://arxiv.org/abs/2305.07759} {Tinystories: How small can language models be and still speak coherent english?}
\newblock \emph{Preprint}, arXiv:2305.07759.

\bibitem[{Ge et~al.(2024)Ge, Chan, Wang, Yu, Mi, and Yu}]{ge2024scalingsyntheticdatacreation}
Tao Ge, Xin Chan, Xiaoyang Wang, Dian Yu, Haitao Mi, and Dong Yu. 2024.
\newblock \href {https://arxiv.org/abs/2406.20094} {Scaling synthetic data creation with 1,000,000,000 personas}.
\newblock \emph{Preprint}, arXiv:2406.20094.

\bibitem[{Gerstgrasser et~al.(2024)Gerstgrasser, Schaeffer, Dey, Rafailov, Sleight, Hughes, Korbak, Agrawal, Pai, Gromov, Roberts, Yang, Donoho, and Koyejo}]{gerstgrasser2024modelcollapseinevitablebreaking}
Matthias Gerstgrasser, Rylan Schaeffer, Apratim Dey, Rafael Rafailov, Henry Sleight, John Hughes, Tomasz Korbak, Rajashree Agrawal, Dhruv Pai, Andrey Gromov, Daniel~A. Roberts, Diyi Yang, David~L. Donoho, and Sanmi Koyejo. 2024.
\newblock \href {https://arxiv.org/abs/2404.01413} {Is model collapse inevitable? breaking the curse of recursion by accumulating real and synthetic data}.
\newblock \emph{Preprint}, arXiv:2404.01413.

\bibitem[{Grattafiori et~al.(2024)}]{grattafiori2024llama3herdmodels}
Aaron Grattafiori et~al. 2024.
\newblock \href {https://arxiv.org/abs/2407.21783} {The llama 3 herd of models}.

\bibitem[{Guo et~al.(2024)Guo, Chen, Wang, Chang, Pei, Chawla, Wiest, and Zhang}]{ijcai2024p890}
Taicheng Guo, Xiuying Chen, Yaqi Wang, Ruidi Chang, Shichao Pei, Nitesh~V. Chawla, Olaf Wiest, and Xiangliang Zhang. 2024.
\newblock \href {https://doi.org/10.24963/ijcai.2024/890} {Large language model based multi-agents: A survey of progress and challenges}.
\newblock In \emph{Proceedings of the Thirty-Third International Joint Conference on Artificial Intelligence, {IJCAI-24}}, pages 8048--8057. International Joint Conferences on Artificial Intelligence Organization.
\newblock Survey Track.

\bibitem[{Hayati et~al.(2024)Hayati, Lee, Rajagopal, and Kang}]{hayati2024farextractdiverseperspectives}
Shirley~Anugrah Hayati, Minhwa Lee, Dheeraj Rajagopal, and Dongyeop Kang. 2024.
\newblock \href {https://arxiv.org/abs/2311.09799} {How far can we extract diverse perspectives from large language models?}
\newblock \emph{Preprint}, arXiv:2311.09799.

\bibitem[{Hoffmann et~al.(2022)Hoffmann, Borgeaud, Mensch, Buchatskaya, Cai, Rutherford, de~Las~Casas, Hendricks, Welbl, Clark, Hennigan, Noland, Millican, van~den Driessche, Damoc, Guy, Osindero, Simonyan, Elsen, Rae, Vinyals, and Sifre}]{hoffmann2022trainingcomputeoptimallargelanguage}
Jordan Hoffmann, Sebastian Borgeaud, Arthur Mensch, Elena Buchatskaya, Trevor Cai, Eliza Rutherford, Diego de~Las~Casas, Lisa~Anne Hendricks, Johannes Welbl, Aidan Clark, Tom Hennigan, Eric Noland, Katie Millican, George van~den Driessche, Bogdan Damoc, Aurelia Guy, Simon Osindero, Karen Simonyan, Erich Elsen, Jack~W. Rae, Oriol Vinyals, and Laurent Sifre. 2022.
\newblock \href {https://arxiv.org/abs/2203.15556} {Training compute-optimal large language models}.
\newblock \emph{Preprint}, arXiv:2203.15556.

\bibitem[{Honovich et~al.(2022)Honovich, Scialom, Levy, and Schick}]{honovich2022unnaturalinstructionstuninglanguage}
Or~Honovich, Thomas Scialom, Omer Levy, and Timo Schick. 2022.
\newblock \href {https://arxiv.org/abs/2212.09689} {Unnatural instructions: Tuning language models with (almost) no human labor}.
\newblock \emph{Preprint}, arXiv:2212.09689.

\bibitem[{Jiang et~al.(2023)Jiang, Sablayrolles, Mensch, Bamford, Chaplot, de~las Casas, Bressand, Lengyel, Lample, Saulnier, Lavaud, Lachaux, Stock, Scao, Lavril, Wang, Lacroix, and Sayed}]{jiang2023mistral7b}
Albert~Q. Jiang, Alexandre Sablayrolles, Arthur Mensch, Chris Bamford, Devendra~Singh Chaplot, Diego de~las Casas, Florian Bressand, Gianna Lengyel, Guillaume Lample, Lucile Saulnier, Lélio~Renard Lavaud, Marie-Anne Lachaux, Pierre Stock, Teven~Le Scao, Thibaut Lavril, Thomas Wang, Timothée Lacroix, and William~El Sayed. 2023.
\newblock \href {https://arxiv.org/abs/2310.06825} {Mistral 7b}.
\newblock \emph{Preprint}, arXiv:2310.06825.

\bibitem[{Jin et~al.(2020)Jin, Pan, Oufattole, Weng, Fang, and Szolovits}]{jin2020diseasedoespatienthave}
Di~Jin, Eileen Pan, Nassim Oufattole, Wei-Hung Weng, Hanyi Fang, and Peter Szolovits. 2020.
\newblock \href {https://arxiv.org/abs/2009.13081} {What disease does this patient have? a large-scale open domain question answering dataset from medical exams}.
\newblock \emph{Preprint}, arXiv:2009.13081.

\bibitem[{Jin et~al.(2019)Jin, Dhingra, Liu, Cohen, and Lu}]{jin2019pubmedqa}
Qiao Jin, Bhuwan Dhingra, Zhengping Liu, William Cohen, and Xinghua Lu. 2019.
\newblock Pubmedqa: A dataset for biomedical research question answering.
\newblock In \emph{Proceedings of the 2019 Conference on Empirical Methods in Natural Language Processing and the 9th International Joint Conference on Natural Language Processing (EMNLP-IJCNLP)}, pages 2567--2577.

\bibitem[{Ke et~al.(2025)Ke, Ming, Nguyen, Xiong, and Joty}]{ke2025demystifyingdomainadaptiveposttrainingfinancial}
Zixuan Ke, Yifei Ming, Xuan-Phi Nguyen, Caiming Xiong, and Shafiq Joty. 2025.
\newblock \href {https://arxiv.org/abs/2501.04961} {Demystifying domain-adaptive post-training for financial llms}.
\newblock \emph{Preprint}, arXiv:2501.04961.

\bibitem[{Kim et~al.(2024)Kim, Suk, Yue, Viswanathan, Lee, Wang, Gashteovski, Lawrence, Welleck, and Neubig}]{kim2024evaluatinglanguagemodelssynthetic}
Seungone Kim, Juyoung Suk, Xiang Yue, Vijay Viswanathan, Seongyun Lee, Yizhong Wang, Kiril Gashteovski, Carolin Lawrence, Sean Welleck, and Graham Neubig. 2024.
\newblock \href {https://arxiv.org/abs/2412.03679} {Evaluating language models as synthetic data generators}.
\newblock \emph{Preprint}, arXiv:2412.03679.

\bibitem[{Kringelum et~al.(2016)Kringelum, Kj{\ae}rulff, Brunak, Lund, Oprea, and Taboureau}]{Kringelum2016ChemProt30AG}
Jens~Vindahl Kringelum, Sonny~Kim Kj{\ae}rulff, S{\o}ren Brunak, Ole Lund, Tudor~I. Oprea, and Olivier Taboureau. 2016.
\newblock \href {https://api.semanticscholar.org/CorpusID:88650050} {Chemprot-3.0: a global chemical biology diseases mapping}.
\newblock \emph{Database: The Journal of Biological Databases and Curation}, 2016.

\bibitem[{Lee et~al.(2023)Lee, Miranda, and Koyejo}]{lee2023scale}
Alycia Lee, Brando Miranda, and Sanmi Koyejo. 2023.
\newblock Beyond scale: the diversity coefficient as a data quality metric demonstrates llms are pre-trained on formally diverse data.
\newblock \emph{arXiv preprint arXiv:2306.13840}.

\bibitem[{Leviathan et~al.(2023)Leviathan, Kalman, and Matias}]{leviathan2023fastinferencetransformersspeculative}
Yaniv Leviathan, Matan Kalman, and Yossi Matias. 2023.
\newblock \href {https://arxiv.org/abs/2211.17192} {Fast inference from transformers via speculative decoding}.
\newblock \emph{Preprint}, arXiv:2211.17192.

\bibitem[{Li et~al.(2024)Li, Wang, Zhang, and Zhao}]{li-etal-2024-self-prompting}
Junlong Li, Jinyuan Wang, Zhuosheng Zhang, and Hai Zhao. 2024.
\newblock \href {https://doi.org/10.18653/v1/2024.naacl-long.17} {Self-prompting large language models for zero-shot open-domain {QA}}.
\newblock In \emph{Proceedings of the 2024 Conference of the North American Chapter of the Association for Computational Linguistics: Human Language Technologies (Volume 1: Long Papers)}, pages 296--310, Mexico City, Mexico. Association for Computational Linguistics.

\bibitem[{Li et~al.(2023)Li, Zhu, Lu, and Yin}]{li2023syntheticdatagenerationlarge}
Zhuoyan Li, Hangxiao Zhu, Zhuoran Lu, and Ming Yin. 2023.
\newblock \href {https://arxiv.org/abs/2310.07849} {Synthetic data generation with large language models for text classification: Potential and limitations}.
\newblock \emph{Preprint}, arXiv:2310.07849.

\bibitem[{Liu et~al.(2019)Liu, Ott, Goyal, Du, Joshi, Chen, Levy, Lewis, Zettlemoyer, and Stoyanov}]{liu2019robertarobustlyoptimizedbert}
Yinhan Liu, Myle Ott, Naman Goyal, Jingfei Du, Mandar Joshi, Danqi Chen, Omer Levy, Mike Lewis, Luke Zettlemoyer, and Veselin Stoyanov. 2019.
\newblock \href {https://arxiv.org/abs/1907.11692} {Roberta: A robustly optimized bert pretraining approach}.
\newblock \emph{Preprint}, arXiv:1907.11692.

\bibitem[{Maia et~al.(2018)Maia, Handschuh, Freitas, Davis, McDermott, Zarrouk, and Balahur}]{10.1145/3184558.3192301}
Macedo Maia, Siegfried Handschuh, Andr\'{e} Freitas, Brian Davis, Ross McDermott, Manel Zarrouk, and Alexandra Balahur. 2018.
\newblock \href {https://doi.org/10.1145/3184558.3192301} {Www'18 open challenge: Financial opinion mining and question answering}.
\newblock In \emph{Companion Proceedings of the The Web Conference 2018}, WWW '18, page 1941–1942, Republic and Canton of Geneva, CHE. International World Wide Web Conferences Steering Committee.

\bibitem[{Malo et~al.(2013)Malo, Sinha, Takala, Korhonen, and Wallenius}]{malo2013gooddebtbaddebt}
Pekka Malo, Ankur Sinha, Pyry Takala, Pekka Korhonen, and Jyrki Wallenius. 2013.
\newblock \href {https://arxiv.org/abs/1307.5336} {Good debt or bad debt: Detecting semantic orientations in economic texts}.
\newblock \emph{Preprint}, arXiv:1307.5336.

\bibitem[{McCreery et~al.(2020)McCreery, Katariya, Kannan, Chablani, and Amatriain}]{10.1145/3394486.3412861}
Clara~H. McCreery, Namit Katariya, Anitha Kannan, Manish Chablani, and Xavier Amatriain. 2020.
\newblock \href {https://doi.org/10.1145/3394486.3412861} {Effective transfer learning for identifying similar questions: Matching user questions to covid-19 faqs}.
\newblock In \emph{Proceedings of the 26th ACM SIGKDD International Conference on Knowledge Discovery \& Data Mining}, KDD '20, page 3458–3465, New York, NY, USA. Association for Computing Machinery.

\bibitem[{Mitra et~al.(2024)Mitra, Corro, Zheng, Mahajan, Rouhana, Codas, Lu, ge~Chen, Vrousgos, Rosset, Silva, Khanpour, Lara, and Awadallah}]{mitra2024agentinstructgenerativeteachingagentic}
Arindam Mitra, Luciano~Del Corro, Guoqing Zheng, Shweti Mahajan, Dany Rouhana, Andres Codas, Yadong Lu, Wei ge~Chen, Olga Vrousgos, Corby Rosset, Fillipe Silva, Hamed Khanpour, Yash Lara, and Ahmed Awadallah. 2024.
\newblock \href {https://arxiv.org/abs/2407.03502} {Agentinstruct: Toward generative teaching with agentic flows}.
\newblock \emph{Preprint}, arXiv:2407.03502.

\bibitem[{Nayab et~al.(2025)Nayab, Rossolini, Simoni, Saracino, Buttazzo, Manes, and Giacomelli}]{nayab2025concisethoughtsimpactoutput}
Sania Nayab, Giulio Rossolini, Marco Simoni, Andrea Saracino, Giorgio Buttazzo, Nicolamaria Manes, and Fabrizio Giacomelli. 2025.
\newblock \href {https://arxiv.org/abs/2407.19825} {Concise thoughts: Impact of output length on llm reasoning and cost}.
\newblock \emph{Preprint}, arXiv:2407.19825.

\bibitem[{OpenAI et~al.(2024)OpenAI, Achiam, Adler, Agarwal, Ahmad, Akkaya, Aleman, Almeida, Altenschmidt, Altman, Anadkat et~al.}]{openai2024gpt4technicalreport}
OpenAI, Josh Achiam, Steven Adler, Sandhini Agarwal, Lama Ahmad, Ilge Akkaya, Florencia~Leoni Aleman, Diogo Almeida, Janko Altenschmidt, Sam Altman, Shyamal Anadkat, et~al. 2024.
\newblock \href {https://arxiv.org/abs/2303.08774} {Gpt-4 technical report}.
\newblock \emph{arXiv preprint arXiv:2303.08774}.

\bibitem[{Radford et~al.(2019)Radford, Wu, Child, Luan, Amodei, and Sutskever}]{Radford2019LanguageMA}
Alec Radford, Jeff Wu, Rewon Child, David Luan, Dario Amodei, and Ilya Sutskever. 2019.
\newblock \href {https://api.semanticscholar.org/CorpusID:160025533} {Language models are unsupervised multitask learners}.

\bibitem[{Ramos(2003)}]{Ramos2003UsingTT}
Juan~Enrique Ramos. 2003.
\newblock \href {https://api.semanticscholar.org/CorpusID:14638345} {Using tf-idf to determine word relevance in document queries}.

\bibitem[{Salinas~Alvarado et~al.(2015)Salinas~Alvarado, Verspoor, and Baldwin}]{salinas-alvarado-etal-2015-domain}
Julio~Cesar Salinas~Alvarado, Karin Verspoor, and Timothy Baldwin. 2015.
\newblock \href {https://aclanthology.org/U15-1010/} {Domain adaption of named entity recognition to support credit risk assessment}.
\newblock In \emph{Proceedings of the Australasian Language Technology Association Workshop 2015}, pages 84--90, Parramatta, Australia.

\bibitem[{Seddik et~al.(2024)Seddik, Chen, Hayou, Youssef, and Debbah}]{seddik2024badtrainingsyntheticdata}
Mohamed El~Amine Seddik, Suei-Wen Chen, Soufiane Hayou, Pierre Youssef, and Merouane Debbah. 2024.
\newblock \href {https://arxiv.org/abs/2404.05090} {How bad is training on synthetic data? a statistical analysis of language model collapse}.
\newblock \emph{Preprint}, arXiv:2404.05090.

\bibitem[{Shaib et~al.(2024{\natexlab{a}})Shaib, Barrow, Sun, Siu, Wallace, and Nenkova}]{shaib2024standardizingmeasurementtextdiversity}
Chantal Shaib, Joe Barrow, Jiuding Sun, Alexa~F. Siu, Byron~C. Wallace, and Ani Nenkova. 2024{\natexlab{a}}.
\newblock \href {https://arxiv.org/abs/2403.00553} {Standardizing the measurement of text diversity: A tool and a comparative analysis of scores}.
\newblock \emph{Preprint}, arXiv:2403.00553.

\bibitem[{Shaib et~al.(2024{\natexlab{b}})Shaib, Elazar, Li, and Wallace}]{shaib2024detectionmeasurementsyntactictemplates}
Chantal Shaib, Yanai Elazar, Junyi~Jessy Li, and Byron~C. Wallace. 2024{\natexlab{b}}.
\newblock \href {https://arxiv.org/abs/2407.00211} {Detection and measurement of syntactic templates in generated text}.
\newblock \emph{Preprint}, arXiv:2407.00211.

\bibitem[{Shumailov et~al.(2024)Shumailov, Shumaylov, Zhao, Gal, Papernot, and Anderson}]{shumailov2024curserecursiontraininggenerated}
Ilia Shumailov, Zakhar Shumaylov, Yiren Zhao, Yarin Gal, Nicolas Papernot, and Ross Anderson. 2024.
\newblock \href {https://arxiv.org/abs/2305.17493} {The curse of recursion: Training on generated data makes models forget}.
\newblock \emph{Preprint}, arXiv:2305.17493.

\bibitem[{Sinha and Khandait(2020)}]{sinha2020impactnewscommoditymarket}
Ankur Sinha and Tanmay Khandait. 2020.
\newblock \href {https://arxiv.org/abs/2009.04202} {Impact of news on the commodity market: Dataset and results}.
\newblock \emph{Preprint}, arXiv:2009.04202.

\bibitem[{Sun and Dredze(2025)}]{sun2025amurocharanalyzingrelationship}
Kaiser Sun and Mark Dredze. 2025.
\newblock \href {https://arxiv.org/abs/2408.06663} {Amuro and char: Analyzing the relationship between pre-training and fine-tuning of large language models}.
\newblock \emph{Preprint}, arXiv:2408.06663.

\bibitem[{Suzgun and Kalai(2024)}]{suzgun2024metaprompting}
Mirac Suzgun and Adam~Tauman Kalai. 2024.
\newblock \href {https://arxiv.org/abs/2401.12954} {Meta-prompting: Enhancing language models with task-agnostic scaffolding}.

\bibitem[{Tang et~al.(2025)Tang, Pang, Liu, Tang, Ye, Jin, Dong, Wang, and Chen}]{tang2025synthesizingposttrainingdatallms}
Shuo Tang, Xianghe Pang, Zexi Liu, Bohan Tang, Rui Ye, Tian Jin, Xiaowen Dong, Yanfeng Wang, and Siheng Chen. 2025.
\newblock \href {https://arxiv.org/abs/2410.14251} {Synthesizing post-training data for llms through multi-agent simulation}.
\newblock \emph{Preprint}, arXiv:2410.14251.

\bibitem[{Villalobos et~al.(2024)Villalobos, Ho, Sevilla, Besiroglu, Heim, and Hobbhahn}]{villalobos2024run}
Pablo Villalobos, Anson Ho, Jaime Sevilla, Tamay Besiroglu, Lennart Heim, and Marius Hobbhahn. 2024.
\newblock \href {https://arxiv.org/abs/2211.04325} {Will we run out of data? limits of llm scaling based on human-generated data}.
\newblock \emph{Preprint}, arXiv:2211.04325.

\bibitem[{Wan et~al.(2023)Wan, Huang, Yang, Quan, Bi, and Shi}]{wan2023exploreinstructenhancingdomainspecificinstruction}
Fanqi Wan, Xinting Huang, Tao Yang, Xiaojun Quan, Wei Bi, and Shuming Shi. 2023.
\newblock \href {https://arxiv.org/abs/2310.09168} {Explore-instruct: Enhancing domain-specific instruction coverage through active exploration}.
\newblock \emph{Preprint}, arXiv:2310.09168.

\bibitem[{Wang et~al.(2022)Wang, Mishra, Alipoormolabashi, Kordi, Mirzaei, Arunkumar, Ashok, Dhanasekaran, Naik, Stap, Pathak, Karamanolakis, Lai, Purohit, Mondal, Anderson, Kuznia, Doshi, Patel, Pal, Moradshahi, Parmar, Purohit, Varshney, Kaza, Verma, Puri, Karia, Sampat, Doshi, Mishra, Reddy, Patro, Dixit, Shen, Baral, Choi, Smith, Hajishirzi, and Khashabi}]{wang2022supernaturalinstructionsgeneralizationdeclarativeinstructions}
Yizhong Wang, Swaroop Mishra, Pegah Alipoormolabashi, Yeganeh Kordi, Amirreza Mirzaei, Anjana Arunkumar, Arjun Ashok, Arut~Selvan Dhanasekaran, Atharva Naik, David Stap, Eshaan Pathak, Giannis Karamanolakis, Haizhi~Gary Lai, Ishan Purohit, Ishani Mondal, Jacob Anderson, Kirby Kuznia, Krima Doshi, Maitreya Patel, Kuntal~Kumar Pal, Mehrad Moradshahi, Mihir Parmar, Mirali Purohit, Neeraj Varshney, Phani~Rohitha Kaza, Pulkit Verma, Ravsehaj~Singh Puri, Rushang Karia, Shailaja~Keyur Sampat, Savan Doshi, Siddhartha Mishra, Sujan Reddy, Sumanta Patro, Tanay Dixit, Xudong Shen, Chitta Baral, Yejin Choi, Noah~A. Smith, Hannaneh Hajishirzi, and Daniel Khashabi. 2022.
\newblock \href {https://arxiv.org/abs/2204.07705} {Super-naturalinstructions: Generalization via declarative instructions on 1600+ nlp tasks}.
\newblock \emph{Preprint}, arXiv:2204.07705.

\bibitem[{Wei et~al.(2023)Wei, Wang, Schuurmans, Bosma, Ichter, Xia, Chi, Le, and Zhou}]{wei2023chainofthoughtpromptingelicitsreasoning}
Jason Wei, Xuezhi Wang, Dale Schuurmans, Maarten Bosma, Brian Ichter, Fei Xia, Ed~Chi, Quoc Le, and Denny Zhou. 2023.
\newblock \href {https://arxiv.org/abs/2201.11903} {Chain-of-thought prompting elicits reasoning in large language models}.
\newblock \emph{Preprint}, arXiv:2201.11903.

\bibitem[{Wong et~al.(2024)Wong, Orlovskiy, Luo, Seshia, and Gonzalez}]{wong2024simplestratdiversifyinglanguagemodel}
Justin Wong, Yury Orlovskiy, Michael Luo, Sanjit~A. Seshia, and Joseph~E. Gonzalez. 2024.
\newblock \href {https://arxiv.org/abs/2410.09038} {Simplestrat: Diversifying language model generation with stratification}.
\newblock \emph{Preprint}, arXiv:2410.09038.

\bibitem[{Wu et~al.(2023)Wu, Bansal, Zhang, Wu, Li, Zhu, Jiang, Zhang, Zhang, Liu, Awadallah, White, Burger, and Wang}]{wu2023autogenenablingnextgenllm}
Qingyun Wu, Gagan Bansal, Jieyu Zhang, Yiran Wu, Beibin Li, Erkang Zhu, Li~Jiang, Xiaoyun Zhang, Shaokun Zhang, Jiale Liu, Ahmed~Hassan Awadallah, Ryen~W White, Doug Burger, and Chi Wang. 2023.
\newblock \href {https://arxiv.org/abs/2308.08155} {Autogen: Enabling next-gen llm applications via multi-agent conversation}.
\newblock \emph{Preprint}, arXiv:2308.08155.

\bibitem[{Wu et~al.(2024)Wu, Huang, Gao, Chen, Zhang, Wan, Zhou, Zhang, Gao, Xiao, and Sun}]{wu2024unigenunifiedframeworktextual}
Siyuan Wu, Yue Huang, Chujie Gao, Dongping Chen, Qihui Zhang, Yao Wan, Tianyi Zhou, Xiangliang Zhang, Jianfeng Gao, Chaowei Xiao, and Lichao Sun. 2024.
\newblock \href {https://arxiv.org/abs/2406.18966} {Unigen: A unified framework for textual dataset generation using large language models}.
\newblock \emph{Preprint}, arXiv:2406.18966.

\bibitem[{Xu et~al.(2023)Xu, Sun, Zheng, Geng, Zhao, Feng, Tao, and Jiang}]{xu2023wizardlmempoweringlargelanguage}
Can Xu, Qingfeng Sun, Kai Zheng, Xiubo Geng, Pu~Zhao, Jiazhan Feng, Chongyang Tao, and Daxin Jiang. 2023.
\newblock \href {https://arxiv.org/abs/2304.12244} {Wizardlm: Empowering large language models to follow complex instructions}.
\newblock \emph{Preprint}, arXiv:2304.12244.

\bibitem[{Xu et~al.(2025)Xu, Cui, Yu, Kan, Shi, Zhuang, Jin, Ho, and Yang}]{xu2025knowledgeinfusedpromptingassessingadvancing}
Ran Xu, Hejie Cui, Yue Yu, Xuan Kan, Wenqi Shi, Yuchen Zhuang, Wei Jin, Joyce Ho, and Carl Yang. 2025.
\newblock \href {https://arxiv.org/abs/2311.00287} {Knowledge-infused prompting: Assessing and advancing clinical text data generation with large language models}.
\newblock \emph{Preprint}, arXiv:2311.00287.

\bibitem[{Yao et~al.(2023)Yao, Zhao, Yu, Du, Shafran, Narasimhan, and Cao}]{yao2023reactsynergizingreasoningacting}
Shunyu Yao, Jeffrey Zhao, Dian Yu, Nan Du, Izhak Shafran, Karthik Narasimhan, and Yuan Cao. 2023.
\newblock \href {https://arxiv.org/abs/2210.03629} {React: Synergizing reasoning and acting in language models}.
\newblock \emph{Preprint}, arXiv:2210.03629.

\bibitem[{Yu et~al.(2023)Yu, Zhuang, Zhang, Meng, Ratner, Krishna, Shen, and Zhang}]{yu2023largelanguagemodelattributed}
Yue Yu, Yuchen Zhuang, Jieyu Zhang, Yu~Meng, Alexander Ratner, Ranjay Krishna, Jiaming Shen, and Chao Zhang. 2023.
\newblock \href {https://arxiv.org/abs/2306.15895} {Large language model as attributed training data generator: A tale of diversity and bias}.
\newblock \emph{Preprint}, arXiv:2306.15895.

\bibitem[{Zhang et~al.(2024)Zhang, Yuan, and Yao}]{zhang2024metapromptingaisystems}
Yifan Zhang, Yang Yuan, and Andrew Chi-Chih Yao. 2024.
\newblock \href {https://arxiv.org/abs/2311.11482} {Meta prompting for ai systems}.
\newblock \emph{Preprint}, arXiv:2311.11482.

\bibitem[{Zhou et~al.(2023)Zhou, Liu, Xu, Iyer, Sun, Mao, Ma, Efrat, Yu, Yu, Zhang, Ghosh, Lewis, Zettlemoyer, and Levy}]{zhou2023limaalignment}
Chunting Zhou, Pengfei Liu, Puxin Xu, Srini Iyer, Jiao Sun, Yuning Mao, Xuezhe Ma, Avia Efrat, Ping Yu, Lili Yu, Susan Zhang, Gargi Ghosh, Mike Lewis, Luke Zettlemoyer, and Omer Levy. 2023.
\newblock \href {https://arxiv.org/abs/2305.11206} {Lima: Less is more for alignment}.
\newblock \emph{Preprint}, arXiv:2305.11206.

\end{thebibliography}

\appendix

\section{Meta-Prompting: Algorithmic Procedure}
\label{sec:appendixA}

\begin{algorithm}
\footnotesize
\caption{Meta Prompting}
\label{alg:meta-prompting-algorithm}
\begin{algorithmic}[1]
\REQUIRE LM : $\mathcal{S} \rightarrow \mathcal{S}$, $x$, $error \in \mathcal{S}$; $T \in \mathbb{N}$; $t_{\text{init}}, t_{\text{mid}}, t_{\text{exp}}, e_{\text{exp}}, e_{\text{ret}} : \mathcal{S} \rightarrow \mathcal{S}$
\STATE $\mathcal{H}_1 \leftarrow t_{\text{init}}(x)$
\FOR{$t \in [1, \ldots, T]$}
    \STATE $y_t \leftarrow LM(\mathcal{H}_t)$
    \IF{$e_{\text{exp}}(y_t) \neq \emptyset$}
        \STATE $prompt \leftarrow t_{\text{exp}}(e_{\text{exp}}(y_t))$
        \STATE $z_t \leftarrow LM(prompt)$
        \STATE $\mathcal{H}_{t+1} \leftarrow \mathcal{H}_t \oplus t_{\text{mid}}(z_t)$ \COMMENT{Meta Model provided expert instructions}
    \ELSIF{$e_{\text{ret}}(y_t) \neq \emptyset$}
        \RETURN $e_{\text{ret}}(y_t)$ \COMMENT{Meta Model returned end of generation token}
    \ELSE
        \STATE $\mathcal{H}_{t+1} \leftarrow \mathcal{H}_t \oplus error$ \COMMENT{Meta Model formatting error}
    \ENDIF
\ENDFOR
\end{algorithmic}
\end{algorithm}

Let $\mathbb{S}$ be the set of finite strings, with $\emptyset$ denoting the empty string. A test-time query $x \in \mathbb{S}$ represents a natural language task. The fixed language model $\mathtt{LM}$ operates from $\mathbb{S}$ to $\mathbb{S}$, taking a prompt history $\mathcal{H}$ as input and producing an output. Template functions $t_\text{init}$, $t_\text{mid}$, and $t_\text{exp}$ map $\mathbb{S}$ to $\mathbb{S}$, formatting input/output for the meta-LM and agent/expert models. String extractors $e_\text{exp}$ and $e_\text{ret}$ retrieve substrings enclosed by delimiters, while $\oplus$ denotes string concatenation, and $\text{error} \in \mathbb{S}$ represents error messages.  At each iteration, $\mathcal{H}_t$ guides $\mathtt{LM}$ to either return a response or consult an agent, with instructions extracted via $e_\text{exp}$. Agents process only what is shared with them by the meta-LM, and their outputs are formatted with $t_\text{mid}$. A final response is extracted using $e_\text{ret}$ and returned. If neither a final response nor a call to an agent is made, $\text{error}$ is appended to $\mathcal{H}_t$ for error handling.

\begin{figure}[t]
    \centering
    \includegraphics[width=\columnwidth, height=0.6\textheight, keepaspectratio]{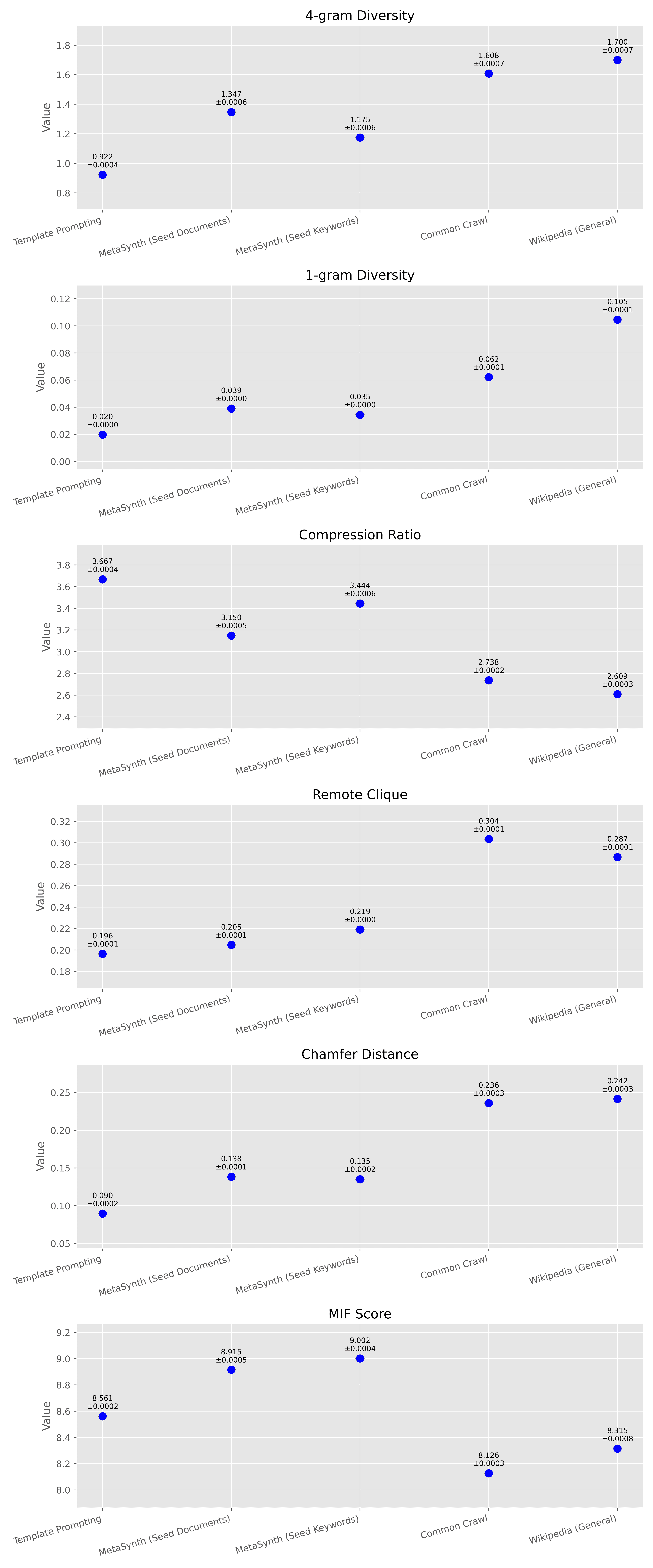}
    \caption{\textbf{Finance Domain: Distribution of diversity metrics for documents synthesized by MetaSynth versus other types of documents (e.g., those generated with template-prompting or real data).}}
    \label{fig:finance_diversity_metrics_metasynth_docs_vs_others}
\end{figure}

\begin{figure}[t]
    \centering
    \includegraphics[width=\columnwidth, height=0.6\textheight, keepaspectratio]{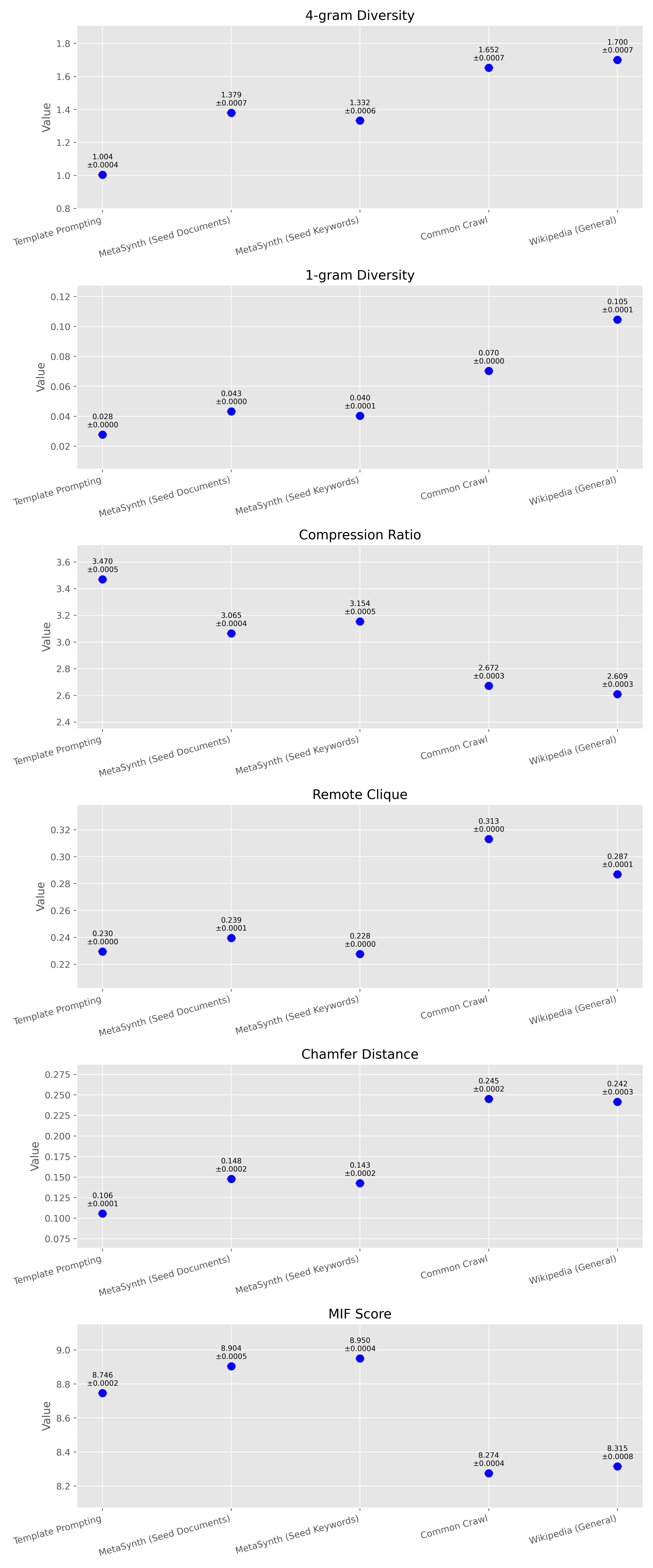}
    \caption{\textbf{Biomedical Domain: Distribution of diversity metrics for documents synthesized by MetaSynth versus other types of documents (e.g., those generated with template-prompting or real data).}}
    \label{fig:biomedicine_diversity_metrics_metasynth_docs_vs_others}
\end{figure}

\section{Prompt Settings \& Datasets For Domain Adaptation Experiments}
\label{sec:domain-datasets}
We follow the prompt setting of AdaptLLM \cite{cheng2024adaptinglargelanguagemodels}: for biomedicine domain, we evaluate zero-shot performance on PubMedQA \cite{jin2019pubmedqa} and USMLE \cite{jin2020diseasedoespatienthave}, few-shot performance on ChemProt \cite{Kringelum2016ChemProt30AG}, MQP \cite{10.1145/3394486.3412861} and RCT \cite{dernoncourt-lee-2017-pubmed}; for finance domain, we evaluate zero-shot performance on ConvFinQA \cite{chen2022convfinqaexploringchainnumerical} and few-shot performance on FPB \cite{malo2013gooddebtbaddebt}, FiQA SA \cite{10.1145/3184558.3192301}, Headline \cite{sinha2020impactnewscommoditymarket}, and NER \cite{salinas-alvarado-etal-2015-domain}.

\begin{figure}[ht]
    \centering
    \includegraphics[width=\columnwidth]{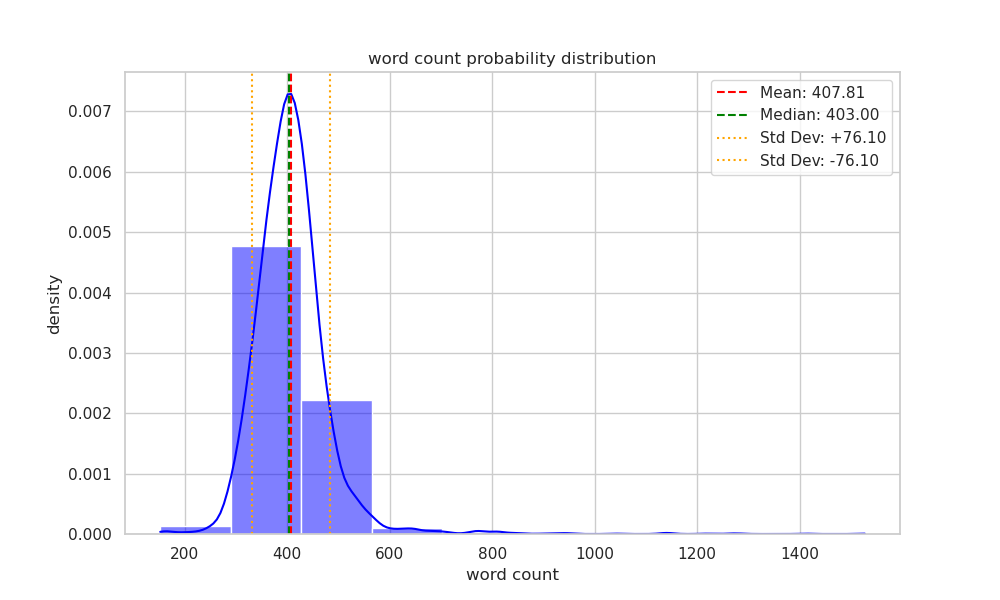}
    \caption{Length distribution (in word count) of documents synthesized by \textit{MetaSynth} from the Finance domain.}
    \label{fig:length1}
\end{figure}

\begin{figure}[ht]
    \centering
    \includegraphics[width=\columnwidth]{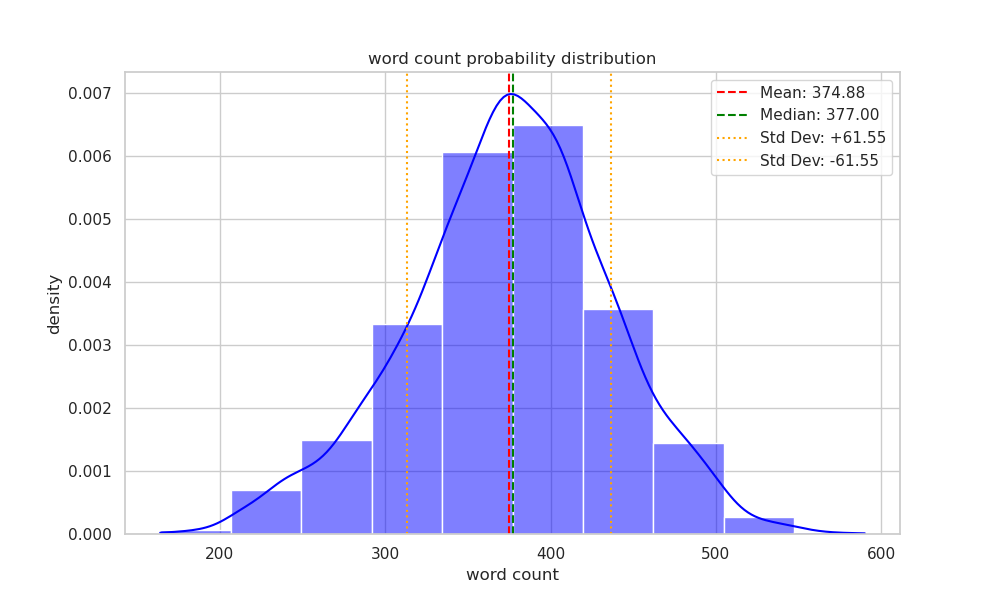}
    \caption{Length distribution (in word count) of documents synthesized by Template-Prompting from the Finance domain.}
    \label{fig:length2}
\end{figure}

\begin{figure}[ht]
    \centering
    \includegraphics[width=\columnwidth]{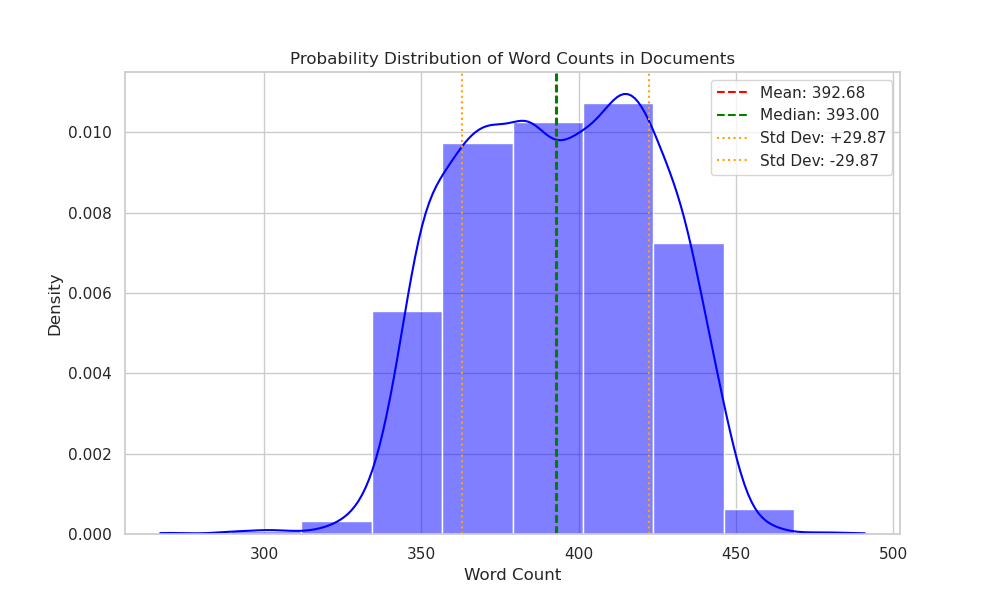}
    \caption{Length distribution (in word count) of Common Crawl documents from the Finance domain.}
    \label{fig:length3}
\end{figure}

\begin{figure}[ht]
    \centering
    \includegraphics[width=\columnwidth]{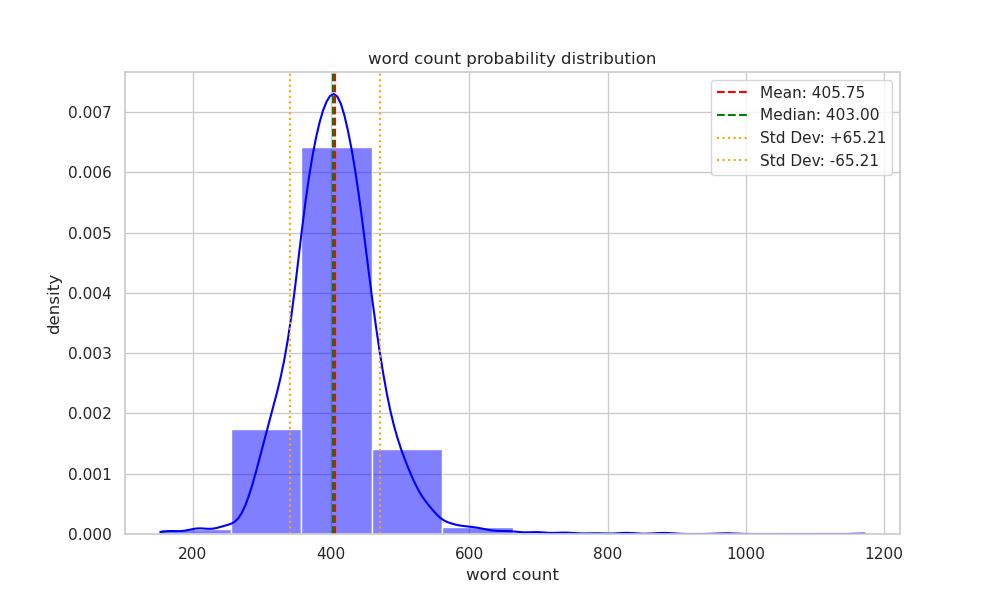}
    \caption{Length distribution (in word count) of documents synthesized by \textit{MetaSynth} from the Biomedicine domain.}
    \label{fig:length4}
\end{figure}

\begin{figure}[ht]
    \centering
    \includegraphics[width=\columnwidth]{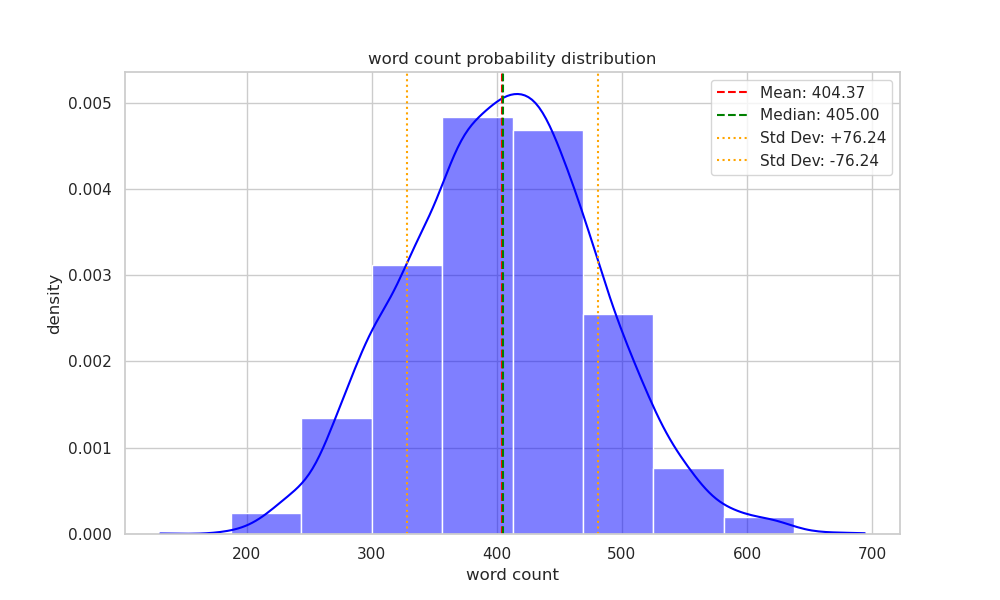}
    \caption{Length distribution (in word count) of documents synthesized by Template-Prompting from the Biomedicine domain.}
    \label{fig:length5}
\end{figure}

\begin{figure}[ht]
    \centering
    \includegraphics[width=\columnwidth]{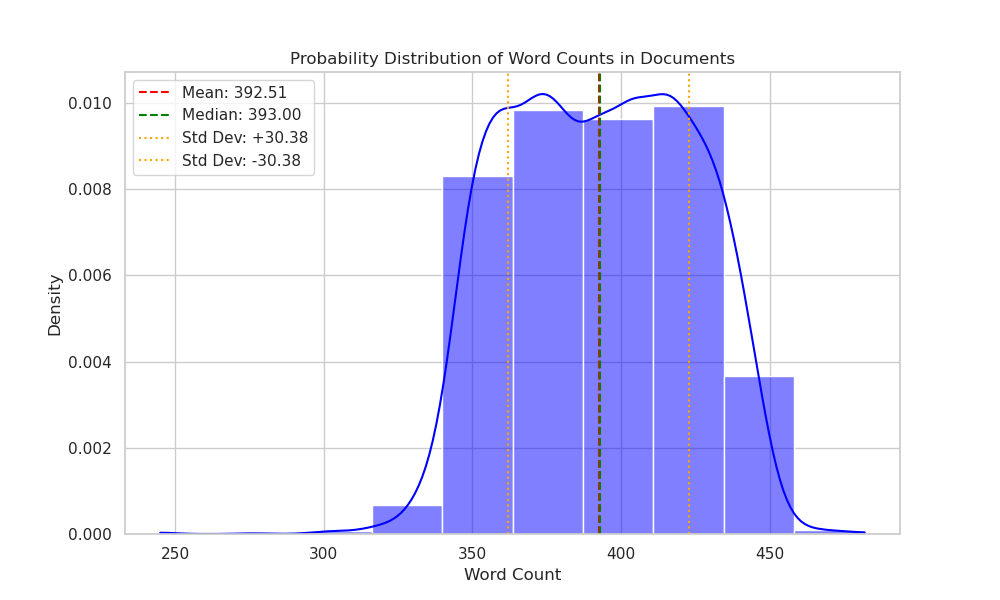}
    \caption{Length distribution (in word count) of Common Crawl documents from the Biomedicine domain.}
    \label{fig:length6}
\end{figure}

\begin{figure}[t]
    \centering
    \includegraphics[width=\columnwidth, height=0.6\textheight, keepaspectratio]{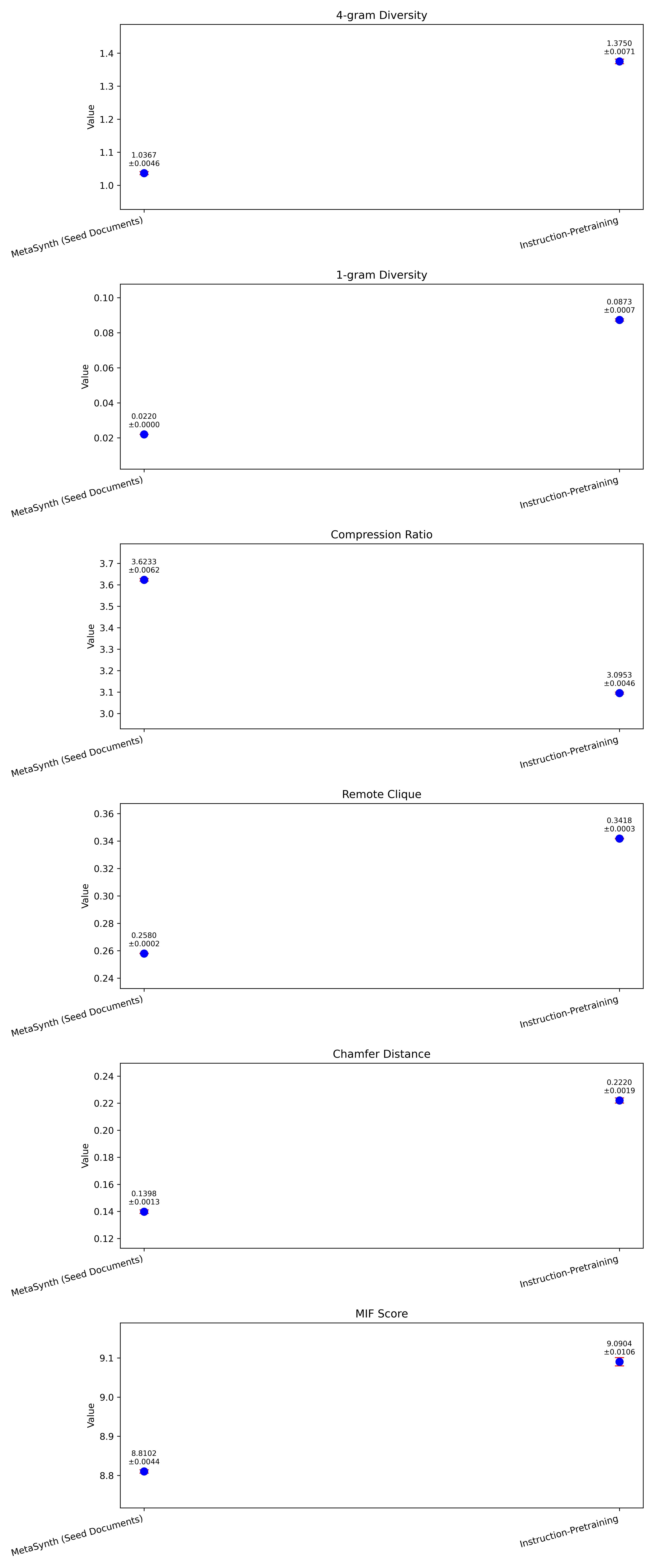}
    \caption{\textbf{Biomedical Domain: Distribution of diversity metrics for instructions synthesized by MetaSynth versus instructions synthesized by Instruction-Pretraining \cite{cheng-etal-2024-instruction}.}}
    \label{fig:biomedicine-metasynth-vs-ipt}
\end{figure}

\section{Diversity Metrics}
\label{sec:diversity-appendix}
\subsection{\textit{Task2Vec} Diversity Coefficient}
The Task2Vec diversity coefficient proposed by \newcite{lee2023scale} quantifies the intrinsic variability of a dataset by measuring the distinctness of different data batches, which can be measured for each batch by computing the diagonal of the Fisher Information Matrix (FIM) using a fixed GPT-2 \cite{Radford2019LanguageMA} probe network. Intuitively, if a dataset is rich in latent concepts, different batches will fine-tune the final layer of the probe network in diverse ways, resulting in Task2Vec embeddings that are more dissimilar (i.e., have larger pairwise cosine distances). Thus, a dataset containing a wide variety of topics and styles should exhibit a higher diversity coefficient than a more homogeneous dataset. We calculate the Task2Vec coefficient as follows:
\begin{itemize}
    \item \textbf{Sampling Batches:} \\
    Sample \(M\) batches from a dataset \(D\) e.g., the corpus of documents synthesized with MetaSynth. Each batch \(B_i\) (for \(i = 1, \ldots, M\))
    consists of \(n\) text sequences:
    \[
    B_i = \{ x_1^{(i)}, x_2^{(i)}, \ldots, x_n^{(i)} \}.
    \]
    
    \item \textbf{Fine-Tuning the Probe Network:} \\
    For each batch \(B_i\), we fine-tune only the final layer of the probe network \(f_w\)
    for next-token prediction, keeping the other layers frozen.
    
    \item \textbf{Computing Gradients:} \\
    For each sequence \(x \in B_i\) and each token position \(t\), we then compute the gradient of the log-likelihood
    with respect to the final-layer parameters:
    \[
    g_t^{(i)} = \nabla_{w} \log \hat{p}_w\Bigl(x_t \mid x_{1:t-1}\Bigr).
    \]
    
    \item \textbf{Estimating the Fisher Information Matrix (FIM):} \\
    For each batch \(B_i\),  FIM is approximated by taking the expected outer product of the gradients:
    \[
    \hat{F}_{B_i} = \mathbb{E}_{(x,t) \sim B_i}\left[ g_t^{(i)} \, \bigl(g_t^{(i)}\bigr)^\top \right].
    \]
    
    \item \textbf{Extracting the Task2Vec Embedding:} \\
    The Task2Vec embedding \(f_{B_i}\) for  each batch \(B_i\) is defined as the diagonal of the FIM:
    \[
    f_{B_i} = \operatorname{diag}\Bigl(\hat{F}_{B_i}\Bigr).
    \]
    
    \item \textbf{Computing Pairwise Cosine Distances:} \\
    For every distinct pair of batches \((B_i, B_j)\) with \(i < j\), we then compute the cosine distance between
    their embeddings:
    \[
    d_{ij} = d\Bigl( f_{B_i}, f_{B_j} \Bigr).
    \]
    
    \item \textbf{Calculating the Diversity Coefficient:} \\
    We then estimate the Task2Vec diversity coefficient as the average pairwise cosine distance across all batches:
    \[
    \hat{div}(D) = \frac{2}{M(M-1)} \sum_{1 \leq i < j \leq M} d_{ij}.
    \]
\end{itemize}

\subsection{Compression Ratio}
\paragraph{Compression Ratio (CR)} Text compression algorithms identify redundancy in variable-length sequences:

\begin{equation}
\textrm{CR}(D) = \frac{\text{size of }D\oplus}{\text{compressed size of }D\oplus}
\end{equation}

where $D\oplus$ denotes the dataset $D$ concatenated into a single string.

\subsection{N-Gram Diversity}
\paragraph{N-Gram Diversity Score (NGD)} NGD extends the idea of token-type ratio (i.e., the unique token
count divided by the total count of tokens) to longer $n$-grams:

\begin{equation}
\textrm{NGD}(D) = \sum_{n=1}^{4} \frac{\#\text{ unique }n\text{-grams in }D\oplus}{\#\text{ }n\text{-grams in }D\oplus}
\end{equation}

where $D\oplus$ denotes the dataset $D$ concatenated into a single string.

\subsection{Remote Clique}

\paragraph{Remote-Clique Distance} Average of mean pairwise distances:

\begin{equation}
\frac{1}{N^2} \sum_{i,j} d(\mathbf{x}_i, \mathbf{x}_j)
\end{equation}

where $\mathbf{x}_i$ represents a document embedding vector computed by a language model.

\subsection{Chamfer Distance}
\paragraph{Chamfer Distance} Average of minimum pairwise distances, also computed over document embeddings:

\begin{equation}
\frac{1}{N} \sum_{i=1}^N \min_{j\neq i} d(\mathbf{x}_i, \mathbf{x}_j)
\end{equation}

\subsection{Mean Inverse Frequency (MIF)}
This metric captures the use of rare vocabulary in synthesized documents. For each word, we calculate its inverse frequency value based on a reference corpus (in this case Wikipedia). We then average these values over all words in the document to produce a document-level score that captures lexical rarity.

\subsection{Diversity Distribution for MetaSynth Documents Vs Real Documents}
\label{sec:boostrap-resampling}
Figures \ref{fig:finance_diversity_metrics_metasynth_docs_vs_others} \& \ref{fig:biomedicine_diversity_metrics_metasynth_docs_vs_others} show diversity metrics for each domain, computed over 5000 synthetic documents with 95\% confidence intervals via 1000 bootstrap resamples. MetaSynth documents generated using common crawl documents as seeds exhibit consistently higher diversity, as measured by these metrics, compared to those generated from seed keywords. In turn, MetaSynth even with seed keywords generates more diverse documents than template-prompting (which uses common crawl documents as in-context exemplars).

\subsection{Diversity Distribution for MetaSynth Instructions Vs Instruction-Pretraining}
\label{sec:metasynth-vs-ipt}
Figure \ref{fig:biomedicine-metasynth-vs-ipt}, 
illustrates the variance for diversity metrics between MetaSynth-\textit{Instruct} and \textit{Instruction-Pretraining}. MetaSynth instructions exhibit lower diversity, as they evolve solely from synthetic documents, whereas Instruction-Pretraining leverages a 1B-token corpus of real data.

\section{Length Distribution: MetaSynth Documents Vs Real Documents}
\label{sec:appendixD}
Figures \ref{fig:length1}, \ref{fig:length2}, \ref{fig:length3}, \ref{fig:length4}, \ref{fig:length5}, \ref{fig:length6} show the length distributions of each type of synthetic or real document used in our work. 

\begin{table}[h]
    \centering
    \small
    \begin{tabular}{lccc}
        \toprule
        & \textbf{Accuracy} & \textbf{Relevance} & \textbf{\# Category} \\
        \midrule
        BioMed.  & 82.0  & 91.0  & 16 \\
        Finance  & 83.0  & 93.0 & 23 \\
        \bottomrule
    \end{tabular}
    \caption{\footnotesize \textbf{Response accuracy, context relevance, and number of task categories} in a sample of 1000 instruction-response pairs synthesized by MetaSynth.}
    \label{tab:response_metrics}
\end{table}

\begin{figure}[t]
    \centering
    \includegraphics[width=\columnwidth, height=0.2\textheight, keepaspectratio]{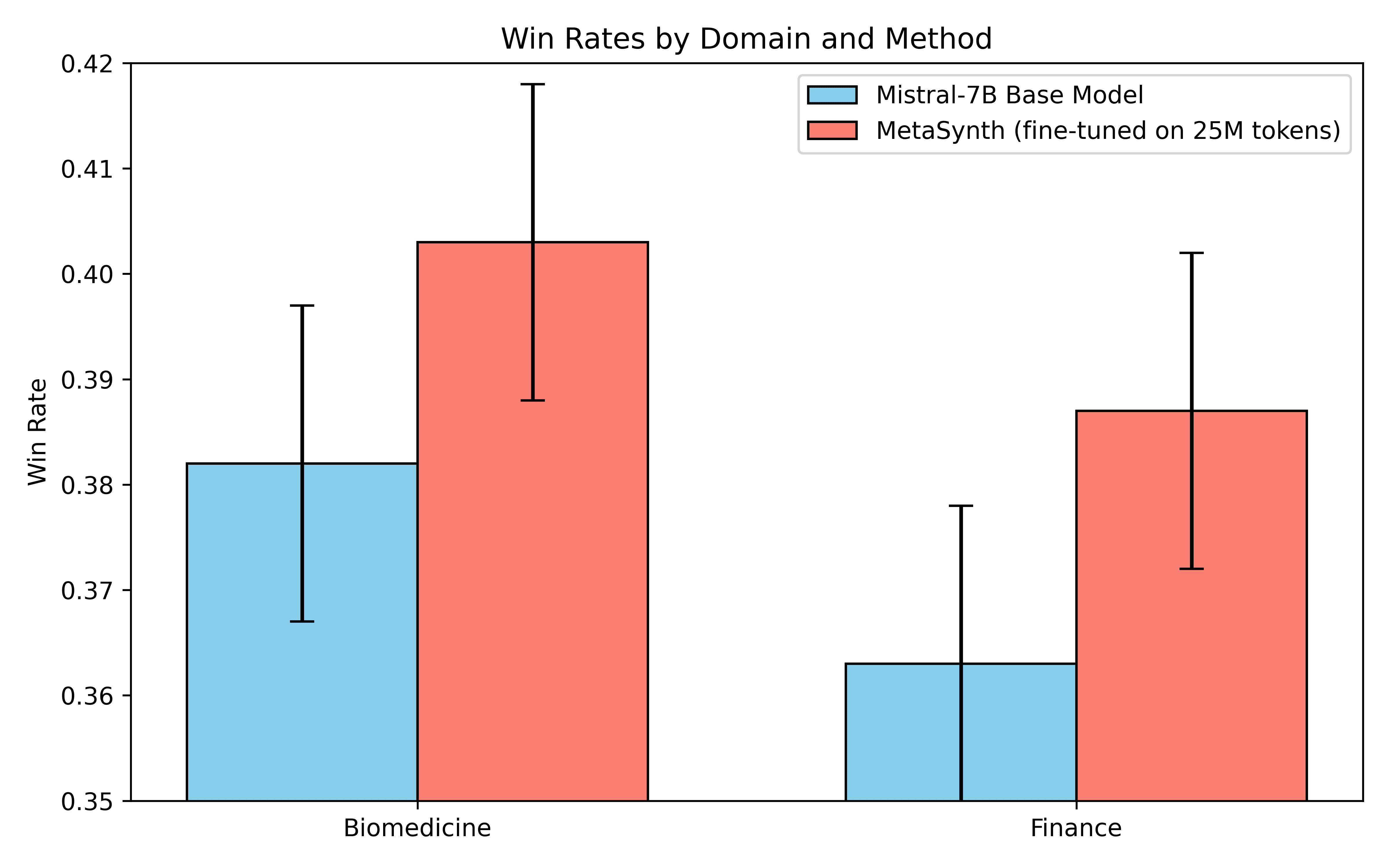}
    \caption{\textbf{Against Claude 3 Sonnet (data generating LLM)}: Win-rates shown for Mistral-7B pretrained on 25M tokens of MetaSynth Documents-Instructions-Responses versus the non-pretrained base model (judged by Claude 3 Opus).}
    \label{fig:winrates}
\end{figure}

\begin{figure}[t]
    \centering
    \includegraphics[width=\columnwidth, height=0.2\textheight, keepaspectratio]{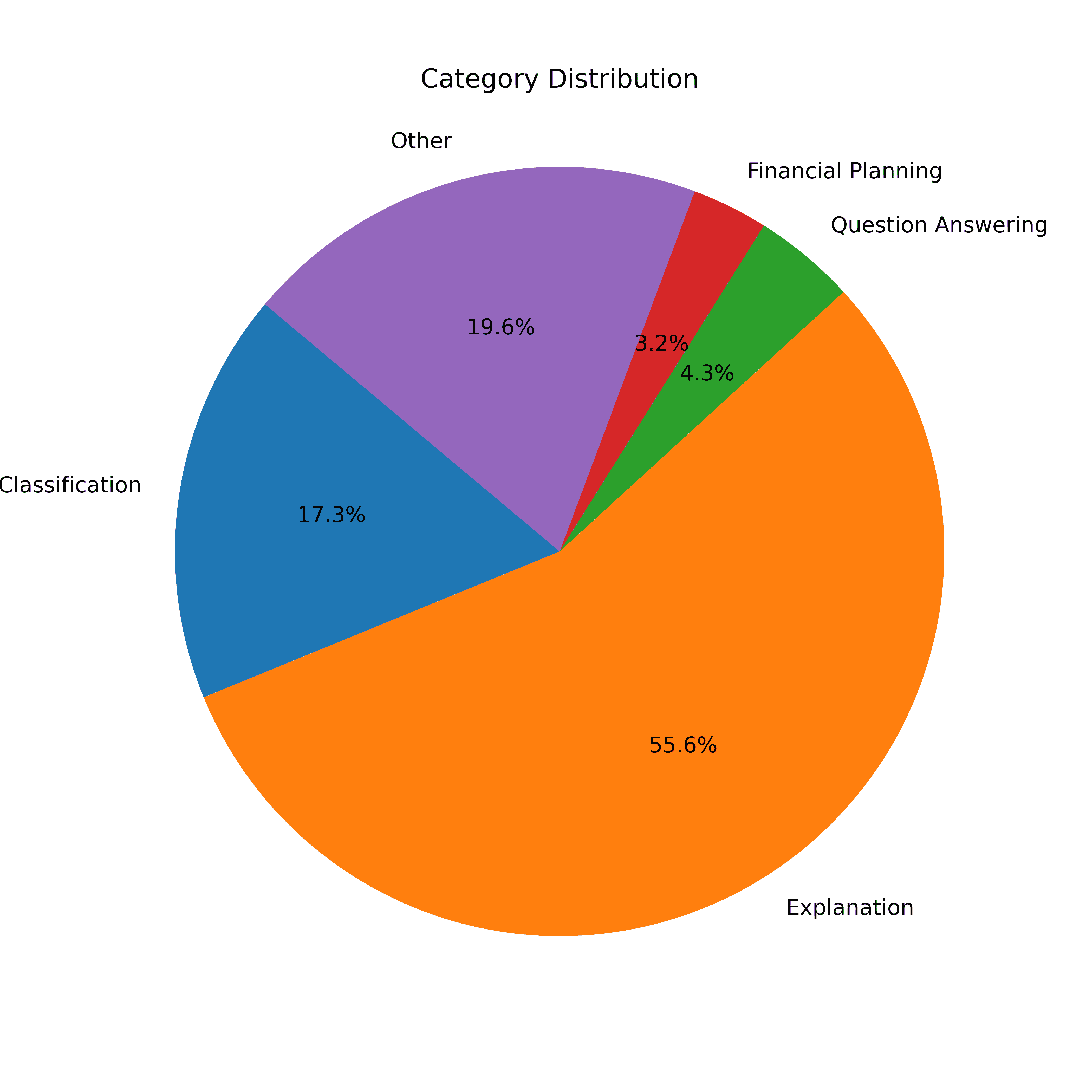}
    \caption{\textbf{Distribution of task scenarios  synthesized by MetaSynth-\textit{Instruct}} in instruction-response pairs from Finance domain.}
    \label{fig:finance_category_pie_chart}
\end{figure}

\begin{figure}[t]
    \centering
    \includegraphics[width=\columnwidth, height=0.2\textheight, keepaspectratio]{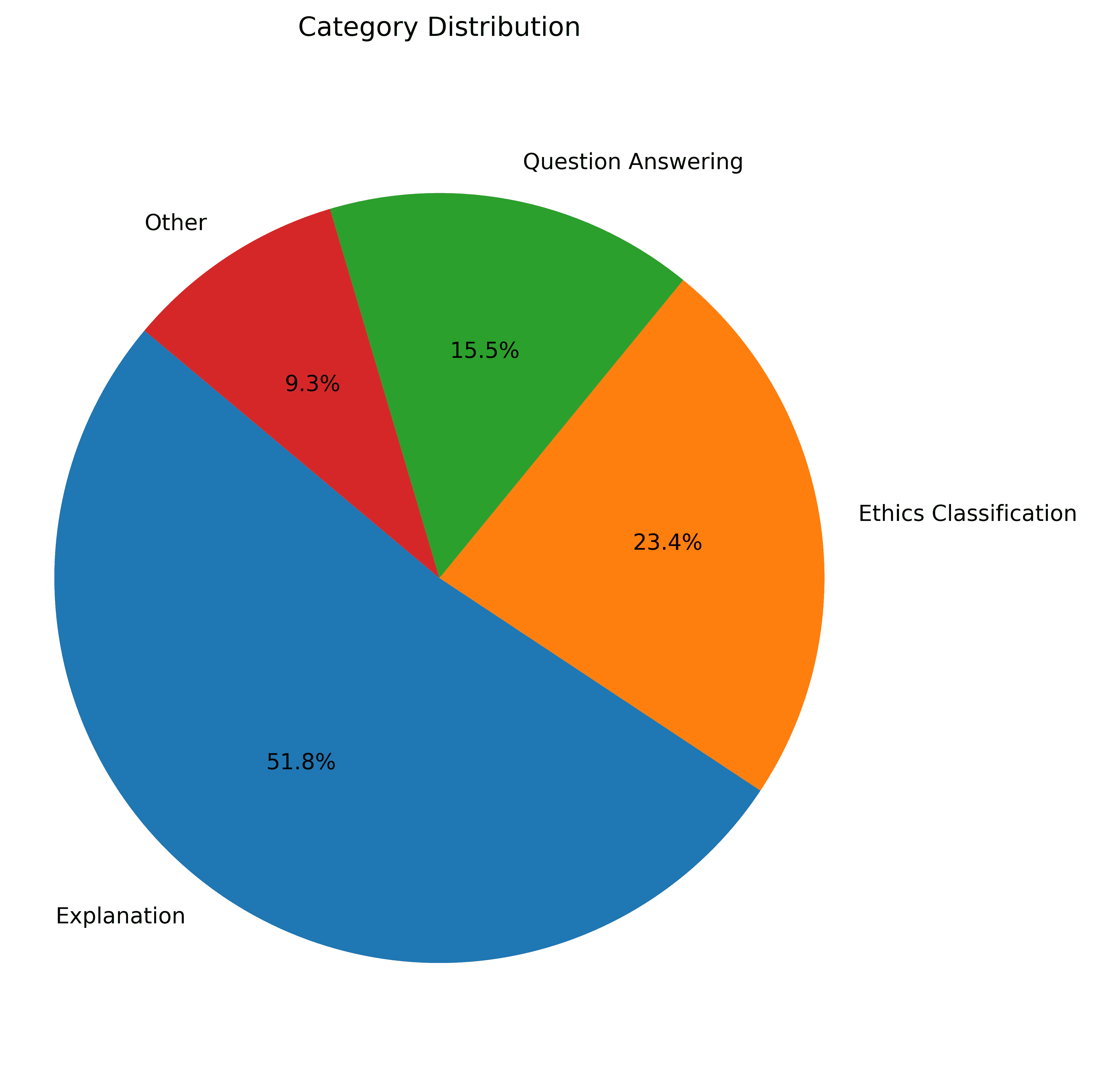}
    \caption{\textbf{Distribution of task scenarios  synthesized by MetaSynth-\textit{Instruct}} in instruction-response pairs from Biomedicine domain.}
    \label{fig:medical_category_pie_chart}
\end{figure}

\section{MetaSynth-\textit{Instruct} Instruction-Response Analysis}
\label{sec:appendixE}
Figure \ref{fig:finance_category_pie_chart} and \ref{fig:medical_category_pie_chart} show the percentages of task scenarios from \newcite{wang2022supernaturalinstructionsgeneralizationdeclarativeinstructions} that occur in a sample of 1000 instruction-response pairs synthesized by MetaSynth, for each domain. Table \ref{tab:response_metrics} shows the number of unique task scenarios that occur in this sample, along with response accuracy and context relevance. \\
\textbf{Response Accuracy} Claude 3 Opus \cite{anthropic2024claude3} is prompted to assess whether a response is accurate based on the instruction and context. A binary indicator score is used to compute accuracy. \\
\textbf{Context Relevance}
The same LLM is also prompted to judge whether the instruction synthesized by MetaSynth is relevant to the context (a synthetic document) given the synthesized response to the context-instruction pair. \\
\textbf{Win Rate} 
We evaluate win-rates—defined as the percentage of times one model is preferred over another in pairwise comparisons—using Claude 3 Opus as the judge. We conduct two comparisons against Claude 3 Sonnet (the synthetic data-generating LLM which also synthesized responses to its own generated instructions): (1) Mistral-7B-v0.3 continually pre-trained on MetaSynth-generated data versus Claude 3 Sonnet, and (2) the base Mistral-7B-v0.3 model (not trained on MetaSynth data) versus Claude 3 Sonnet. Results are shown in figure \ref{fig:winrates}.

Models continually pre-trained on MetaSynth-generated synthetic data achieve higher win rates than their respective base models. Specifically, Mistral-7B-v0.3 continually pretrained on 25M MetaSynth tokens achieves a 40.3\% win rate against Claude 3 Sonnet in biomedicine, outperforming the base (non-pretrained) model's 38.2\%. In finance, it wins 38.7\% of the time compared to the base model's 36.3\%, indicating the utility of our synthetic data.





\section{Templates for Synthesizing Responses to MetaSynth Instructions}
\label{sec:appendixF}
To further elicit diverse responses to our synthesized instructions, we reformat each context and its associated instruction pairs through templated variations. Specifically, for each pair, we apply one of three formats: free-form completion, chain-of-thought (CoT) completion \cite{wei2023chainofthoughtpromptingelicitsreasoning}, and constrained chain-of-thought (cCoT) completion \cite{nayab2025concisethoughtsimpactoutput}. In the cCoT case, a random word limit (a multiple of 50 between 50 and 500) is inserted into the template.  We then construct multiple prompt variants by concatenating each context with randomly sampled subsets of these reformatted instructions--ensuring that every example is incorporated at least once--until a full set of variations is obtained. From these variations, we subsequently sample a diverse, non-redundant set of context–instruction pairs for response synthesis.


\begin{table}[t]
\centering
\footnotesize
\begin{tabular}{@{}l@{\hskip 9pt}r@{\hskip 9pt}r@{\hskip 9pt}r@{\hskip 9pt}r@{}}
\toprule
\textbf{Dataset} & \makecell{\textbf{Synth} \\ \textbf{Data}} & \makecell{\textbf{Real} \\ \textbf{Train}} & \makecell{\textbf{Test} \\ \textbf{Data}} & \makecell{\textbf{Real} \\ \textbf{Total}} \\
\midrule
Headlines & 3,000 & 3,000 & 5,723 & 28,619 \\
FiQA-SA & 646 & 646 & 235 & 881 \\
FPB & 3,876 & 3,876 & 969 & 4,845 \\
\bottomrule
\end{tabular}
\caption{\footnotesize Dataset statistics showing synthetic samples (generated using MetaSynth), real training samples, test samples, and total available samples.}
\label{tab:encoder-dataset-stats}
\end{table}

\section{Dataset Statistics for Encoder Fine-tuning Experiments}
\label{sec:encoder-dataset-stats}


Table~\ref{tab:encoder-dataset-stats} shows the real and synthetic data samples used for training BERT (Section~\ref{encoder-tasks}). For Headlines, we created an 80/20 train-test split, sampled 3,000 real instances from the 22,895 training examples (in the 80\% split), and generated an equivalent number of synthetic samples. We trained BERT on and evaluated on the 20\% test split (5,723 samples). For FiQA-SA, we used the provided training split of 646 samples, generated an equivalent number of synthetic samples with MetaSynth, and evaluated on the provided test split of 235 samples. For FPB, we created an 80/20 split with 3,876 samples for training (also generating an equivalent synthetic set) and evaluated on the remaining 969 samples.

\section{MetaSynth-Execution History}
\label{sec:metasynth-execution-history}

\onecolumn  
\begin{figure*}[t]
    \centering
    \includegraphics[width=\textwidth, height=0.9\textheight, keepaspectratio]
    {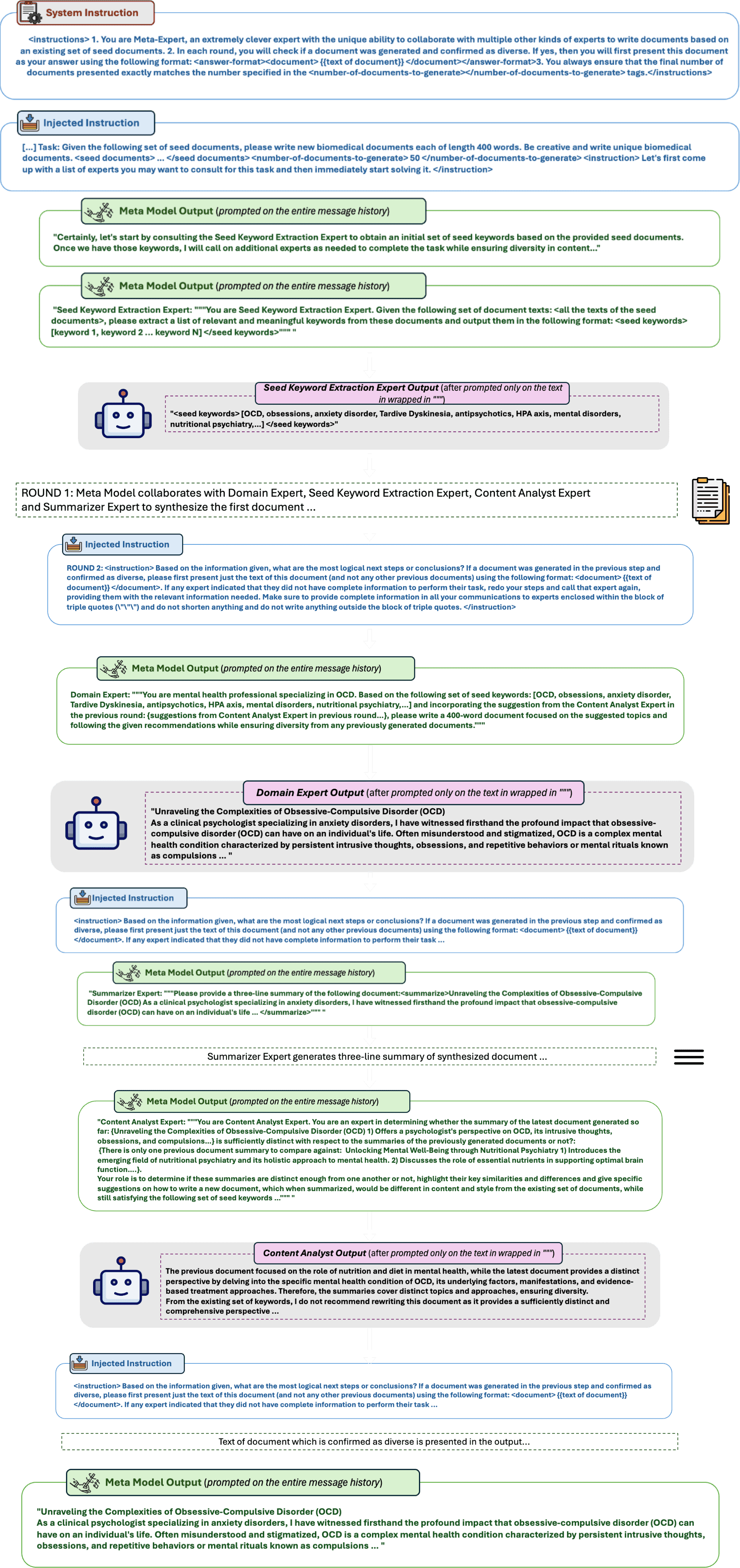}
    \caption{An example \textsc{MetaSynth} meta-prompting history for document synthesis, initialized with system prompt, meta-prompt, and user task description. Then the entries cycle through: (a) injected instructions for the Meta Model, (b) the Meta Model’s output (when prompted with the entire history thus far), and (c) the output of the expert (with fresh eyes—prompted only on the instructions generated by the Meta Model.}
    \label{fig:metasynth_history}
\end{figure*}

\clearpage

\begin{figure*}[t]
    \centering
    \includegraphics[width=\textwidth, height=0.5\textheight, keepaspectratio]
    {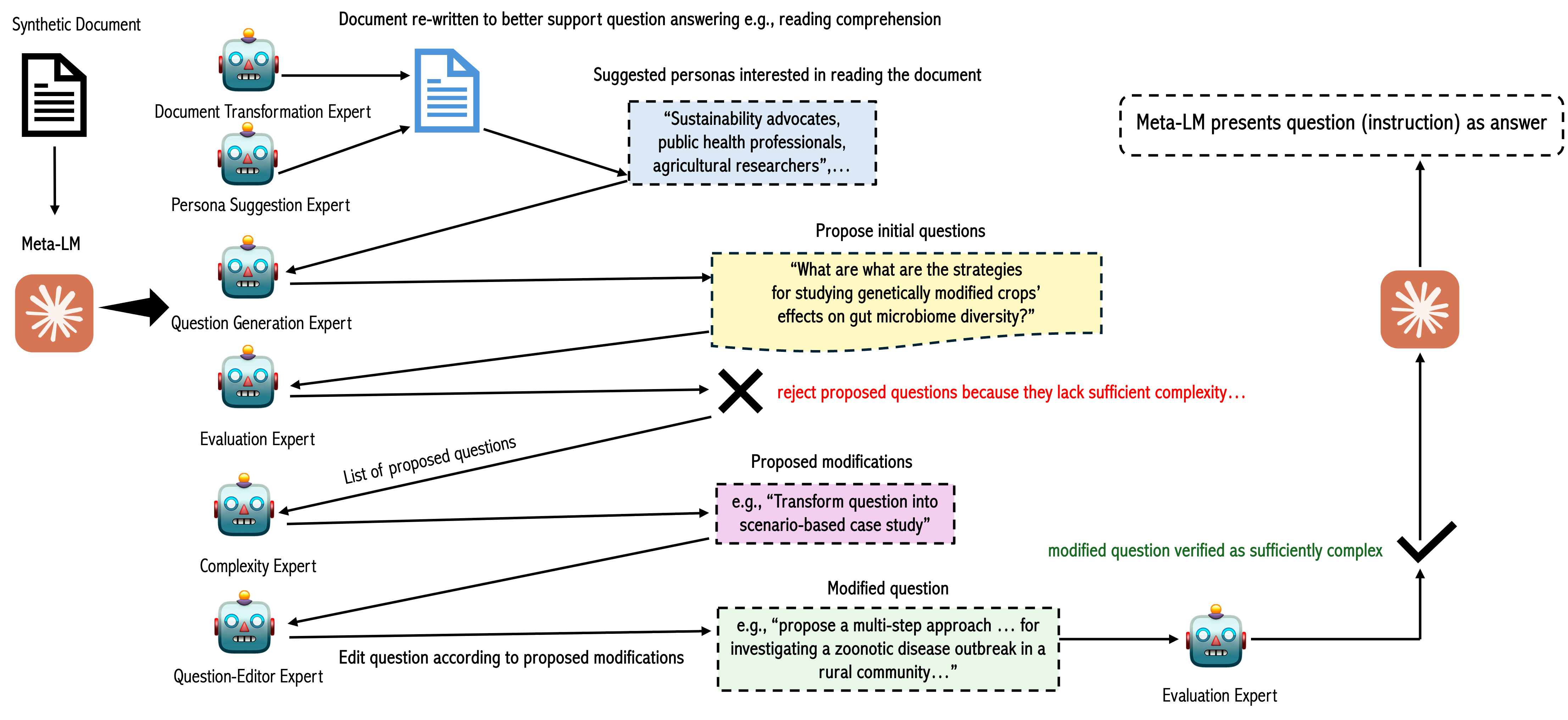}
    \caption{Demonstration of an example \textsc{MetaSynth}-\textit{Instruct} agentic workflow for synthesizing an instruction from a synthetic biomedical document. A meta-LM orchestrates various expert agents that iteratively refine and generate complex instructions in the form of questions conditioned on the text of the synthetic document.}
    \label{fig:metasynth_instruct_figure}
\end{figure*}

\section{MetaSynth-\textit{Instruct}}
\label{sec:appendixH}

Using the above figure as reference, we describe a possible execution history for synthesizing an instruction from a biomedical document as follows:







\textbf{(1)} Given a synthetic document on agriculture and its impact on human health, the meta-LM conjures the following experts to consult: Document Transformation Expert, Persona Suggestion Expert, Question Generation, Evaluation Expert, Complexity Expert, Question Editor expert and two Domain Experts in healthcare and agriculture.

\textbf{(2)} Given the original document text, the Document Transformation Expert is first called by the meta-LM, who identifies the mention of \texttt{Genetically Modified Organisms} and risks/controversies surrounding their use. This expert then reformulates the document to focus more on this aspect.

\textbf{(3)} The meta-LM then calls on the two domain experts (healthcare \& agriculture) to analyze both the original and rewritten document(s) and to provide foundational knowledge. At the same time, the meta-LM also calls Persona Suggestion Expert for a list of readers who would find the document engaging; this expert suggests that sustainability advocates, public health professionals and agricultural researchers would be interested in reading the document.

\textbf{(4)} This feedback along with other information aggregated from various experts by the meta-LM is then passed to a Question Generation Expert which then proposes a set of initial questions e.g. ``\texttt{what are the strategies for studying genetically modified crops' effects on gut microbiome diversity?}''

\textbf{(5)} The meta-LM then calls an Evaluation Expert to determine if the proposed questions are sufficiently complex. However Evaluation Expert may decide that the proposed questions are not sufficiently difficult and reject them. In this case the meta-LM would then call a Complexity Expert to suggest ways on how to make the question more complex. Complexity Expert may suggest to transform the question into a scenario-based case study.

\textbf{(6)} The meta-LM passes Complexity Expert's suggestions to Question Editor Expert which makes the necessary modifications. For example the transformed question might then become: ``\texttt{Propose a multi-step approach which encompasses epidemiological analysis, risk assessment, and stakeholder collaboration, for investigating a zoonotic disease outbreak in a rural community}''. If the Evaluation Expert verifies that this question is sufficiently complex, the Meta-LM accepts the question as an instruction

\section{Template Prompt For Generating Synthetic Documents}
\label{sec:appendixI}

\begin{tcolorbox}[
  enhanced,
  breakable,  
  colback=boxbg,
  colframe=boxborder,
  arc=15pt,
  boxrule=1pt,
  left=5pt, 
  right=5pt 
]

\textbf{\orangetag{\textless{}instruction\textgreater{}}}
Given the following set of seed documents: \orangetag{text of five randomly selected seed documents} please write new \orangetag{financial|biomedical} documents each of length 400 words. Be creative and write unique \orangetag{financial|biomedical} documents. Note: You are not allowed to copy the text of any document in your output verbatim.
\textbf{\orangetag{\textless{}/instruction\textgreater{}}}
\\
\begin{tcolorbox}[colback=blue!10, colframe=blue!20, boxrule=0pt, left=5pt, right=5pt, breakable]
\textbf{\orangetag{\textless{}injected instruction\textgreater{}}}

You are also provided with the text(s) of all documents that you have previously generated: \orangetag{texts of all previously synthesized documents.}

Please generate the next document, ensuring that it is diverse from all previously generated documents, while also being similar to the set of seed documents. Remember to present the text of each new document using the following format: \orangetag{\textless{}document\textgreater{}} \orangetag{text of document} \orangetag{\textless{}/document\textgreater{}}

\textbf{\orangetag{\textless{}/injected instruction\textgreater{}}}

\textbf{answer format:}\orangetag{\textless{}document\textgreater{}} \orangetag{text of document} \orangetag{\textless{}/document\textgreater{}}
\end{tcolorbox}

\end{tcolorbox}



\section{Meta-Prompts For Generating Synthetic Documents}
\label{sec:appendixJ}
\subsubsection{System Prompt}

\begin{tcolorbox}[
  enhanced,
  breakable,  
  colback=boxbg,
  colframe=boxborder,
  arc=15pt,
  boxrule=1pt,
  left=5pt, 
  right=5pt 
]
\textbf{\orangetag{\textless{}instructions\textgreater{}}}
\begin{enumerate}[leftmargin=*]
  \item You are Meta-Expert, an extremely clever expert with the unique ability to collaborate with multiple other kinds of experts to write documents based on an existing set of seed documents.
  \item In each round, you will check if a document was generated and confirmed as diverse. If yes, then you will first present this document as your answer using the following format: \orangetag{\textless{}answer-format\textgreater{}}\orangetag{\textless{}document\textgreater{}} \orangetag{text of document} \orangetag{\textless{}/document\textgreater{}}\orangetag{\textless{}/answer-format\textgreater{}}
  \item You always ensure that the final number of documents presented exactly matches the number specified in the \orangetag{\textless{}number-of-documents-to-generate\textgreater{}\textless{}/number-of-documents-to-generate\textgreater{}} tags. 
\end{enumerate}

If you have presented the last document and the number of documents you have presented equals to what was specified in the \orangetag{\textless{}number-of-documents-to-generate\textgreater{}\textless{}/number-of-documents-to-generate\textgreater{}} tags, please output: \orangetag{\textless{}END\textgreater{}}.

Otherwise, based on the information given, what are the most logical next steps or conclusions? Make sure to provide complete information in all your communications to experts enclosed within the block of triple quotes (\textrm{\textquotedblleft\textquotedblleft\textquotedblleft}) and do not shorten anything and do not write anything outside the block of triple quotes. If a document was generated in the previous step and confirmed as diverse then you first need to present just the text of this document (and not any other previous documents) as your answer using the following format: \orangetag{\textless{}document\textgreater{}} \orangetag{text of document} \orangetag{\textless{}/document\textgreater{}} before proceeding to the next round

\textbf{\orangetag{\textless{}/instructions\textgreater{}}}
\end{tcolorbox}


\subsubsection{User Prompt}
\begin{tcolorbox}[
  enhanced,
  breakable,  
  colback=boxbg,
  colframe=boxborder,
  arc=15pt,
  boxrule=1pt,
  left=5pt, 
  right=5pt 
]
\textbf{\orangetag{\textless{}role of meta-expert\textgreater{}}}
\begin{itemize}[leftmargin=*]
  \item oversees communication between experts
  \item calls different kinds of experts to write diverse documents e.g. ``Seed Keyword Extraction Expert'', ``Domain Expert'', ``Summarizer Expert'', ``Writing/Linguistics Expert'', ``Content Analyst Expert'' etc.
  \item applies critical thinking and judgment skills
  \item always calls other experts in the right order
  \item assigns personas to experts if needed e.g. ``You are a policy analyst specialized in...\orangetag{some domain}''
  \item always remembers how many documents have been written so far
  \item always consults with "Seed Keyword Extraction Expert" to extract a set of seed keywords using the texts of all of the documents provided in the \orangetag{\textless{}seed documents\textgreater{} \textless{}/seed documents\textgreater{}} tags below. Make sure to provide ``Seed Keyword Extraction Expert'' with the full texts of all of the documents which are enclosed in the \orangetag{\textless{}seed documents\textgreater{} \textless{}/seed documents\textgreater{}} tags below
  \item always consults with ``Summarizer Expert'' after each new document is written for a three-line summary. To obtain a summary from ``Summarizer Expert'', make sure to give ``Summarizer Expert'' the full text of each new document that is generated
  \item always memorizes the summaries of all documents generated so far
  \item always consults with ``Content Analyst Expert'' to compare the summary of each new generated document with the summaries of all previously generated documents in order to successfully determine the content diversity of the new document
  \item if ``Content Analyst Expert'' determines that the summary of the new document is not sufficiently distinct with respect to the existing set of summaries, then please reject this document and use the feedback from ``Content Analyst Expert'' to call another expert and ask them to write a new document from scratch
  \item only interacts with one expert at a time and waits for the expert to reply back before calling for another expert
  \item your interactions with each of the other experts are isolated, so please include all relevant information in every call
  \item provide clear, unambiguous instructions with complete information when communicating with experts
  \item always keep in mind that except for you, all other experts have no memory! Therefore always provide all relevant information when contacting them
  \item verify that the new document that was written is a valid document if you are uncertain
  \item verify that the length of the new document is exactly 400 words
  \item consult with at least two experts for confirmation that a document is sufficiently diverse before presenting it as your answer
  \item if an expert verifies that the new document is not very diverse, call a new expert to rewrite it
  \item aim to present all of the requested documents within 256 rounds or fewer
  \item avoid repeating identical questions to experts
  \item only you as the Meta-Expert can communicate with other experts. The other experts cannot talk among themselves
  \item when presenting your answer, make sure that you or any other expert(s) did not copy and paste the text of any seed document verbatim
  \item ensure that the count of the number of generated documents matches the number specified in the
  \orangetag{\textless{}number-of-documents-to-generate\textgreater{} \textless{}/number-of-documents-to-generate\textgreater{}} 
  \item ensure that each document you present as an answer contains the actual texts of the documents in full and not it's summary
  \item once you are certain that a document is sufficiently diverse, present it in the answer format specified below before proceeding to the next round
\end{itemize}
\textbf{\orangetag{\textless{}/role of meta-expert\textgreater{}}}
\\\\
\textbf{\orangetag{\textless{}rules for communicating with other experts\textgreater{}}}

  \orangetag{\textless{}format \textgreater{}} expert name: \textrm{\textquotedblleft\textquotedblleft\textquotedblleft}\orangetag{detailed instructions}\textrm{\textquotedblright\textquotedblright\textquotedblright} \orangetag{\textless{}/format \textgreater{}}
  
\orangetag{\textless{}example\textgreater{}}
  
  \orangetag{\textless{}name\textgreater{}} Seed Keyword Extraction Expert \orangetag{\textless{}/name\textgreater{}}
  
  \orangetag{\textless{}instruction\textgreater{}}
      You are Seed Keyword Extraction Expert. Given the following set of document texts: \orangetag{text of each seed document}, please extract a list of relevant and meaningful keywords from these documents and output them in the following format: \orangetag{\textless{}seed keywords\textgreater{}} [keyword 1, keyword 2 ... keyword N] \orangetag{\textless{}/seed keywords\textgreater{}}.
      
    \orangetag{\textless{}/instruction\textgreater{}}  
\orangetag{\textless{}/example\textgreater{}}

\orangetag{\textless{}example\textgreater{}}

\orangetag{\textless{}name\textgreater{}} Content Analyst Expert \orangetag{\textless{}/name\textgreater{}}

  \orangetag{\textless{}instruction\textgreater{}}
  
    You are Content Analyst Expert. You are an expert in determining whether the summary of the latest document generated so far: \orangetag{three-line summary of last generated document} is sufficiently distinct with respect to the summaries of the previously generated documents or not?: \orangetag{set of three-line summaries of each previously generated document}.
    
    Your role is to determine if these summaries are distinct enough from one another or not, highlight their key similarities and differences and give specific suggestions on how to write a new document, which when summarized, would be different in content and style from the existing set of documents, while still satisfying the following set of seed keywords: \orangetag{list of seed keywords which were generated by Seed Keyword Extraction Expert and which may also include suggestions from other experts}.
    
    You must also indicate whether this document, based on its summary, should be re-written if it is not sufficiently distinct. If you think it should be re-written, please give specific suggestions on how to re-write it.
    
    You must also suggest new seed keywords to be added to the current set of keywords that are related yet sufficiently distinct from the current set of seed keywords. For your reference, here are the current set of seed keywords: \orangetag{list of seed keywords which were generated by Seed Keyword Extraction Expert and which may also include suggestions from other experts}.
    
    Keep in mind that summaries are just a proxy for comparing documents and you should always suggest how to write a new document, NOT a new summary.
    
    You must monitor the diversity of topics in recent document summaries. If you detect a pattern of focusing on subtopics related to only a few keywords, suggest a change of topic. Your role is to encourage exploration of a wide range of themes, rather than allowing deep dives into a limited number of areas. Propose new directions that broaden the scope of discussion and ensure a balanced coverage of topics.
    
    To enhance diversity, you should also suggest new persona's for another document writer to adopt, or to write a document in a new format or to write a document from a different perspective.
    
    Ideally, your suggestion(s) must ensure that the next document covers a theme or perspective that is different from the previously generated documents.

  \orangetag{\textless{}/instruction\textgreater{}}

\orangetag{\textless{}/example\textgreater{}}
\\\\
\orangetag{\textless{}example\textgreater{}}

\orangetag{\textless{}name\textgreater{}} Summarizer Expert \orangetag{\textless{}/name\textgreater{}}
  
  \orangetag{\textless{}instruction\textgreater{}}
  
    You are Summarizer Expert. Please provide a three-line summary of the following document: \orangetag{\textless{}summarize\textgreater{}} \orangetag{text of document to be summarized}\orangetag{\textless{}/summarize\textgreater{}}.
  
  \orangetag{\textless{}/instruction\textgreater{}}

\orangetag{\textless{}/example\textgreater{}}
\\\\
    \orangetag{\textless{}example\textgreater{}}
    
    \orangetag{\textless{}name\textgreater{}} Domain Expert \orangetag{\textless{}/name\textgreater{}}
    
    \orangetag{\textless{}instruction\textgreater{}}
    
    You are an expert in the following domain: \orangetag{name of domain}. Given the following set of seed keywords: \orangetag{list of seed keywords which were extracted by Seed Keyword Extraction Expert and which may also include suggestions from other experts}, and the following feedback from another expert: \orangetag{one or more suggestions from another expert}, write a document that follows these suggestions and focuses on a subset of the seed keywords. Ensure that the length of the document is exactly 400 words. Be creative and write a unique document. 
    
    \orangetag{\textless{}/instruction\textgreater{}}
    
    \orangetag{\textless{}/example\textgreater{}}
    
\textbf{\orangetag{\textless{}/rules for communicating with other experts\textgreater{}}}
\\
\begin{tcolorbox}[colback=blue!10, colframe=blue!20, boxrule=0pt, left=5pt, right=5pt, breakable]
\textbf{\orangetag{\textless{}important note\textgreater{}}}
  \begin{itemize}[leftmargin=*]
    \item The expert types listed above are just examples; you should consult completely new kinds of experts based on the task's needs.
    \item Please ensure that you are presenting the full text of each document in your answer and NOT its summary.
    \item In each round, you will check if a document was generated and confirmed as diverse. If yes, then you must first present this document as your answer using the \orangetag{answer format} below.
  \end{itemize}
\textbf{\orangetag{\textless{}/important note\textgreater{}}}

\textbf{answer format:}\orangetag{\textless{}document\textgreater{}} \orangetag{text of document} \orangetag{\textless{}/document\textgreater{}}
\end{tcolorbox}
\end{tcolorbox}

\subsection{Task Description}

\begin{tcolorbox}[
  enhanced,
  breakable,  
  colback=boxbg,
  colframe=boxborder,
  arc=15pt,
  boxrule=1pt,
  left=5pt, 
  right=5pt 
]
Given the following set of seed documents, please write new finance|biomedical documents each of length 400 words. Be creative and write unique finance|biomedical documents.
\end{tcolorbox}


\section{Meta-Prompts For Synthetic Instructions}
\label{sec:appendixK}
\subsection{System Prompt}

\begin{tcolorbox}[
  enhanced,
  breakable,  
  colback=boxbg,
  colframe=boxborder,
  arc=15pt,
  boxrule=1pt,
  left=5pt, 
  right=5pt 
]
\textbf{\orangetag{\textless{}instructions\textgreater{}}}
\begin{enumerate}[leftmargin=*]
  \item You are Meta-Expert, an extremely clever expert with the unique ability to collaborate with multiple other kinds of experts to create complex questions from a given document.
  \item In each round, you will check if one or more questions(s) were generated and confirmed as unique and diverse. If yes, then you will present each of these question(s) as your output using the following format: \orangetag{\textless{}questions\textgreater{}}\orangetag{\textless{}question\textgreater{}}\orangetag{first question}\orangetag{\textless{}/question\textgreater{}}
  ...
  \orangetag{\textless{}question\textgreater{}}\orangetag{last question}\orangetag{\textless{}/question\textgreater{}}
  \orangetag{\textless{}/questions\textgreater{}}
\end{enumerate}

If you have presented a sufficient number of diverse and complex questions from this document, please output: \orangetag{\textless{}END\textgreater{}}.

Otherwise, based on the information given, what are the most logical next steps or conclusions? Make sure to provide complete information in all your communications to experts enclosed within the block of triple quotes (\textrm{\textquotedblleft\textquotedblleft\textquotedblleft}) and do not shorten anything and do not write anything outside the block of triple quotes. If one or more examples(s) were generated in the previous step, then you need to present each of these example(s) as your output using the following format: \orangetag{\textless{}questions\textgreater{}}
\orangetag{\textless{}question\textgreater{}}\orangetag{first question}\orangetag{\textless{}/question\textgreater{}}
...
\orangetag{\textless{}question\textgreater{}}\orangetag{last question}\orangetag{\textless{}/question\textgreater{}}
\orangetag{\textless{}/questions\textgreater{}}
\textbf{\orangetag{\textless{}/instructions\textgreater{}}}
\end{tcolorbox}


\subsection{User Prompt}

\begin{tcolorbox}[
  enhanced,
  breakable,  
  colback=boxbg,
  colframe=boxborder,
  arc=15pt,
  boxrule=1pt,
  left=5pt, 
  right=5pt 
]
\textbf{\orangetag{\textless{}role of meta-expert\textgreater{}}}
\begin{itemize}[leftmargin=*]
  \item oversees communication between experts
  \item uses the following task description: \orangetag{PLACEHOLDER}, and the text of the document given below, to call different kinds of experts to generate diverse questions
  \item for any given document, calls a ``Document Transformation Expert'' which can re-write the text of the document to better support generating diverse questions
  \item for any given document, calls a ``Persona Suggestion Expert'' to suggest a list of persona's or other expert types that would be interested in the contents of that document
  \item for any given document, calls an ``Question Generation Expert'' which:
    \begin{enumerate}
      \item uses the document text (which can either be the original document text or the transformed document text as suggested by Document Transformation Expert)
      \item uses the list of persona's suggested by the ``Persona Suggestion Expert'' in the previous round, to create questions from the point of view of each suggested persona, based upon the following task description: \orangetag{PLACEHOLDER}
    \end{enumerate}
  \item for any given document, calls other unique types of experts that can give suggestions on how to create complex and diverse questions, using the either the original text of the document or the transformed document text as suggested by ``Document Transformation Expert''
  \item Before presenting the final set of questions, calls ``Complexity Expert'' which:
    \begin{enumerate}
      \item uses the document text (which can either be the original document text or the transformed document text as suggested by Document Transformation Expert)
      \item uses the set of questions generated by ``Question Generation Expert''
      \item Gives suggestions on how to modify each question in order to complicate it
    \end{enumerate}
  \item Before presenting the final set of questions, calls ``Question Editor Expert'' which uses the suggestions of ``Complexity Expert'' to output a final set of re-written/modified questions
  \item applies critical thinking and judgment skills
  \item always calls other experts in the right order
  \item always remembers how many questions have been generated so far
  \item only interacts with one expert at a time and waits for the expert to reply back before calling for another expert
  \item your interactions with each of the other experts are isolated, so you must include all relevant information in every call
  \item provide clear, unambiguous instructions with complete information when communicating with experts
  \item always keep in mind that except for you, all other experts have no memory! Therefore always provide all relevant information when contacting them
  \item consult at least two or more experts to verify that each new question that was generated is a valid and diverse question if you are uncertain
  \item if you or any other expert thinks that the question(s) generated are not very diverse or complex, call a new expert to rewrite them or re-do your steps
  \item aim to present all of the questions within 128 rounds or fewer
  \item avoid repeating identical information to experts
  \item only you as the Meta-Expert can communicate with other experts. The other experts cannot talk among themselves
  \item once the final set of questions are ready and you are certain that all of the generated questions are sufficiently complex and diverse and no more questions can be generated from the given document, then at the end, present the final set of questions in the output format specified below
\end{itemize}
\textbf{\orangetag{\textless{}/role of meta-expert\textgreater{}}}
\\\\
\textbf{\orangetag{\textless{}rules for communicating with other experts\textgreater{}}}

  \orangetag{\textless{}format \textgreater{}} expert name: \textrm{\textquotedblleft\textquotedblleft\textquotedblleft}\orangetag{detailed instructions}\textrm{\textquotedblright\textquotedblright\textquotedblright} \orangetag{\textless{}/format \textgreater{}}
  
\orangetag{\textless{}example\textgreater{}}
  
  \orangetag{\textless{}name\textgreater{}} Document Transformation Expert \orangetag{\textless{}/name\textgreater{}}
  
  \orangetag{\textless{}instruction\textgreater{}}
  
    You are Document Transformation Expert. Given the following document: \orangetag{text of document}, and given the following task description: \orangetag{PLACEHOLDER},
    transform or re-write the document in such a way that would make it easier to create questions from the document text as stated in the given task.
    
  \orangetag{\textless{}/instruction\textgreater{}}
    
\orangetag{\textless{}/example\textgreater{}}
\\\\
\orangetag{\textless{}example\textgreater{}}

\orangetag{\textless{}name\textgreater{}} Persona Suggestion Expert \orangetag{\textless{}/name\textgreater{}}

  \orangetag{\textless{}instruction\textgreater{}}
  
    You are Persona Suggestion Expert. Given the text of the following document: 
    \orangetag{text of document}, suggest a list of people that would be interested in this document.
    
  \orangetag{\textless{}/instruction\textgreater{}}

\orangetag{\textless{}/example\textgreater{}}
\\\\
\orangetag{\textless{}example\textgreater{}}

\orangetag{\textless{}name\textgreater{}} Question Generation Expert \orangetag{\textless{}/name\textgreater{}}
  
  \orangetag{\textless{}instruction\textgreater{}}
  
    You are Question Generation Expert. Given the following information:
    1. document text: \orangetag{text of document} 
    2. list of persona's: \orangetag{full list of persona's suggested by the Persona Suggestion Expert}
    3. Task: \orangetag{PLACEHOLDER}

    your job is to create diverse and complex questions as described in the given task role-playing as the following persona:
    \orangetag{each persona in the list of persona's suggested by the Persona Suggestion Expert}
    The questions you create must satisfy the given task description and must be based only on the text of the document.
    Ensure that each question can be answered entirely from the information present in the contexts.
    Phrases like `based on the document', `according to the document', `As a ...' etc., are not allowed to appear in the question.
    Ensure the each question is clear and unambiguous.
    
  \orangetag{\textless{}/instruction\textgreater{}}

\orangetag{\textless{}/example\textgreater{}}
\\\\
\orangetag{\textless{}example\textgreater{}}

\orangetag{\textless{}name\textgreater{}} Complexity Expert \orangetag{\textless{}/name\textgreater{}}
  
  \orangetag{\textless{}instruction\textgreater{}}
  
    You are Complexity Expert. Given the following questions: \orangetag{text of each question proposed by ``Question Generation Expert''} and the following context: \orangetag{text of document}
    please suggest ways to modify each question to increase its complexity or make it more intricate based on the context. For example you may suggest to: add some context to the original question, which states the importance of the question, explains background knowledge, or adds other reasonable information.
    You may also suggest to change the questions into a different format or style, e.g., imperative statements, length requirements for the answer, etc. You may also suggest to change the questions into elongated questions that require to elaborate on specific topics or discuss a certain point. 
    You may also suggest including some examples, data points, or references or putting some constraints on the answer for e.g. that it must follow specific formats or styles, e.g., no more than 100 words including specific words, etc.
    You may also suggest adding a scenario or condition that affects the context of the question.
    You may also suggest rewriting the question into a multi-hop reasoning question based on the provided context, which would require the reader to make multiple logical connections or inferences using the information available.
    You may also suggest any other reasonable modification not described above, that would make the task more detailed. Be creative and come up with novel modifications. 
    Return both the text of the original question and the proposed modification in the following format:
    
    \orangetag{\textless{}original question\textgreater{}}
    \orangetag{text of original question}\orangetag{\textless{}/original question\textgreater{}}
    \orangetag{\textless{}proposed modification\textgreater{}}\orangetag{proposed modification}\orangetag{\textless{}/proposed modification\textgreater{}}
    
  \orangetag{\textless{}/instruction\textgreater{}}

\orangetag{\textless{}/example\textgreater{}}
\\\\
\orangetag{\textless{}example\textgreater{}}

\orangetag{\textless{}name\textgreater{}} Question Editor Expert \orangetag{\textless{}/name\textgreater{}}
  
  \orangetag{\textless{}instruction\textgreater{}}
  
    You are Question Editor Expert. You are given the following pairs of questions and proposed modifications to those questions: \orangetag{each pair of \textless{}original question\textgreater{}\orangetag{text of original question}\textless{}/original question\textgreater{} \textless{}proposed modification\textgreater{} \orangetag{the proposed modification} \textless{}/proposed modification\textgreater{} as suggested by ``Complexity Expert''}
    Rewrite each question according its corresponding proposed modification and output the modified questions.
    Ensure that the rewritten questions are clear and unambiguous.
    
  \orangetag{\textless{}/instruction\textgreater{}}

\orangetag{\textless{}/example\textgreater{}}

\textbf{\orangetag{\textless{}/rules for communicating with other experts\textgreater{}}}
\\
\begin{tcolorbox}[colback=blue!10, colframe=blue!20, boxrule=0pt, left=5pt, right=5pt, breakable]
\textbf{\orangetag{\textless{}important note\textgreater{}}}
  \begin{itemize}[leftmargin=*]
    \item The expert types listed above are just suggestions; you should also consult completely new kinds of experts based on the task requirements
    \item When outputting the final list of questions the name of any Expert or Persona must not appear in the text of any question
    \item Ensure that only the generated questions are present in the output with no extraneous information
  \end{itemize}
\textbf{\orangetag{\textless{}/important note\textgreater{}}}

\textbf{output format:}\orangetag{\textless{}questions\textgreater{}} 
\orangetag{\textless{}question\textgreater{}}\orangetag{first question}\orangetag{\textless{}/question\textgreater{}}
...
\orangetag{\textless{}question\textgreater{}}\orangetag{last question}\orangetag{\textless{}/question\textgreater{}}
\orangetag{\textless{}/questions\textgreater{}}
\end{tcolorbox}

\end{tcolorbox}

\subsection{Task Description For Synthesizing Instructions}

\begin{tcolorbox}[
  enhanced,
  breakable,  
  colback=boxbg,
  colframe=boxborder,
  arc=15pt,
  boxrule=1pt,
  left=5pt, 
  right=5pt 
]
\orangetag{\textless{}task\textgreater{}}
    \orangetag{\textless{}name\textgreater{}}Creating Complex Questions\orangetag{\textless{}/name\textgreater{}}
    
    \orangetag{\textless{}description\textgreater{}}
    
        The task is to:
        \begin{enumerate}[leftmargin=*]
            \item Create complex questions or problems.
            \item Ensure that the questions require multi-step reasoning, critical thinking, or creative problem-solving.
            \item Each question should not be more than one-hundred (100) words.
            \item The questions should in various styles and in the formats of various tasks e.g. reading comprehension, mathematical problems or other complex domain-specific tasks etc.
            \item Reading comprehension style questions can be divided into: multiple-choice questions (MCQs), literal comprehension questions with short answers, numerical/discrete reasoning, critical comprehension, evaluative comprehension, vocabulary and language use (e.g. fill-in-the-blank),
            relationship comprehension, sequencing events, argument strengthening/weakening, or assumption, inference, flaws in reasoning type of questions etc.,
            \item The question style must test the ability to consider multiple perspectives, engage in hypothetical scenarios and problem-solving.
            \item The questions may requiring making unexpected connections, analyzing arguments, identifying logical fallacies, paradoxes, or evaluating evidence.
            \item Ensure that the questions are clear, well-structured and unambiguous, despite their complexity.
        \end{enumerate}
    \orangetag{\textless{}/description\textgreater{}}
    
    \orangetag{\textless{}evaluation\textgreater{}}
    
        \orangetag{\textless{}metric\textgreater{}} Human evaluation of the diversity, complexity, difficulty, and level of thinking required to answer each question. \orangetag{\textless{}/metric\textgreater{}}

    \orangetag{\textless{}/evaluation\textgreater{}}

\orangetag{\textless{}/task\textgreater{}}
\end{tcolorbox}

\subsection{Task Description For Synthesizing Encoder LM Datasets}
\label{encoder-task-description}
\subsubsection{Headlines:}
\begin{tcolorbox}[
  enhanced,
  breakable,  
  colback=boxbg,
  colframe=boxborder,
  arc=15pt,
  boxrule=1pt,
  left=5pt, 
  right=5pt 
]
\orangetag{\textless{}task\textgreater{}}
    \orangetag{\textless{}name\textgreater{}}News Headline Generation\orangetag{\textless{}/name\textgreater{}}
    
    \orangetag{\textless{}description\textgreater{}}
        The task is to:
        \begin{enumerate}[leftmargin=*]
            \item Generate creative headlines in the style of The Onion and HuffPost that can serve as high quality examples for sarcasm classification. 
            \item Ensure there is a balance of sarcastic and serious headlines.
            \item The headlines should not contain the literal word "sarcasm" or "serious".
            \item The headlines should be grammatical and well-written.
        \end{enumerate}
    \orangetag{\textless{}/description\textgreater{}}
    
    \orangetag{\textless{}task-examples\textgreater{}}
        \begin{enumerate}[leftmargin=*]
            \item ``helpful waitress asks recently seated couple if they`ve eaten food before''
            \item ``must-see tv shows you can't miss this fall''
            \item ``as per tradition, election results officially certified with two barks of approval from electoral collie''
        \end{enumerate}
    \orangetag{\textless{}/task-examples\textgreater{}}
    
    \orangetag{\textless{}evaluation\textgreater{}}
        \orangetag{\textless{}metric\textgreater{}}Human evaluation of the creativity and relevance of generated headlines.\orangetag{\textless{}/metric\textgreater{}}
    \orangetag{\textless{}/evaluation\textgreater{}}
\orangetag{\textless{}/task\textgreater{}}
\end{tcolorbox}


\subsubsection{FiQA-SA ABSA:}

\begin{tcolorbox}[
  enhanced,
  breakable,  
  colback=boxbg,
  colframe=boxborder,
  arc=15pt,
  boxrule=1pt,
  left=5pt, 
  right=5pt 
]
\orangetag{\textless{}task\textgreater{}}
        \orangetag{\textless{}name\textgreater{}}Data Generation For Aspect Based Sentiment Analysis (ABSA) \orangetag{\textless{}/name\textgreater{}}
        \orangetag{\textless{}description\textgreater{}}
            The task is to:
                \begin{enumerate}[leftmargin=*]
                    \item Generate diverse example sentences that mention specific aspects related to companies, products, or services.
                    \item Each example sentence should contain only one clear aspect that could be subject to sentiment analysis.
                    \item The aspects should be varied and could include company names, stock symbols, product features, or service characteristics.
                    \item The aspects must always be present as a substring in the generated sentence.
                    \item The example sentences should be written in a style similar to social media posts, news headlines, or customer reviews.
                    \item The format of each generated example should be as follows: sentence: \{text of sentence\} aspect: \{the relevant aspect\}
                    \item Ensure a balance of potentially positive, negative, and neutral contexts for the aspects.
                    \item The sentences should be in English.
                \end{enumerate}
        \orangetag{\textless{}/description\textgreater{}}
        \orangetag{\textless{}examples\textgreater{}}
            \begin{enumerate}[leftmargin=*]
                \item sentence: \#Tesla: Model X Recall Adds To Reliability Issues \$TSLA https://t.co/jVXQ4DoXnP aspect: TSLA
                \item sentence: \$AAPL AAPL: Gundlach Slams iPad mini, Sees Downside to \$425. http://stks.co/bDqV aspect: AAPL
                \item sentence: \$UBNT still having some trouble at the resistance line. Should resolve soon.@cheri1 @strattonite http://stks.co/c0sU4 aspect: UBNT
            \end{enumerate}
        \orangetag{\textless{}/examples\textgreater{}}
        \orangetag{\textless{}evaluation\textgreater{}}
            \orangetag{\textless{}metric\textgreater{}}Human evaluation of the diversity, relevance, and quality of generated example sentences and their corresponding aspects.\orangetag{\textless{}/metric\textgreater{}}
        \orangetag{\textless{}/evaluation\textgreater{}}
\orangetag{\textless{}/task\textgreater{}}
\end{tcolorbox}

\subsubsection{Financial Phrase Bank (FPB):}

\begin{tcolorbox}[
  enhanced,
  breakable,  
  colback=boxbg,
  colframe=boxborder,
  arc=15pt,
  boxrule=1pt,
  left=5pt, 
  right=5pt 
]
\orangetag{\textless{}task\textgreater{}}
    \orangetag{\textless{}name\textgreater{}}Data Generation for Sentiment Analysis Task\orangetag{\textless{}/name\textgreater{}}
    \orangetag{\textless{}description\textgreater{}}
        The task is to:
        \begin{enumerate}[leftmargin=*]
            \item Write some financial news that expresses polar sentiments.
            \item Consider each piece of financial news from the viewpoint of an investor, i.e., whether the news could have a positive, negative, or neutral influence on a stock price.
            \item Sentences whose sentiment is not relevant from an economic or financial perspective are deemed neutral.
            \item Ensure a balance of positive, negative, and neutral sentiments across the generated sentences.
            \item The length of each piece of financial news must be between 12--18 words.
            \item Be creative and write unique financial news.
            \item Avoid including explicit sentiment words like “positive,” “negative,” or “neutral” in the sentences themselves.
            \item Generate only the news without adding any additional commentary.
        \end{enumerate}
    \orangetag{\textless{}/description\textgreater{}}

    \orangetag{\textless{}examples\textgreater{}}
    \begin{enumerate}[leftmargin=*]
        \item Cramo slipped to a pretax loss of EUR 6.7 million from a pretax profit of EUR 58.9 million.
        \item In Finland, insurance company Pohjola and the Finnish motorcyclist association have signed an agreement with the aim of improving motorcyclists' traffic safety.
        \item The agreement was signed with Biohit Healthcare Ltd, the UK-based subsidiary of Biohit Oyj, a Finnish public company which develops, manufactures, and markets liquid handling products and diagnostic test systems.
    \end{enumerate}
    \orangetag{\textless{}/examples\textgreater{}}

    \orangetag{\textless{}evaluation\textgreater{}}
        \orangetag{\textless{}metric\textgreater{}}Human evaluation of the diversity, relevance, and quality of generated sentences considering financial context.\orangetag{\textless{}/metric\textgreater{}}
    \orangetag{\textless{}/evaluation\textgreater{}}
\orangetag{\textless{}/task\textgreater{}}
\end{tcolorbox}

\section{Judge LLM Prompts}
\label{sec:appendixL}

\textbf{Prompt For Win Rate}

\begin{tcolorbox}[
  enhanced,
  breakable,  
  colback=boxbg,
  colframe=boxborder,
  arc=15pt,
  boxrule=1pt,
  left=5pt, 
  right=5pt 
]
Please act as an impartial judge and evaluate the quality of the responses provided by two AI assistants to the user question displayed below. You should choose the assistant that follows the user's instructions and answers the user's question better. Your evaluation should consider factors such as the helpfulness, relevance, accuracy, depth, creativity, and level of detail of their responses. Begin your evaluation by comparing the two responses. Avoid any position biases and ensure that the order in which the responses were presented does not influence your decision. Do not allow the length of the responses to influence your evaluation. Do not favor certain names of the assistants. Be as objective as possible. Output your final verdict by strictly following this format: ``[[A]]'' if assistant A is better, ``[[B]]'' if assistant B is better, and ``[[C]]'' for a tie.
\end{tcolorbox}



\textbf{Prompt For Response Accuracy}

\begin{tcolorbox}[
  enhanced,
  breakable,  
  colback=boxbg,
  colframe=boxborder,
  arc=15pt,
  boxrule=1pt,
  left=5pt, 
  right=5pt 
]
You are an impartial and strict judge of answer accuracy.

Given the context and the user instruction below, decide whether the assistant's response is correct and complete.

Return 1 if the response is accurate, 0 if it is inaccurate.
Do not provide any explanation; only return a single digit (1 or 0).

Context: \{context\}

Instruction: \{instruction\}

Response: \{assistant response\}

Judge: Is the response accurate based on the instruction and context?
\end{tcolorbox}


\textbf{Prompt For Task Categorization}

\begin{tcolorbox}[
  enhanced,
  breakable,  
  colback=boxbg,
  colframe=boxborder,
  arc=15pt,
  boxrule=1pt,
  left=5pt, 
  right=5pt 
]
Given this list of categories: \{categories list\},

Classify the following instruction-response pair into exactly one of these categories. 

Return only the category name with no additional text.

Instruction: \{instruction\}

Response: \{assistant response\}

Category: 
\end{tcolorbox}


\textbf{Prompt For Context Relevance}

\begin{tcolorbox}[
  enhanced,
  breakable,  
  colback=boxbg,
  colframe=boxborder,
  arc=15pt,
  boxrule=1pt,
  left=5pt, 
  right=5pt 
]
You are an impartial and strict judge of context relevance.

Given the context, the user instruction, and the assistant's response, decide if the instruction-response pair is relevant to the context.

Return 1 if relevant, 0 if irrelevant.

Do not provide any explanation; only return a single digit (1 or 0).

Context: \{context\}

Instruction: \{instruction\}

Response: \{assistant response\}

Judge: Is this instruction-response pair relevant to the context?
\end{tcolorbox}



\onecolumn





\section{Examples of Finance Documents Synthesized with Template Prompting}
\label{sec:appendixM}
\begin{table}[h]
    \centering
    \footnotesize 
    \renewcommand{\arraystretch}{1.2} 
    \begin{adjustbox}{max width=0.95\textwidth} 
    \begin{tabular}{@{}p{0.95\textwidth}@{}} 
        \toprule
        \textbf{Financial Documents} \\
        \midrule
        \textbf{Apex Financial Partners: Investment Strategies and Risk Management} 
        In today's ever-changing financial landscape, it's crucial to have a solid understanding of investment strategies and risk management. At Apex Financial Partners, we believe that knowledge is power, and our mission is to empower individuals and families to make informed decisions about their financial future. Our team of seasoned professionals offers a comprehensive range of services tailored to meet your unique needs. Whether you're looking to build a robust retirement portfolio, navigate the complexities of estate planning, or explore alternative investment opportunities, we have the expertise to guide you every step of the way... \\
        \midrule
        \textbf{Sustainable Investing: A New Era of Portfolio Diversification} \\
        As the global financial landscape continues to evolve, investors are increasingly seeking diversified portfolios to mitigate risk and maximize returns. One emerging opportunity lies in the realm of sustainable investing, which incorporates environmental, social, and governance (ESG) factors into the investment decision-making process.
        Sustainable investing has gained significant traction in recent years, driven by a growing awareness of the impact that businesses have on the environment and society. Investors are recognizing that companies that prioritize sustainability and ethical practices not only contribute to a better world but also tend to be more resilient and better positioned for long-term success... \\
        \midrule
        \textbf{The Digital Transformation of the Financial Sector} \\
        The global financial landscape has undergone a dramatic transformation in recent years, driven by technological advancements, regulatory changes, and shifting consumer preferences. As we navigate this ever-evolving terrain, it is crucial for financial institutions to adapt and innovate to stay ahead of the curve. One area that has garnered significant attention is the rise of digital banking and mobile finance. With the proliferation of smartphones and the increasing demand for convenience, consumers are seeking seamless and secure ways to manage their finances on-the-go... \\
        \midrule
        \textbf{Achieving Financial Freedom through Smart Budgeting and Debt Management} \\
        The path to financial freedom begins with taking control of your spending habits. One of the most effective ways to do this is by creating a budget and sticking to it. A well-designed budget allows you to allocate your income towards necessary expenses, while also setting aside funds for saving and debt repayment. Start by tracking your monthly income and expenditures. Categorize your expenses into essentials like rent, utilities, and groceries, as well as non-essentials like entertainment and dining out. Identify areas where you can cut back and redirect those funds towards paying off debts or building an emergency fund... \\
        \midrule
        \textbf{Navigating the Evolving Financial Landscape: Trends and Challenges} \\
        The world of finance is a vast and ever-evolving landscape, with new opportunities and challenges arising every day. In this dynamic environment, staying informed and adaptable is crucial for success. Whether you're an investor, a business owner, or simply someone seeking to manage your personal finances, understanding the latest trends and developments can open up a world of possibilities. One area that has seen tremendous growth in recent years is the field of fintech, or financial technology. Innovative companies are leveraging cutting-edge technologies like blockchain, artificial intelligence, and big data to revolutionize how we conduct financial transactions, manage investments, and secure our assets... \\
        \bottomrule
    \end{tabular}
    \end{adjustbox}
    \caption{A sample of five financial documents synthesized by Template prompting using common crawl documents as seeds.}
    \label{tab:template_finance_docs_samples}
\end{table}

\section{Examples of Documents Synthesized with MetaSynth}
\label{sec:appendixN}

\subsection{Finance}
\begin{table}[h]
    \centering
    \footnotesize 
    \renewcommand{\arraystretch}{1.2} 
    \begin{adjustbox}{max width=0.95\textwidth} 
    \begin{tabular}{@{}p{0.95\textwidth}@{}} 
        \toprule
        \textbf{Financial Documents} \\
        \midrule
        \textbf{The Cranston Crypt-Coin Caper: A Cautionary Tale of Cryptocurrency Fraud} \\
        In the burgeoning world of digital currencies, bad actors are always lurking, seeking to exploit the anonymity and decentralized nature of the blockchain for illicit gains. One such case that sent shockwaves through the industry was the infamous Cranston Crypt-Coin scandal. It began in late 2021 when Cranston Crypt-Coin, a new cryptocurrency touted as the future of secure digital transactions, burst onto the scene. The founders, a group shrouded in mystery, launched an aggressive marketing blitz across social media and online forums, making grandiose claims about the coin's revolutionary technology and potential for astronomical returns. Unsuspecting investors, dazzled by the promises of life-changing wealth, poured millions into the new cryptocurrency. However, unbeknownst to them, the entire operation was a meticulously orchestrated pump-and-dump scheme...\\
        \midrule
        \textbf{Where Bloodline Meets Bottom Line: Inside a Millionaire Concierge Club} \\
        Discreetly tucked away in a stately Berkeley Square townhouse, Linton \& Co. represents the apex of white-glove family office services. An elite membership club providing comprehensive wealth management for multi-generational clans with net worths spanning billions, not mere millions.
        ``We like to think of ourselves as outsourced chiefs of staff,'' reveals Sir Edmond Baines, a consummate English gentleman who has presided over Linton's operations for nearly three decades. ``Our members are globally dispersed, dizzyingly wealthy alpha families seeking high-watchmanship oversight of their entire capital engines.'' 
        Baines gestures around his paneled study, its walls lined with portraits of haughty Victorian patrons... \\
        \midrule
        \textbf{The ABACUS Scandal: A Seismic Shift for Structured Finance} \\
        By Jane Doe, Financial Times  The ABACUS scandal, which erupted in 2010, sent shockwaves through the structured finance industry and precipitated sweeping regulatory changes that reshaped the landscape. At the center of the controversy was Goldman Sachs, the Wall Street titan accused of defrauding investors by failing to disclose material information about a synthetic collateralized debt obligation (CDO) called ABACUS 2007-AC1. The SEC alleged that Goldman allowed billionaire John Paulson's hedge fund, which was betting against the CDO, to influence the selection of the underlying mortgage assets. This conflict of interest was not disclosed to investors on the other side of the trade. The SEC charged Goldman and its employee Fabrice Tourre with securities fraud, setting off a legal firestorm. While Goldman ultimately paid \$550 million to settle the charges, the fallout extended far beyond the bank... \\
        \midrule
        \textbf{Quarterly M\&A Report: Analyzing the Microsoft-Activision Blizzard Acquisition} \\
        In January 2022, Microsoft announced its intentions to acquire video game company Activision Blizzard for a staggering \$68.7 billion. This deal, if approved, would be the largest acquisition in the history of the gaming industry and one of the biggest technology acquisitions ever. As a leading M\&A expert, I aim to analyze the strategic rationale, valuation methods, deal dynamics, and potential implications of this groundbreaking transaction. Strategic Rationale: Microsoft's acquisition of Activision Blizzard aligns with its broader strategy of expanding its gaming business and fortifying its position in the metaverse. By acquiring Activision Blizzard's portfolio of popular franchises like Call of Duty, Overwatch, and World of Warcraft, Microsoft can bolster its gaming content and attract more users to its Xbox ecosystem and cloud gaming services. Additionally, the deal provides Microsoft with a significant presence in mobile gaming, a rapidly growing segment. Valuation and Deal Negotiation: Microsoft's offer of \$68.7 billion represents a 45\% premium over Activision Blizzard's pre-announcement share price... \\
        \midrule
        \textbf{LEGAL MEMORANDUM RE: Risk Retention Rules and Disclosure Requirements for Asset-Backed Securities} From the Desk of: [Your Name], Structured Finance Legal Expert Date: [Current Date] This memorandum provides an overview of the risk retention rules and disclosure requirements applicable to asset-backed securities (ABS) issuances under the U.S. federal securities laws. Given the complexity of these regulations and the potential legal risks associated with non-compliance, it is crucial for issuers, sponsors, and other participants in the structured finance market to be well-versed in these requirements. Risk Retention Rules: The risk retention rules, implemented under the Dodd-Frank Wall Street Reform and Consumer Protection Act, require sponsors of ABS transactions to retain an economic interest in a portion of the credit risk associated with the securitized assets. The primary objective of these rules is to align the interests of sponsors with those of investors and incentivize sound underwriting practices. The risk retention requirements vary based on the type of ABS transaction and the underlying asset class. For example, in a residential mortgage-backed securities (RMBS) transaction, the sponsor must retain at least 5\% of the credit risk associated with the pool of residential mortgages. The retained interest can take various forms, such as a vertical slice, a horizontal residual interest, or a combination of both. It is essential for sponsors to carefully structure their risk retention strategy to ensure compliance with the applicable rules while considering the potential impact on the transaction's economics and marketability. Disclosure Requirements: The disclosure requirements for ABS transactions are governed by Regulation AB under the Securities Act of 1933...
        \\ \bottomrule
    \end{tabular}
    \end{adjustbox}
    \caption{A sample of five financial documents synthesized by MetaSynth using common crawl documents as seeds.}
    \label{tab:metasynth_finance_docs_samples}
\end{table}
\clearpage
\subsection{Medicine}
\footnotesize 
\begin{longtable}
{@{}p{0.95\textwidth}@{}} 
    \toprule
    \textbf{Biomedical Documents} \\
    \midrule
    \endfirsthead 
    
    \toprule
    \textbf{Biomedical Documents (continued)} \\
    \midrule
    \endhead

    \textbf{Melatonin and Natural Sleep Aids for Better Sleep} \\
    Sleep disorders are a common and often debilitating issue that affects millions of people worldwide. While there are various pharmaceutical options available to aid in achieving better sleep quality, many individuals seek natural alternatives to avoid potential side effects or dependencies. One such natural remedy that has gained significant attention is the use of melatonin, a hormone naturally produced by the pineal gland in the brain.
    Melatonin plays a crucial role in regulating the body's internal clock, also known as the circadian rhythm. Its levels naturally rise in the evening as it gets darker, signaling to the body that it's time to prepare for sleep. However, in today's modern world... \\
    \midrule

    \textbf{Managing Hypoglycemia: A Critical Concern for Individuals with Diabetes} \\
    Living with diabetes comes with its own set of challenges, but one of the most concerning is the risk of hypoglycemia or low blood sugar levels. This condition can strike without warning and can quickly become a medical emergency if not treated promptly. One of the most effective ways to manage hypoglycemia is by keeping fast-acting glucose tablets or gel on hand at all times. These compact and portable sources of carbohydrates can rapidly raise blood sugar levels within minutes, potentially averting a crisis. The American Diabetes Association recommends that individuals with diabetes always carry a supply of fast-acting glucose, along with testing supplies, as part of their self-care routine... \\
    \midrule

    \textbf{Physical Therapy and Shoulder Rehabilitation: Strengthening and Recovery} \\
    Shoulder pain is one of the most common musculoskeletal issues that physical therapists treat. The shoulder is a complex ball-and-socket joint with an incredible range of motion, making it susceptible to injuries and strain from overuse, poor posture, or trauma. Common shoulder conditions include rotator cuff tears, impingement, tendinitis, and osteoarthritis. As a physical therapist, my goal is to help patients manage their shoulder pain, improve mobility and strength, and prevent further injury through targeted exercises and rehabilitation techniques. One of the most effective exercises for shoulder issues is the ``newspaper arm openings.'' This deceptively simple exercise strengthens the rotator cuff muscles that stabilize the shoulder joint... \\
    \midrule

    \textbf{The Affordable Care Act: Impact and Ongoing Debates in Healthcare Policy} \\
    The Affordable Care Act (ACA), signed into law in 2010, aimed to make healthcare more accessible and affordable for millions of Americans. However, its implementation and long-term impact have been the subject of intense debate and scrutiny within the medical community.
    As a healthcare policy analyst, I've closely examined the ACA's key provisions and their real-world effects. One of the most significant changes brought about by the legislation was the expansion of Medicaid coverage to adults with incomes up to 138\% of the federal poverty level. This move has enabled millions of low-income individuals to gain access to essential healthcare services they previously could not afford... \\
    \midrule

    \textbf{Wisdom Teeth Removal: A Common Oral Surgery with Potential Complications} \\
    Wisdom teeth, also known as third molars, are the last set of permanent teeth to emerge in the mouth, typically between the ages of 17 and 25. While some individuals have enough space in their jaws to accommodate these teeth, many others experience issues due to impaction or lack of room for proper eruption.
    When wisdom teeth become impacted, they can cause a range of problems, including pain, swelling, infection, and damage to adjacent teeth. In such cases, oral surgery is often recommended to remove these problematic teeth... \\
    \bottomrule
\end{longtable}
\captionof{table}{A sample of five biomedical documents synthesized by MetaSynth using common crawl documents as seeds.}
\label{tab:metasynth_biomedical_docs_samples}

\section{Examples of Instructions Synthesized with MetaSynth}
\label{sec:appendixO}
\footnotesize 
\begin{longtable}
{@{}p{0.95\textwidth}@{}} 
    \toprule
    \textbf{Biomedical Instructions} \\
    \midrule
    \endfirsthead 
    
    \toprule
    \textbf{Biomedical Instructions (continued)} \\
    \endhead

    Question: A 45-year-old patient is interested in exploring personalized medicine options to better manage their chronic condition. They have undergone multi-omics testing, which revealed a unique genetic variant linked to their disease. Which of the following would be the most appropriate next step for this patient? Answer options: A. Begin astandard treatment protocol without considering individual genetics B. Consult with a precision medicine specialist to develop a tailored treatment plan C. Undergo tissue engineering to regenerate the affected organ D. Receive a bioelectronicimplant to modulate physiological processes \\
 \\
    \midrule

    A research team is developing a novel nano robotic drug delivery system for targeted cancer treatment. The nano robots are designed to navigate the body's intricate pathways and release precise doses of chemotherapy directly to tumor cells. Which of the following is a potential challenge they may face during development? Answer options: A. Ensuring the nano robots can effectively identify and target cancer cells B. Preventing the body's immune system from attacking the nano robots C. Maintaining structural integrity of the nano robots during extended circulation D. All of the above \\
\\
    \midrule

    A biotechnology company is exploring the use of bio printing and tissue engineering to create personalized organ replacements. They plan to use a patient's own stem cells to seed biomimetic scaffolds, allowing for the regeneration of damaged organs. Which of the following factors would be crucial for the success of this approach? Answer options: A. Sourcing compatible donor stem cells for each patient B. Ensuring the bio printed scaffolds accurately mimic the native organ structure C. Developing methods to induce differentiation of stem cells into desired cell types D. B and C \\
 \\
    \midrule

    Question: A 25-year-old professional soccer player presents with a partial tear of the Achilles tendon sustained during a match. After discussing the available treatment options, the patient expresses interest in exploring orthobiologic therapies for faster recovery. Which of the following orthobiologic treatments would be most appropriate for this patient's condition?A. Stem cell therapy to promote regeneration of the damaged tendon tissue. B. Platelet-rich plasma (PRP) therapy to stimulate the body's natural healing process and reduce inflammation. C. Tissue engineering using a biomaterial scaffold to replace the damaged portion of the Achilles tendon. D. Bone marrow aspiration to harvest stem cells for cartilage regeneration in the ankle joint \\
 \\
    \midrule
    Your patient is a 55-year-old post-menopausal woman, 18 months out from sleeve gastrectomy, presenting with complaints of fatigue, loss of muscle mass/strength despite exercising, frequent hot flashes, brain fog, and lab results showing low ferritin and vitamin D levels. As her naturopathic doctor, devise a detailed treatment protocol incorporating dietary recommendations, supplement regimen, botanical medicines, and lifestyle interventions to address her symptoms which are suggestive of low iron/nutrient status as well as hormonal imbalances related to the surgery. Provide scientific rationale for each component of your treatment plan \\
    \bottomrule
\end{longtable}
\captionof{table}{A sample of five biomedical instructions synthesized by MetaSynth using synthetic documents as seeds.}
\label{tab:metasynth_biomedical_instructions_samples}

\begin{table}[t]
\centering
\begin{tabular}{lc}
\hline
\textbf{Hyper-Parameter} & \textbf{Value} \\
\hline
Learning rate & 1e-5 \\
LR scheduler type & cosine \\
Number of train epochs & 3 \\
Per device train batch size & 16 \\
Per device eval batch size & 8 \\
Gradient accumulation steps & 2 \\
Optimizer & AdamW \\
Adam beta1 & 0.9 \\
Adam beta2 & 0.95 \\
Weight decay & 0.1 \\
Warmup ratio & 0.1 \\
Max gradient norm & 1.0 \\
BF16 & True \\
Gradient checkpointing & True \\
Save steps & 500 \\
Computing infrastructure & 8 A100-40GB GPUs \\
\hline
\end{tabular}
\caption{Hyper-parameters used in continual pre-training experiments.}
\label{tab:hyperparams-continual-pretraining}
\end{table}

\section{Continual Pre-Training Hyperparameters}
\label{sec:appendixP}
All models in this work were trained on a single, high-performance computing node featuring eight interconnected, high-bandwidth NVIDIA A100 GPUs, each possessing 40GB of memory, providing a total of 320GB of GPU memory. The node's processing power was supplied by a high-clock speed, multi-core Intel Xeon processor, paired with 1152 GB of RAM. Table \ref{tab:hyperparams-continual-pretraining} shows the hyperparameters used in continual-pretraining (CPT) experiments.

\twocolumn 

\end{document}